\documentclass[11pt,a4paper]{article}
\usepackage[T1]{fontenc}
\usepackage[utf8]{inputenc}
\usepackage{newtxtext,newtxmath}
\usepackage[margin=2.5cm]{geometry}
\usepackage[english]{babel}
\usepackage{csquotes}
\usepackage{booktabs,tabularx,longtable}
\usepackage{enumitem}
\usepackage{xcolor}
\usepackage{listings}
\usepackage{tikz}
\usepackage{graphicx}
\usepackage{placeins}

\usetikzlibrary{positioning,arrows.meta,shapes.geometric,shapes.symbols}
\tikzset{bx/.style={font=\scriptsize, node distance=3mm and 3.4mm},
  bxinst/.style={draw, rectangle, minimum height=5.5mm, inner sep=1.5pt, align=center},
  bxmodel/.style={draw, signal, signal to=east and west, signal pointer angle=125, minimum height=5mm, inner xsep=0.5pt, inner ysep=1pt, align=center, fill=black!7},
  bxproc/.style={draw, ellipse, minimum height=5.5mm, inner sep=0.5pt, align=center},
  bxactor/.style={draw, isosceles triangle, shape border rotate=90, isosceles triangle apex angle=60, inner sep=0pt, minimum height=4.5mm},
  bxnote/.style={font=\tiny\itshape, text=black!65, align=center, inner sep=0.5pt, text width=13mm},
  bxflow/.style={->, >={Stealth[length=1.5mm]}},
  bxreq/.style={->, >={Stealth[length=1.5mm]}, dashed}}

\usepackage[style=authoryear,backend=biber,maxcitenames=2,uniquelist=false,useprefix=true]{biblatex}
\usepackage[hidelinks]{hyperref}
\usepackage[capitalise,nameinlink]{cleveref}
\crefname{pattern}{Pattern}{Patterns}
\AtBeginBibliography{\sloppy}

\definecolor{fxkw}{HTML}{1F4E9A}\definecolor{fxvar}{HTML}{0E6E5C}
\definecolor{fxstr}{HTML}{9A3412}\definecolor{fxcom}{HTML}{6B7280}
\lstdefinelanguage{framex}{
  emph={external,using,with,AND,OR,NOT,closed,expect,import},
  morekeywords={zzfxdummy}, keywordsprefix=?, alsoletter={?\#},
  sensitive=true, morecomment=[l]{//}, morestring=[b]"}
\lstdefinestyle{fxprog}{numbers=left,numberstyle=\tiny\sffamily\color{black!55},
  numbersep=5pt,xleftmargin=14pt}

\newenvironment{patternform}
  {\begin{description}[style=nextline,leftmargin=1.2em,font=\normalfont\bfseries\small]}
  {\end{description}}
\makeatletter
\newcommand{\knownlit}[1]{\@ifundefined{knownlit@#1}{}{\csname knownlit@#1\endcsname}}
\newcommand{\defknownlit}[2]{\expandafter\def\csname knownlit@#1\endcsname{#2}}
\makeatother
\defknownlit{sensor}{In the literature: neural perception feeding a reasoner, Kautz's
Neuro\,|\,Symbolic \parencite{kautz2022third}; learning from data with symbolic output
\parencite{vanharmelen2019boxology,vanbekkum2021modular}; knowledge-graph construction
by a language model \parencite{pan2024roadmap}.}
\defknownlit{author}{In the literature: domain-model acquisition with experts in
LLM-Modulo \parencite{kambhampati2024llmmodulo}; action-schema generation
\parencite{rani2026nesyllm}; the engineered model of an actor (pattern 1c) of
\textcite{vanbekkum2021modular}, with the model in the engineer's role.}
\defknownlit{translator}{In the literature: the model formalises, a solver answers,
in Logic-LM \parencite{pan2023logiclm}, LINC \parencite{olausson2023linc} and SatLM
\parencite{ye2023satlm}; knowledge-graph question answering and graph-to-text
\parencite{pan2024roadmap}; solver-aided methods \parencite{yang2025nesy}.}
\defknownlit{dispose}{In the literature: rule-based safeguards around a model
\parencite{washizaki2020sep4mla,washizaki2022sepml}; hard critics in LLM-Modulo
\parencite{kambhampati2024llmmodulo}; guardrails \parencite{rebedea2023nemo,liu2025agentpatterns};
gates in prompt chains \parencite{schluntz2024agents}.}
\defknownlit{steer}{In the literature: programmable dialogue rails
\parencite{rebedea2023nemo}, routing and incremental querying
\parencite{schluntz2024agents,liu2025agentpatterns}; knowledge-enhanced inference and
retrieval \parencite{pan2024roadmap,deboer2025patterns}. In these the flows are scripted
or the model drives; here a rule base derives every step from its state.}
\defknownlit{repair}{In the literature: refinement from solver error messages in Logic-LM
\parencite{pan2023logiclm}, tool-grounded critique \parencite{gou2024critic}, the
back-prompt controller of LLM-Modulo \parencite{kambhampati2024llmmodulo}; self-feedback
without an external check \parencite{madaan2023selfrefine,shinn2023reflexion}, whose gains
\textcite{huang2024cannot} trace to oracle stopping signals.}
\defknownlit{competence}{In the literature: separating business logic from the model
\parencite{washizaki2020sep4mla}; routing \parencite{schluntz2024agents}; the inverse
direction, a model that decides when to call a reasoner, is Kautz's Neuro[Symbolic]
\parencite{kautz2022third}; extracting rules from a model is the ``decompiler'' of the
neural-symbolic cycle \parencite{garcez2023third}.}
\defknownlit{hypothesis}{In the literature: few-shot abduction tested against retrieved
events \parencite{shi2023lamp}; hypotheses implemented as programs and verified on
examples \parencite{wang2024hypothesis}. Neither asks for the evidence that decides between
surviving candidates.}
\defknownlit{critic}{In the literature: soft critics in LLM-Modulo
\parencite{kambhampati2024llmmodulo} and cross-reflection
\parencite{liu2025agentpatterns} judge a candidate; we found no catalogue in which the model
reviews the rules themselves.}
\defknownlit{teach}{In the literature: symbolic rules generate training data, Kautz's
Neuro:Symbolic$\to$Neuro \parencite{kautz2022third}; informed learning with prior knowledge
and meta-reasoning for control \parencite{vanbekkum2021modular}; the ``compiler'' of the
neural-symbolic cycle \parencite{garcez2023third}; natural-language principles as labels in
Constitutional AI and RLAIF \parencite{bai2022constitutional,lee2024rlaif}.}
\defknownlit{memory}{In the literature: memory as a component of agent architectures
\parencite{liu2025agentpatterns}; reflections kept across trials
\parencite{shinn2023reflexion}; knowledge graphs as updatable memory for language models
\parencite{pan2024roadmap}. In these the model decides what to keep and recall; here the
rules do, and forgetting withdraws what rested on the forgotten input.}
\defknownlit{select}{In the literature: template-based generation, in which the content
is fixed before the text is planned \parencite{reiter2000nlg}; guardrails that restrict a
reply to retrieved passages \parencite{rebedea2023nemo}. In these the selection is checked
afterwards or the template is written by hand; here the model fills neither content nor
value.}
\defknownlit{failsafe}{In the literature: the deny-overrides combining algorithm of
access-control policies, in which any applicable deny wins over permits
\parencite{oasis2013xacml}; layered guardrails in which every rail can block
\parencite{rebedea2023nemo}. Here a model is one of the sources, and its silence counts
as a deny.}

\title{Hybrid AI Patterns\\[0.3em]
\large Thirteen Ways to Couple Language Models with Rules, and How to Choose Among Them}
\author{%
  \begin{minipage}[t]{0.31\textwidth}
    \centering\normalsize
    Siegfried Handschuh \\
    {\small Institute of Computer Science} \\
    {\small University of St.~Gallen (HSG)} \\
    {\footnotesize\texttt{siegfried.handschuh@unisg.ch}}
  \end{minipage}%
  \hspace{0.02\textwidth}%
  \begin{minipage}[t]{0.31\textwidth}
    \centering\normalsize
    Götz-Henrik Wiegand \\
    {\small Institute of Computer Science} \\
    {\small University of St.~Gallen (HSG)} \\
    {\footnotesize\texttt{goetz-henrik.wiegand@unisg.ch}}
  \end{minipage}%
  \hspace{0.02\textwidth}%
  \begin{minipage}[t]{0.31\textwidth}
    \centering\normalsize
    Lorena Raichle \\
    {\small Institute of Computer Science} \\
    {\small University of St.~Gallen (HSG)} \\
    {\footnotesize\texttt{lorena.raichle@unisg.ch}}
  \end{minipage}%
}
\date{October 2026}

\AddToHook{cmd/section/before}{\clearpage}

\begin{document}
\maketitle

\begin{abstract}
Language models handle text well, and rule engines give conclusions with proofs.
Systems that combine the two are usually described by example, and the same coupling
turns up under different names in agent frameworks, guardrail toolkits and
neuro-symbolic research. This document defines thirteen patterns for coupling a language
model with an executable rule language, in the catalogue form introduced for
object-oriented design: intent, motivation, applicability, participants, collaborations,
consequences, implementation, known uses and related patterns. Four patterns concern
what crosses the boundary between model and rules, five who controls whom, three how
the model takes part in the reasoning, and one how knowledge is kept over time. Five invariants hold for every combination, and five failure patterns name the couplings that break them. We
relate the catalogue to published pattern catalogues and taxonomies of hybrid and
neuro-symbolic systems and show where they overlap with it. A decision guide leads from
a problem to a pattern. It rests on our measurements, including the cases in which a
model alone did as well as the hybrid or better, and it reports what each design cost.
\end{abstract}

\tableofcontents

\section{Introduction}

More and more organisations let language models act on documents, conversations and
transactions that are governed by explicit rules: a regulation, a protocol, an
authority limit. Two designs are possible. The rules can be written into the prompt and
the model asked to apply them, or the rules can stay executable and the model be given
the parts of the task that rules cannot do: reading text, phrasing a question, proposing
a candidate. Between these two designs lie many couplings, and practitioners tend to
reinvent them project by project.

A pattern names a recurring problem in a context and the core of a solution that can be
reused without being copied \parencite{alexander1977pattern}. The catalogue form of \textcite{gamma1994design} made patterns a working vocabulary for
object-oriented design, and \textcite{buschmann1996posa} carried it to the architecture of
whole systems: each pattern has a name, an intent, a motivating example, conditions of
applicability, a structure of participants and collaborations, consequences, notes on
implementation, known uses and related patterns. We use the same form for couplings of
language models with rules. A shared name lets designers say which coupling they mean.
The consequences make the trade-offs explicit, including the cases where a model alone
is the better choice, and the known uses tie each pattern to running systems and to
measurements.

The patterns were found in the applications of FrameX, a rule and knowledge engine
based on F-Logic \parencite{kifer1995flogic}, and checked against published catalogues
(\cref{sec:related}). They do not depend on FrameX: any rule engine that can request a
model call, store the answer as data and derive conclusions over it can host them. The
examples use FrameX's notation because we made the measurements in it.
An early version of the FrameX engine and framework is publicly available at
\url{https://gitlab.com/ds-unisg/framex-core}; a full release is forthcoming.

\paragraph{Contributions.} The following contributions are made in this document: 

\noindent\textbf{1:} Thirteen patterns in catalogue form, composed from four elementary moves, with
five invariants that hold for all of them
(\cref{sec:foundations,sec:boundary,sec:control,sec:reasoning,sec:time}), and five failure
patterns that break the invariants (\cref{sec:failures}).

\noindent\textbf{2:} A mapping onto published pattern catalogues and taxonomies that shows which patterns
are known, which are new, and which published patterns this catalogue lacks
(\cref{sec:related}).

\noindent\textbf{3:} A decision guide from problem to pattern, grounded in
pre-registered measurements (\cref{sec:guide}).

\section{Foundations}
\label{sec:foundations}

\subsection{What FrameX adds for coupling}\label{sec:coupling}

The listings use FrameX, a rule language in the F-Logic tradition. We assume that the
reader knows its basic syntax: frames \texttt{?X:Class[slot -> ?V]}, rules
\texttt{head <- body}, \texttt{AND}, \texttt{NOT}, queries \texttt{?- \ldots}, and the
open-world reading of \emph{unknown}. Every program declares its world. Under
\texttt{world open.} a missing fact is \emph{unknown}, and \texttt{NOT} gives
\emph{true} only over a slot, method or class declared closed
(\texttt{closed slot allowsSlot.}, \texttt{closed method refuted/1.}, where \texttt{/1} is
the number of parameters,
\texttt{closed class Admitted.}); over anything else the negation stays \emph{unknown}.
The programs also compute values (\texttt{?M := ?N + 1}), compare them (\texttt{<},
\texttt{>}, \texttt{!=}), use methods with parameters (\texttt{slot(?I) -> ?V}) and
strings as objects (\texttt{"mbi":Programme}), and write an object whose name is a term
in backquotes (\texttt{\textasciigrave read(d1)\textasciigrave}). This section
introduces only what coupling with a model adds. All thirteen patterns rest on it.

\paragraph{External requests.} A rule asks for something outside the engine with the
keyword \texttt{external}:
\begin{lstlisting}[language=framex]
external read(?D) using _llm[prompt -> "Report what the document states ..."; document -> ?T]
  <- ?D:Document[pending -> true; text -> ?T].
\end{lstlisting}
The engine evaluates this rule like any other. Its conclusion is a request: an object \texttt{\textasciigrave read(d1)\textasciigrave} with the fields in brackets as attributes, the
tool (\texttt{tool\#name -> "\_llm"}), a mark \texttt{tool\#requested -> true}, and link
facts for the request's arguments (\texttt{tool\#arg1 -> d1}, and \texttt{tool\#arg2}
for a second argument). The engine never calls a
network. A \emph{host}, a program beside the engine, reads the open requests, runs the
tool and adds its answer as a \emph{receipt}; the engine then recomputes, and rules that
read the receipt can fire, including rules that issue the next request. The keyword is
kept although the tool name would identify a request, because such a rule follows other
laws than a derivation: a request is carried out at most once in a run and is never
withdrawn, whereas a derived fact disappears when its premises do. Every point at which a
program reaches outside the engine is marked by it.

\paragraph{Receipts.} An answer returns as facts about the request object:

\begin{lstlisting}[language=framex]
// Written by the host after the call.
`read(d1)`[tool#status -> "succeeded"; tool#completed -> true;
    row(1, "slot") -> "customer"; row(1, "value") -> "Acme AG";
    row(1, "quote") -> "Acme AG (the Customer)"; row(1, "quoteFound") -> true].

// Read by the rules, joined to the document through the request's argument.
?D[slot(?I) -> ?S]
  <- ?R[tool#name -> "_llm"; tool#arg1 -> ?D; row(?I, "slot") -> ?S].
\end{lstlisting}
A scalar answer field \texttt{key} becomes \texttt{result("key")}; the reserved field
\texttt{rows}, a list of at most 64 flat objects, becomes \texttt{row(N, "key")}. Every
receipt ends with \texttt{tool\#completed -> true} and \texttt{tool\#status}, which is
\texttt{"succeeded"} or \texttt{"failed"}. A failed call (the model did not answer in
time, the reply was not valid JSON, a field the host needs was missing) adds no
\texttt{result} and no \texttt{row}, only an attribute \texttt{error} with the reason,
\texttt{\textasciigrave read(d1)\textasciigrave[error -> "model response is not valid
JSON; no facts accepted"]}. A rule that reads \texttt{tool\#status -> "failed"} keeps such
a case apart from a rejection (invariant~5). Answers must be flat: values are stored as
literals and never evaluated, nested objects and lists are refused, and a \texttt{null}
states nothing. Rules usually key rows by method parameters (\texttt{?D[slot(?I) ->
\ldots]}); since FrameX allows function terms in rule heads, a rule can also give each row
an object of its own, \texttt{st(?U, ?I):Statement[\ldots]} (\cref{pat:memory}).
\Cref{tab:receipts} lists what each tool reads from a request and writes into its
receipt.

\paragraph{Identity and order.} A request is identified by its name and arguments.
\texttt{read(d1)} runs once; a second attempt needs a new identity, as in
\texttt{propose(?T, ?N)} with a step number (\cref{pat:dispose}). A request without
arguments, \texttt{external propose using \ldots}, is the object \texttt{propose}
itself, and its receipt is read as \texttt{propose[\ldots]} (\cref{pat:hypothesis}).
Instances of one
request name form a queue; requests of different names that are open at the same time
need \texttt{tool\#order -> N}, and the lowest goes first. Two values for one request
field are an error. A run ends when a rule derives
\texttt{workflow[status -> \ldots]} or when no request is open.

\paragraph{Tools.} The tool after \texttt{using} names what the host does
(\cref{tab:tools}). Tools beyond those shipped with FrameX come from adapter files that
the operator names when starting the host; a program can never load code.

\begin{table}[ht]
\centering\small
\caption{Tools used in this document.}
\label{tab:tools}
\begin{tabularx}{\linewidth}{@{}>{\raggedright\arraybackslash}p{3.3cm}>{\raggedright\arraybackslash}p{3.2cm}X@{}}
\toprule
Tool & Provided by & What the host does \\
\midrule
\texttt{\_llm} & \texttt{framex.tools} & one model call; see below \\
\texttt{\_query}, \texttt{\_evidence\_needs} & \texttt{framex.tools} & a query on the rules themselves; the evidence that would settle a goal \\
\texttt{\_turn}, \texttt{\_judge}, \texttt{\_repairTurn} & \texttt{framex.dialogue} & the dialogue model phrases a question; a judge decides a criterion; a repeated turn after a broken reply \\
\texttt{\_ask} & \texttt{framex.dialogue} & one question to a person, as written, without a model \\
\texttt{\_checkCandidate} & \texttt{framex.candidates}, by adapter & checks a drafted rule against cases, before and after \\
\texttt{\_forgetAnswers} & \texttt{framex.tools} & removes stored model answers that contain a forgotten text \\
\bottomrule
\end{tabularx}
\end{table}

\begin{table}[ht]
\centering\small
\caption{What each tool reads from a request and writes into its receipt. Every receipt
also has \texttt{tool\#status} and \texttt{tool\#completed}; a failed call has
\texttt{error} instead of results.}
\label{tab:receipts}
\begin{tabularx}{\linewidth}{@{}>{\raggedright\arraybackslash}p{2.8cm}>{\raggedright\arraybackslash}p{4.2cm}>{\raggedright\arraybackslash}X@{}}
\toprule
Tool & Request fields & Receipt \\
\midrule
\texttt{\_llm} & \texttt{prompt} (or \texttt{with P}) and any named fields; \texttt{groundIn}, \texttt{mustSay}, \texttt{contextKey}, \texttt{tool\#maxContext} for the host & the reply's fields as \texttt{result("k")} and \texttt{row(N, "k")}; with \texttt{groundIn} \texttt{quoteFound} and \texttt{valueQuoted} per row; with \texttt{mustSay} \texttt{saysDecision}; with \texttt{contextKey} \texttt{tool\#sent("id")} \\
\texttt{\_query} & \texttt{query} & \texttt{result("status")} (\texttt{true}, \texttt{false}, \texttt{unknown} or \texttt{bindings}) and the whole answer as JSON in \texttt{result("text")} \\
\texttt{\_checkCandidate} & \texttt{candidate} & \texttt{status} (\texttt{checked}, \texttt{rejected}, \texttt{error}), \texttt{feedback}, and after evaluation \texttt{regressions}, \texttt{failedAfter}, \texttt{newlyFailing}, \texttt{newlyPassing}, \texttt{newViolations}, \texttt{changesConclusions}, all as \texttt{result("k")} \\
\texttt{\_turn} & \texttt{question}, \texttt{attempt} & \texttt{answer}, \texttt{utterance}; after a broken reply \texttt{formatError} and \texttt{received} \\
\texttt{\_judge} & \texttt{criterion}, \texttt{question}, \texttt{answer} & \texttt{verdict} (\texttt{yes}, \texttt{no}, \texttt{unclear}), \texttt{reason}, \texttt{label}; after a broken reply \texttt{formatError} and \texttt{received} \\
\texttt{\_ask} & \texttt{question} & \texttt{question}, \texttt{answer} \\
\texttt{\_forgetAnswers} & \texttt{text} & \texttt{removed}, the number of stored answers deleted \\
\bottomrule
\end{tabularx}
\end{table}

\paragraph{The model call.} \texttt{\_llm} sends \texttt{prompt} as the instruction
and the other fields as one JSON message, so a rule hands over separate, traceable
slots. Fields that start with \texttt{tool\#} and the fields read by the host
(\texttt{groundIn}, \texttt{mustSay}, \texttt{contextKey}) are not sent. Long prompts live in prompt objects (\texttt{readPrompt:Prompt[text -> "..."]})
that a request names with \texttt{using \_llm with readPrompt[...]}. Every answer is
stored with its date, model and messages and replayed on the next identical request
until a refresh is asked for; a run can therefore be repeated without the model. Which
model answers is set in the host's environment (a local server or a hosted endpoint),
and keys never enter programs or receipts.

\paragraph{Measurements on text.} Two request fields are read by the host and never
shown to the model. With \texttt{groundIn -> ?Text}, every answer row gets
\texttt{quoteFound} (its quote occurs verbatim in that text) and \texttt{valueQuoted}
(its value occurs in its quote). With \texttt{mustSay -> ?Word}, the receipt gets
\texttt{saysDecision} (the reply contains the word). Both are measurements; whether a
row or a reply counts is decided by rules.

\paragraph{Rule-selected context.} A request with \texttt{contextKey -> ?Step} receives
in its field \texttt{context} exactly the statements that rules select with
\texttt{?S[selectedFor -> step]}, ordered by identifier and at most
\texttt{tool\#maxContext} of them; more makes the call fail.
Each statement goes to the model as one JSON object with its identifier (\texttt{id},
the object's printed name, such as \texttt{st(u2, 1)}) and its slots; the default limit
is 32. The receipt records only the identifiers of what was sent, as
\texttt{tool\#sent("st(u2, 1)") -> st(u2, 1)}. A reply cites statements by
identifier in its rows, so rules can check that every cited statement was sent and is
current (\cref{pat:memory}).

\paragraph{One trap.} Because the fields of a request become attributes of its object, a
condition without a class can match the request itself: \texttt{?Q[decision -> ?D]} also
matches \texttt{\textasciigrave phrase(q1)\textasciigrave} when that request has a field \texttt{decision}. FrameX
stops such a run and names the cause. A field named like a slot that rules derive for the
receipt gives it two values; \texttt{framex validate} warns about it. Give rule subjects a
class (\texttt{?Q:Request[...]}) and name request fields apart from derived slots.

\paragraph{The other direction.} A model can also call the rules, through a protocol of
read-only operations (\texttt{query}, \texttt{explain}, \texttt{why\_not},
\texttt{evidence\_needs}, \texttt{validate}, \texttt{preview}).

The implementation of each pattern shows the whole chain: the rule that requests the
model call, the rules that read its receipt, and the rules that decide. Each program is
printed in full, with its declarations, its data, its prompts and its queries; it runs
against a model or, from recorded answers, without one%
.

\subsection{Five invariants}

Every pattern and every combination of patterns keeps five invariants.
\begin{enumerate}[leftmargin=*]
\item \textbf{Model text is data.} The engine never executes model-written rules or
code unchecked; an answer is a receipt that rules read. A rule or query written by a model
is a candidate that a check admits or refuses before it runs
(\cref{pat:author,pat:translator}).
\item \textbf{Every model answer is a dated observation} with its prompt, model and
inputs, so a run can be replayed and audited.
\item \textbf{Rules decide.} Whether a statement counts, which step comes next, when a
task is complete: the rules derive each of these.
\item \textbf{Evidence decides acceptance.} A model's confidence is stored as a number.
A statement is accepted on evidence the rules can check, such as a quote found in the
source.
\item \textbf{Failures stay distinct.} A failed call, a reply in the wrong format, an
\emph{unclear} judgement and a logical \emph{unknown} are four different results.
\end{enumerate}
The division of labour follows from these invariants. The model handles perception,
language and ambiguity. The rules handle whatever must be guaranteed, audited or
explained. What passes between them is an explicit, validated record.

\subsection{Four elementary moves}\label{sec:moves}
\begin{figure}[ht]
\centering
\begin{tikzpicture}[font=\small,
  box/.style={draw, rounded corners, minimum width=2.6cm, minimum height=0.9cm, align=center},
  ->, >={Stealth[length=2mm]}]
\node[box, fill=black!6] (r) {Rules};
\node[box, right=3.2cm of r] (m) {Model or tool\\(through the host)};
\node[box, left=3.2cm of r] (p) {Person};
\node[box, below=1.3cm of r] (f) {Facts};
\draw (r.north east) to[bend left=18] node[above]{request} (m.north west);
\draw (m.south west) to[bend left=18] node[below]{receipt} (r.south east);
\draw (r) -- node[right]{accept} (f);
\draw (r.north west) to[bend right=18] node[above]{ask} (p.north east);
\draw (p.south east) to[bend right=18] node[below]{answer, as receipt} (r.south west);
\end{tikzpicture}
\caption{The four elementary moves. Every arrow that enters the rules is a receipt;
only the accept move writes facts.}
\label{fig:moves}
\end{figure}
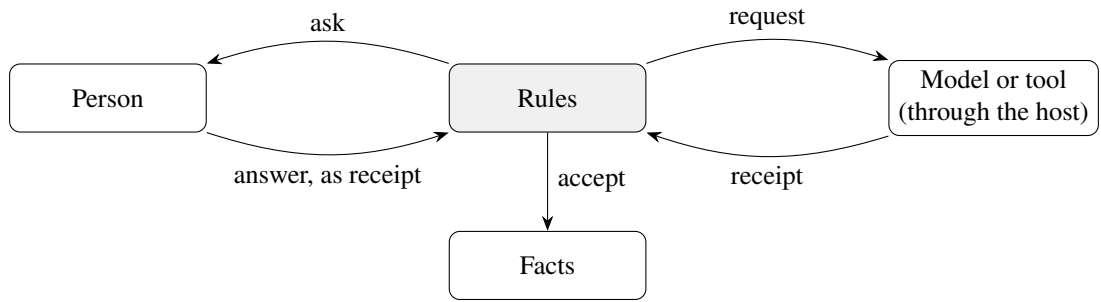

Every pattern is assembled from four moves between the rules and the outside
(\cref{fig:moves}).
\begin{description}[leftmargin=1.2em,font=\normalfont\bfseries\small]
\item[Request] A rule whose conditions hold issues an external request to a model or a
tool. What the model sees is what the rule puts into the request.
\item[Receipt] The answer returns as dated facts about the request, joined to its subject
through the request's arguments. Nothing else enters the engine from outside.
\item[Accept] A rule decides whether what a receipt proposes becomes a fact, is refused
with its reason, or stays \emph{unknown}.
\item[Ask] A rule derives a question for a person, or the evidence that would settle an
open goal; the answer returns as a receipt.
\end{description}

\Cref{tab:moves} shows how the thirteen patterns compose the moves. They differ in what
triggers the request and in what the accept move decides; Critique and Repair is the only
pattern that repeats a move, under a bound.

\begin{table}[ht]
\centering\small
\caption{The patterns as compositions of the four moves.}
\label{tab:moves}
\begin{tabularx}{\linewidth}{@{}>{\raggedright\arraybackslash}p{3.1cm}>{\raggedright\arraybackslash}p{2.9cm}>{\raggedright\arraybackslash}p{3.0cm}X@{}}
\toprule
Pattern & Moves & Request triggered by & Accept decides \\
\midrule
1 Model as Sensor & request, receipt, accept & a pending input & whether a reading becomes a fact \\
2 Model as Author & request, receipt, accept, ask & a policy text or a failing case & whether a draft is offered; a person adopts it \\
3 Model as Translator & request, receipt, accept, request, receipt & a person's question & whether the request may run; whether the reply states the rules' decision \\
4 Model Proposes, Rules Dispose & request, receipt, accept & the next step of a task & admit, refuse with a reason, or narrow \\
5 Rules Steer the Model & request, receipt, ask & the derived step & which step follows; the judge's reading as in 1 \\
6 Critique and Repair & accept, then request again, bounded & a refused reply & the kind of failure and when to stop \\
7 Division by Competence & accept, then request or ask & an \emph{unknown} or a conflict & what the rules decide themselves \\
8 Model as Hypothesis Generator & request, receipt, accept in a copy, ask & an observation to explain & which candidates survive; ask names the deciding evidence \\
9 Model as Critic of the Rules & request, receipt, accept, ask & a change before release & which remarks become findings; a reviewer disposes \\
10 Rules Teach the Model & request, receipt, accept & labelling, selection, grading & the grade of a model answer \\
11 Rule-Governed Memory & request, receipt, accept; later request & an utterance; the next step & admission, supersession, withdrawal; what enters the prompt \\
12 Model Selects, Rules Write & request, receipt, accept, twice & a person's question; a value to state & whether an identifier names a passage; whether the words leave the value to the rules \\
13 Fail-Safe Union & request, receipt, accept, from several sources & every case, for every source & the strictest verdict in the declared order; silence as caution \\
\bottomrule
\end{tabularx}
\end{table}

\subsection{The catalogue form}

Each pattern is described under the headings of \textcite[sec.~1.3]{gamma1994design},
adapted to couplings. \emph{Structure} is drawn in the boxology notation of hybrid
learning and reasoning systems \parencite{vanharmelen2019boxology,vanbekkum2021modular},
with the generative model of \textcite{deboer2025patterns}: rectangles are instances
(\emph{data} such as text, or \emph{symbol}), hexagons are models (\emph{model:gen} for a
language model, \emph{model:sem} for a rule base), ovals are processes (\emph{infer},
\emph{transform}, \emph{generate}) and triangles are actors, here persons. The dashed
arrows are ours: they mark a request, from the process that issues it to the process
that answers, and so show who triggers whom where the solid arrows show what flows. A
reply of the model is text; where it becomes symbols, the receipt is drawn as a
transformation. \Cref{tab:legend} in \cref{sec:legend} lists every symbol;
\cref{sec:boxology} relates the notation to the catalogue.
\emph{Sample code} is folded into \emph{Implementation}: each pattern prints its
complete program and draws it as a \emph{rule graph}, generated from the program by a
script of the companion folder. Its boxes are the facts, the requests, the receipts and the
derived slots; a small circle is one rule, labelled with the line where it starts in the
listing (``l.~23'' for line 23), with arrows from what it reads to what it derives (\cref{tab:rglegend}). The
boxology shows the roles of a pattern, the rule graph the chain of one program. For the
dialogue flow of pattern~5 the graph is a flow diagram of its steps. We add one
heading, \emph{Evidence}, for measured results where they exist, because the
consequences of a coupling are empirical in a way the consequences of a class structure
are not.

Patterns 1 to 11 were drawn from the applications of FrameX. Patterns 12 and 13 were added
later from couplings that recurred in the submissions to a hackathon; they keep their
numbers, which other documents cite, and stand in the group they belong to
(\cref{tab:overview}).

\begin{table}[ht]
\centering\small
\caption{The thirteen patterns in four groups.}
\label{tab:overview}
\begin{tabularx}{\linewidth}{@{}rlXX@{}}
\toprule
\# & Pattern & The model & The rules \\
\midrule
\multicolumn{4}{@{}l}{\emph{What crosses the boundary}} \\
1 & Model as Sensor & turns text into proposed facts & accept or reject each fact \\
2 & Model as Author & drafts rules, tables, flows, tests & check them; a person adopts \\
3 & Model as Translator & turns questions into checked requests, proofs into prose & answer and prove \\
12 & Model Selects, Rules Write & chooses passage identifiers, phrases around values & show passages and values word for word \\
\multicolumn{4}{@{}l}{\emph{Who controls whom}} \\
4 & Model Proposes, Rules Dispose & proposes an answer or action & admit, refuse with a reason, narrow \\
5 & Rules Steer the Model & answers what it is asked & decide what to ask, when, with what \\
6 & Critique and Repair & tries again & classify the fault, state it, bound retries \\
7 & Division by Competence & handles what the rules leave open & decide what they can, hand on the rest \\
13 & Fail-Safe Union & gives a verdict that can only tighten & combine by a declared order; silence counts as caution \\
\multicolumn{4}{@{}l}{\emph{The model in the reasoning}} \\
8 & Model as Hypothesis Generator & proposes possible causes & test each, ask for deciding evidence \\
9 & Model as Critic of the Rules & reviews rules and conclusions & record the critique, open a review \\
10 & Rules Teach the Model & learns from labels, examples, grades & supply them with proofs \\
\multicolumn{4}{@{}l}{\emph{Over time}} \\
11 & Rule-Governed Memory & proposes what to remember, cites what it used & admit, supersede, withdraw, select \\
\bottomrule
\end{tabularx}
\end{table}


\section{Patterns at the boundary: what crosses it}
\label{sec:boundary}

\FloatBarrier
\subsection{Pattern 1: Model as Sensor}
\label{pat:sensor}

\begin{patternform}
\item[Intent] Let the model read what rules cannot (free text, documents, answers in a
conversation) and turn it into proposed facts; a rule decides whether each proposal
becomes a fact.

\item[Also known as] Extraction, classification as a predicate, perception, learned
predicate, judge.

\item[Motivation] An assignment review must know whether a contract names a
signatory with authority, whether a consent form is signed and which party a letter
addresses. Rules can decide what follows from these facts, but they cannot read the
documents. A model can read them, but it also reports things the document does not
say, and a fluent wrong reading looks like a right one. The pattern lets the model
read and keeps the decision about what counts with the rules: the model reports each
statement with a verbatim quote, the host measures whether the quote occurs, and a
rule accepts a statement only if the quote was found and the kind of document can
make such a statement.

\item[Applicability] Use it when
\begin{itemize}
\item the inputs are unstructured (text, speech, images) but the decision over them is
known and can be written as rules;
\item a reading can be checked by something measurable: a quote that occurs, a value
from a closed list, a hash of the source;
\item the same reading is needed by several rules, so it should exist once as a fact
with a provenance.
\end{itemize}
Do not use it when the decision itself is what is unknown (then see
Division by Competence, \cref{pat:competence}).

\item[Structure]\leavevmode
\begin{center}
\begin{tikzpicture}[bx]
\node[bxinst] (a1) at (0,0) {data:\\text};
\node[bxnote, below=0.6mm of a1] {document};
\node[bxproc, right=of a1] (a2) {infer};
\node[bxmodel, above=of a2] (a2m) {model:\\gen};
\node[bxnote, above=0.3mm of a2m] {language model};
\draw[bxflow] (a2m)--(a2);
\draw[bxflow] (a1)--(a2);
\node[bxinst, right=of a2] (a3) {data:\\text};
\node[bxnote, below=0.6mm of a3] {reply};
\draw[bxflow] (a2)--(a3);
\node[bxproc, right=of a3] (a4) {trans-\\form};
\node[bxnote, below=0.6mm of a4] {receipt};
\draw[bxflow] (a3)--(a4);
\node[bxinst, right=of a4] (a5) {symbol};
\node[bxnote, below=0.6mm of a5] {proposed};
\draw[bxflow] (a4)--(a5);
\node[bxproc, right=of a5] (a6) {infer:\\deduce};
\node[bxmodel, above=of a6] (a6m) {model:\\sem};
\node[bxnote, above=0.3mm of a6m] {rules};
\draw[bxflow] (a6m)--(a6);
\draw[bxflow] (a5)--(a6);
\node[bxinst, right=of a6] (a7) {symbol};
\node[bxnote, below=0.6mm of a7] {accepted, refused, unknown};
\draw[bxflow] (a6)--(a7);
\draw[bxreq] (a6.south) .. controls +(0,-10mm) and +(0,-10mm) .. node[pos=0.5, below, inner sep=1pt, font=\tiny]{request} (a2.south);
\end{tikzpicture}
\end{center}

\item[Participants] \emph{Model}: reads one input and answers one question about it in
a fixed format. \emph{Host}: calls the model, measures what can be measured (format,
quote occurrence), stores the answer as a receipt. \emph{Rules}: request the reading,
accept or reject each proposed fact with a reason. \emph{Source}: the document or
utterance, identified by hash.

\item[Collaborations] A rule fires an external request for a document that is
pending; the host returns a receipt with the model's answer and its measurements; an
acceptance rule turns admissible statements into facts; rejected statements keep
their reason. A later revocation of the source withdraws what rested on it.

\item[Consequences] Each fact has a provenance (source, model, prompt version, date)
and a reason for acceptance. An answer that cannot be grounded stays a rejected
statement with its reason. A model's \emph{unknown} leaves the criterion open. The cost is one call per input and question; a
plausible but wrong value that passes every measurable check slips through this
pattern and needs a second reader, a person or a cross-check.

\item[Implementation] Ask for one thing per call: a value from a fixed list, or a
statement with its quote. The request names the source text as \texttt{groundIn}, which
the model does not see; the host then records for every row of the answer whether its
quote occurs verbatim in that text (\texttt{quoteFound}). The program is complete: it
declares its world and the slot it negates, states what a contract may say, gives one
contract and the prompt, and then the steps every pattern has: the request, the rule
that reads the receipt (row \texttt{?I} is one statement about document \texttt{?D}),
the rules that accept or reject, and the rule for a failed call.
\begin{lstlisting}[style=fxprog,basicstyle=\footnotesize\ttfamily,language=framex]
// Pattern 1, Model as Sensor: the model reads, the rules decide what counts.
world open.
closed slot allowsSlot.

// What a kind of document can state, and one document to read.
"contract"[allowsSlot -> "customer"; allowsSlot -> "provider"].
d1:Document[pending -> true; kind -> "contract";
    text -> "Service Agreement between Beta GmbH (the Provider) and Acme AG (the Customer). This agreement is governed by Swiss law."].

// The instruction the model receives.
readPrompt:Prompt[text -> "Report what the document states: one row {slot, value, quote} per statement, slot one of customer, provider, governingLaw, the quote copied verbatim. Answer as JSON {\"rows\": [...]}."].

// Request: one call per pending document; groundIn makes the host check every quote.
external read(?D) using _llm with readPrompt[document -> ?T; groundIn -> ?T]
  <- ?D:Document[pending -> true; text -> ?T].

// Receipt: each answer row becomes a proposed statement about the document.
?D[slot(?I) -> ?S; value(?I) -> ?V; quoteFound(?I) -> ?Q]
  <- ?R[tool#name -> "_llm"; tool#arg1 -> ?D; row(?I, "slot") -> ?S;
        row(?I, "value") -> ?V; row(?I, "quoteFound") -> ?Q].

// Accept: a quoted statement that this kind of document may make.
?D[accepted(?I) -> ?V]
  <- ?D:Document[kind -> ?K; slot(?I) -> ?S; value(?I) -> ?V;
        quoteFound(?I) -> true] AND ?K[allowsSlot -> ?S].

?D[rejected(?I) -> "quote not found in the document"]
  <- ?D[quoteFound(?I) -> false].

?D[rejected(?I) -> "not a statement this kind of document can make"]
  <- ?D:Document[kind -> ?K; slot(?I) -> ?S] AND NOT ?K[allowsSlot -> ?S].

// A failed call (no answer, or a reply that is not valid JSON) proposes nothing;
// the document stays unread, with the host's reason.
?D[notRead -> ?E]
  <- ?R[tool#name -> "_llm"; tool#arg1 -> ?D; tool#status -> "failed"; error -> ?E].

?- d1[accepted(?I) -> ?V].
?- d1[rejected(?I) -> ?Why].
?- d1[notRead -> ?E].
\end{lstlisting}
\begin{figure}[p]
\centering
\includegraphics[width=\linewidth,height=0.82\textheight,keepaspectratio]{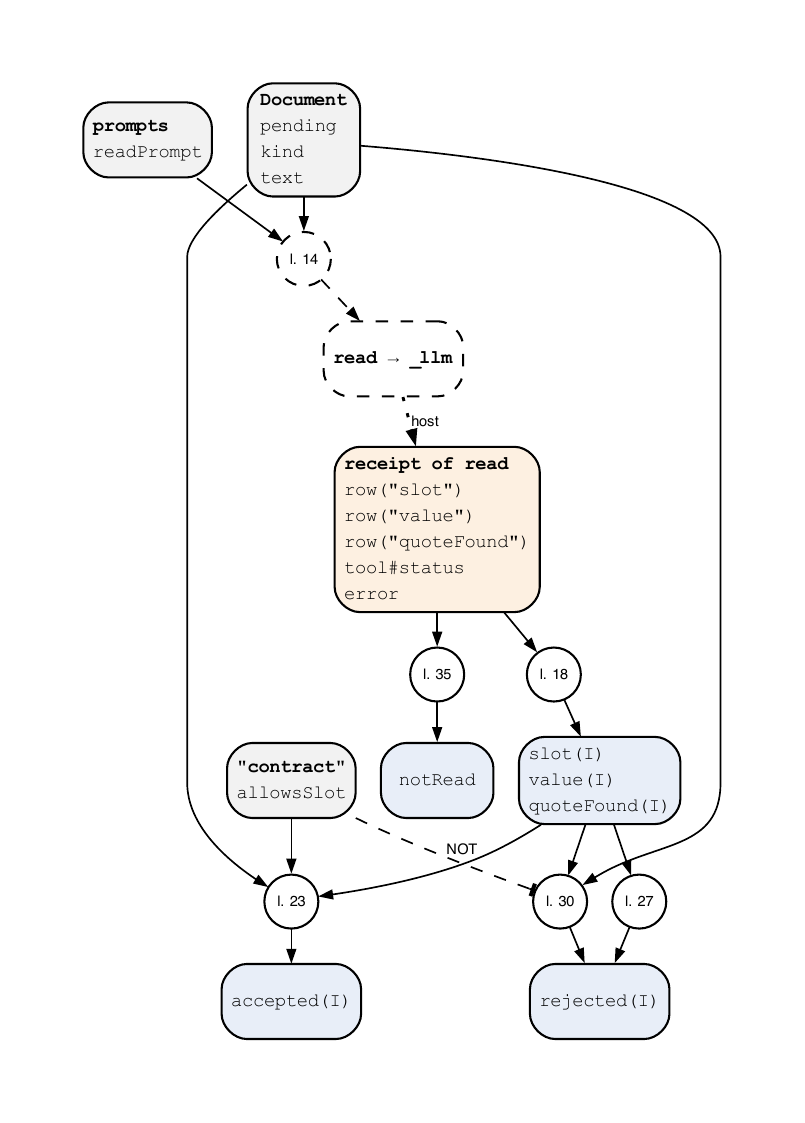}
\caption{Rule graph of \texttt{p01-sensor.fx}, generated from the program. A circle is one rule; ``l.~23'' means that the rule starts in line 23 of the listing. Legend in \cref{tab:rglegend}.}
\label{fig:rg-sensor}
\end{figure}
\Cref{fig:rg-sensor} draws the program as a rule graph.
A statement the model does not report stays \emph{unknown}: no rule accepts or rejects
it. A failed call is no rejection either; \texttt{notRead} keeps the host's reason
(invariant~5).

\item[Library form] The rules above are the same for every source, so they can live in a
library, as the dialogue flows of pattern~5 do. With the library \texttt{sensor.fx}
(\cref{sec:libraries}) loaded first, the program states facts only, and the library
derives \texttt{sensor\#accepted}, \texttt{sensor\#rejected} and
\texttt{sensor\#notRead} for every source:
\begin{lstlisting}[style=fxprog,basicstyle=\footnotesize\ttfamily,language=framex]
// Pattern 1 in library form: load sensor.fx first. Same document, same prompt as p01-sensor.fx.
world open.

"contract"[sensor#allows -> "customer"; sensor#allows -> "provider"].

readPrompt:Prompt[text -> "Report what the document states: one row {slot, value, quote} per statement, slot one of customer, provider, governingLaw, the quote copied verbatim. Answer as JSON {\"rows\": [...]}."].

d1:sensor#Source[sensor#kind -> "contract"; sensor#prompt -> readPrompt;
    sensor#text -> "Service Agreement between Beta GmbH (the Provider) and Acme AG (the Customer). This agreement is governed by Swiss law."].

?- d1[sensor#accepted(?I) -> ?V].
?- d1[sensor#rejected(?I) -> ?Why].
?- d1[sensor#notRead -> ?E].
\end{lstlisting}
The recorded run gives the same results as the program above. Write the rules out when a
case needs its own acceptance rule, and use the library when it does not.
A confidence reported by the model is recorded as a number and never read as truth.
Where the serving stack supports it, decode every reply under a JSON schema derived
from the closed set of allowed values.

\item[Known uses] Assignment review with quote-checked extraction; campervan search
(one classification per listing, from a fixed list or \emph{unknown}); the judge in
dialogue flows (one criterion about one answer); memory claims with quotes in
FrameX-Agents; model measurements per object in the Q-calculus example.
\knownlit{sensor}

\item[Related patterns] The judge of Rules Steer the Model (\cref{pat:steer}) is a
sensor for one answer. Critique and Repair (\cref{pat:repair}) handles the replies
that fail a check. A Sensor whose readings are aggregated by rules over many objects (lists, counts) is
where the hybrid's accuracy advantage was largest.
\item[Variants] \emph{Several Readers, One Rule.} The same question goes to several
models or samples, and rules adjudicate: agreeing readings are accepted, a majority is
accepted with its count recorded, and a disagreement becomes \emph{unclear} and is handed
on (\cref{pat:competence}); it is never settled by a coin toss. Voting and parallel
sampling are published agent and workflow patterns
\parencite{liu2025agentpatterns,schluntz2024agents}, and self-consistency beat
multi-agent debate at the same number of answers \parencite{huang2024cannot}. Asking a
judge twice with the answers in swapped order and requiring agreement is the same variant
applied to position bias \parencite{zheng2023judge}.

\item[Evidence] In a portfolio of 100 agreements (376 documents), a model alone
answered single decisions and four-step chains correctly but produced exact
exhaustive lists in 0 of 6 questions at low effort; with the model as sensor and the
lists derived by rules, 10 of 10 questions were correct, with a completeness check
leaving no silently wrong answer. The extraction was the expensive step: 43.6 cents
once against 1.2 cents per question for the model alone. With an 8-billion-parameter
open model, the strict quote rule refused correct values and made extraction worse
than the model alone (45\,\% against 60\,\%): the acceptance rule must fit the
reader's ability to quote.
\end{patternform}

\FloatBarrier
\subsection{Pattern 2: Model as Author}
\label{pat:author}

\begin{patternform}
\item[Intent] Let the model write formal material (rules, decision tables, schemas,
tests, dialogue flows, thresholds) and check every draft offline before a person
adopts it. The rules never run what the model wrote unchecked.

\item[Also known as] Rule drafting, policy-to-rules, specification-driven authoring,
knowledge acquisition with a model.

\item[Motivation] A policy sentence (``claims above 500 francs need a receipt'') must
become a rule, and an amended regulation must become a changed decision table. A model
drafts the rule in seconds, and the expensive part is checking it. The pattern moves
the cost to a check that is complete for the faults that matter: the
draft must parse, may only use allowed heads, and must pass the regression cases and
expectations, and the check reports which decisions the draft would change.

\item[Applicability] Use it when
\begin{itemize}
\item the target notation has a complete check for the faults that matter
(syntax, allowed vocabulary, structural anomalies of a decision table, regression
cases);
\item regression cases or expectations exist or can be written before the draft;
\item a person is available to adopt what passes.
\end{itemize}
Where no complete check exists (semantic faults in a large rule base), the draft can
only open a review (see Model as Critic, \cref{pat:critic}).

\item[Structure]\leavevmode
\begin{center}
\begin{tikzpicture}[bx]
\node[bxinst] (b1) at (0,0) {data:\\text};
\node[bxnote, below=0.6mm of b1] {policy};
\node[bxproc, right=of b1] (b2) {infer};
\node[bxmodel, above=of b2] (b2m) {model:\\gen};
\node[bxnote, above=0.3mm of b2m] {language model};
\draw[bxflow] (b2m)--(b2);
\draw[bxflow] (b1)--(b2);
\node[bxmodel, right=of b2] (b3) {model:\\sem};
\node[bxnote, below=0.6mm of b3] {draft rule};
\draw[bxflow] (b2)--(b3);
\node[bxproc, right=of b3] (b4) {infer:\\deduce};
\node[bxmodel, above=of b4] (b4m) {model:\\sem};
\node[bxnote, above=0.3mm of b4m] {checker, cases};
\draw[bxflow] (b4m)--(b4);
\draw[bxflow] (b3)--(b4);
\node[bxinst, right=of b4] (b5) {symbol};
\node[bxnote, below=0.6mm of b5] {verdict};
\draw[bxflow] (b4)--(b5);
\node[bxactor, right=of b5] (b6) {};
\node[bxnote, below=0.6mm of b6] {expert};
\draw[bxflow] (b5)--(b6);
\node[bxproc, right=of b6] (b7) {generate:\\engineer};
\draw[bxflow] (b6)--(b7);
\node[bxmodel, right=of b7] (b8) {model:\\sem};
\node[bxnote, below=0.6mm of b8] {adopted rule};
\draw[bxflow] (b7)--(b8);
\draw[bxreq] (b4.south) .. controls +(0,-10mm) and +(0,-10mm) .. node[pos=0.5, below, inner sep=1pt, font=\tiny]{request} (b2.south);
\end{tikzpicture}
\end{center}

\item[Participants] \emph{Model}: drafts one artefact from a policy text, a change
request or failing cases. \emph{Checker}: parses, restricts heads, evaluates the
candidate in a copy of the session against expectations and regression cases.
\emph{Rules}: request the check and derive \emph{adoptable}, or a failure with the
checker's message that drives the repair.
\emph{Expert}: adopts, edits or refuses.

\item[Collaborations] The model's draft is stored as a receipt; a rule requests its
check; the checker's receipt states regressions and changed decisions; a rule derives
either a recommendation for the expert or a precise repair message for the model
(\cref{pat:repair}).

\item[Consequences] Writing becomes cheap and checking becomes the bottleneck; the
quality of the engine's error messages becomes a design parameter. A draft outside the
target language is refused at its first cell or token. What the organisation keeps is the
adopted artefact, and the model can be exchanged. The pattern does not detect a draft that is
well-formed, passes every case and is still wrong where no case covers it.

\item[Implementation] Give the model a template to fill (a rule skeleton, the cell
language of a decision table). Check in a copy of the session, and let a person adopt.
Phrase engine messages for a reader who repairs: ``separate the slots of a frame with
`;', not `,''' led to more repairs than ``expected `;'\,''.

The first request calls the model, the second the checker \texttt{\_checkCandidate}. The
operator fixes in an adapter file what the checker compares against: the vocabulary
file, the hidden cases (\texttt{expect true: c3:NeedsReceipt.}) and the only head a
draft may have%
. Its receipt carries \texttt{status}
(\texttt{checked}; \texttt{rejected} before evaluation, for example for a syntax error or
a head outside the allowed set; or \texttt{error}), and after evaluation
\texttt{regressions} (a case that passed before fails now), \texttt{failedAfter} (cases
failing after) and one line of \texttt{feedback}. The second half of the program is the
repair of Critique and Repair:
\begin{lstlisting}[style=fxprog,basicstyle=\footnotesize\ttfamily,language=framex]
// Pattern 2, Model as Author, with Pattern 6, Critique and Repair: the model drafts a
// rule, the checker measures it against hidden cases, and the checker's message
// drives a bounded repair. Nothing is adopted here; a person adopts.
world open.

p1:Policy[pending -> true; maxAttempts -> 3;
    text -> "A claim above 500 francs needs a receipt.";
    vocabulary -> "Class Claim with slot amount (a number). Write rules whose head is ?C:NeedsReceipt and nothing else. A rule is HEAD <- BODY. Slots: ?C:Claim[amount -> ?A]. Comparisons: > >= < <=."].

draftPrompt:Prompt[text -> "Write the FrameX rule for the policy sentence, using only the vocabulary. Answer as JSON {\"rule\": ...}."].
repairPrompt:Prompt[text -> "Your FrameX rule for the policy sentence failed the check. previousRule is your rule, problem is the checker's message. Write the corrected rule, using only the vocabulary. Answer as JSON {\"rule\": ...}."].

// Attempt 1: the model drafts.
external draft(?P, 1) using _llm with draftPrompt[policy -> ?T; vocabulary -> ?V;
        tool#order -> 2]
  <- ?P:Policy[pending -> true; text -> ?T; vocabulary -> ?V].

?P[rule(?N) -> ?T]
  <- ?R[tool#name -> "_llm"; tool#arg1 -> ?P; tool#arg2 -> ?N; result("rule") -> ?T].

// Every draft is checked; the cases stay hidden from the model.
external check(?P, ?N) using _checkCandidate[candidate -> ?T; tool#order -> 1]
  <- ?P:Policy[rule(?N) -> ?T].

?P[passed(?N) -> true]
  <- ?R[tool#name -> "_checkCandidate"; tool#arg1 -> ?P; tool#arg2 -> ?N;
        result("status") -> "checked"; result("regressions") -> false;
        result("failedAfter") -> 0].

?P[adoptable -> ?T] <- ?P[passed(?N) -> true; rule(?N) -> ?T].

// A draft fails in one of three ways; each failure keeps the checker's message.
?P[failed(?N) -> ?F]
  <- ?R[tool#name -> "_checkCandidate"; tool#arg1 -> ?P; tool#arg2 -> ?N;
        result("status") -> ?St; result("feedback") -> ?F] AND ?St != "checked".

?P[failed(?N) -> ?F]
  <- ?R[tool#name -> "_checkCandidate"; tool#arg1 -> ?P; tool#arg2 -> ?N;
        result("regressions") -> true; result("feedback") -> ?F].

?P[failed(?N) -> ?F]
  <- ?R[tool#name -> "_checkCandidate"; tool#arg1 -> ?P; tool#arg2 -> ?N;
        result("failedAfter") -> ?K; result("feedback") -> ?F] AND ?K > 0.

// Pattern 6: ask again with the rule and the problem in the request, as a new
// request with the next number. Stop at the bound or when the message repeats.
?P[newProblem(1) -> true] <- ?P[failed(1) -> ?F].

?P[newProblem(?N) -> true]
  <- ?P[failed(?N) -> ?F] AND ?K := ?N - 1 AND ?P[failed(?K) -> ?G] AND ?F != ?G.

external draft(?P, ?M) using _llm with repairPrompt[policy -> ?T; vocabulary -> ?V;
        previousRule -> ?Old; problem -> ?F; tool#order -> 2]
  <- ?P:Policy[text -> ?T; vocabulary -> ?V; maxAttempts -> ?Max;
        rule(?N) -> ?Old; failed(?N) -> ?F; newProblem(?N) -> true]
     AND ?N < ?Max AND ?M := ?N + 1.

?P[stalled -> ?N]
  <- ?P[failed(?N) -> ?F] AND ?K := ?N - 1 AND ?P[failed(?K) -> ?F].

?P[gaveUp -> ?F] <- ?P:Policy[maxAttempts -> ?N; failed(?N) -> ?F].

?- p1[rule(?N) -> ?T].
?- p1[failed(?N) -> ?F].
?- p1[adoptable -> ?T].
?- p1[stalled -> ?N].
\end{lstlisting}
\begin{figure}[p]
\centering
\includegraphics[width=\linewidth,height=0.82\textheight,keepaspectratio]{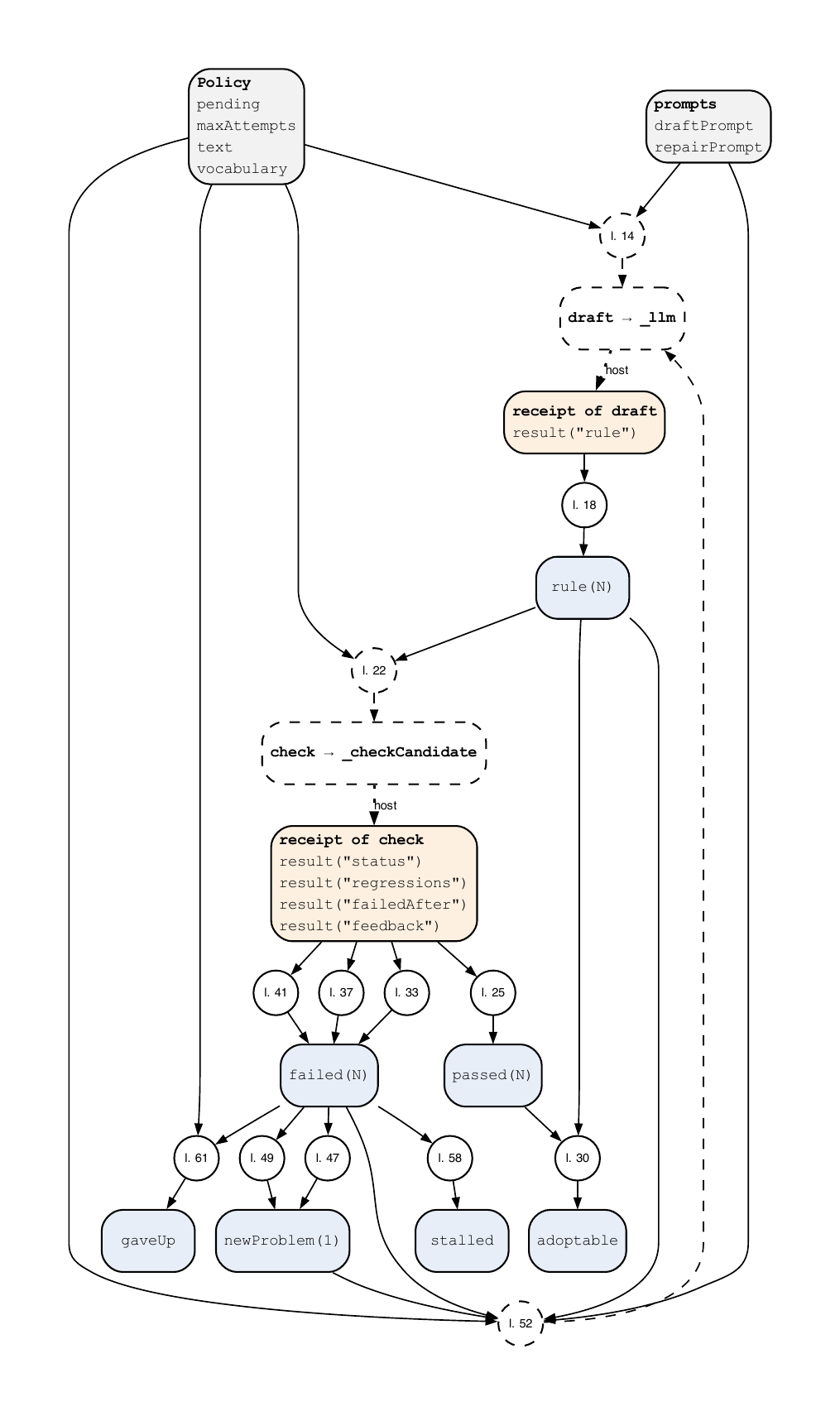}
\caption{Rule graph of \texttt{p02-author.fx}, generated from the program. A circle is one rule; ``l.~23'' means that the rule starts in line 23 of the listing. Legend in \cref{tab:rglegend}.}
\label{fig:rg-author}
\end{figure}
\Cref{fig:rg-author} draws the program as a rule graph.
\texttt{adoptable} is a recommendation for a person; nothing in the program adopts a rule.

\item[Library form] With the library \texttt{author.fx} (\cref{sec:libraries}) loaded first
and the adapter of the checker named to the host, the program states the policy, its
vocabulary, the two prompts and the bound. The library drafts, checks, repairs and stops;
the repair of Critique and Repair is part of it:
\begin{lstlisting}[style=fxprog,basicstyle=\footnotesize\ttfamily,language=framex]
// Patterns 2 and 6 in library form: load author.fx first; the host needs ../author_adapter.py.
world open.

draftPrompt:Prompt[text -> "Write the FrameX rule for the policy sentence, using only the vocabulary. Answer as JSON {\"rule\": ...}."].
repairPrompt:Prompt[text -> "Your FrameX rule for the policy sentence failed the check. previousRule is your rule, problem is the checker's message. Write the corrected rule, using only the vocabulary. Answer as JSON {\"rule\": ...}."].

p1:author#Policy[author#maxAttempts -> 3;
    author#draftPrompt -> draftPrompt; author#repairPrompt -> repairPrompt;
    author#text -> "A claim above 500 francs needs a receipt.";
    author#vocabulary -> "Class Claim with slot amount (a number). Write rules whose head is ?C:NeedsReceipt and nothing else. A rule is HEAD <- BODY. Slots: ?C:Claim[amount -> ?A]. Comparisons: > >= < <=."].

?- p1[author#failed(?N) -> ?F].
?- p1[author#adoptable -> ?T].
\end{lstlisting}

\item[Known uses] Rule drafting from one policy sentence with six hidden cases;
decision-table drafts and change requests in the DMN round trip, checked against every
case and reviewed; the EU Regulation 261/2004 case, where a draft outside the cell
language was refused and a change request fixed six of six failing cases, broke none and
passed review; mapping suggestions between vocabularies kept as recorded proposals.
\knownlit{author}

\item[Related patterns] Critique and Repair (\cref{pat:repair}) closes the loop with
the checker's message. Division by Competence (\cref{pat:competence}) supplies drafts
from recurring model decisions. Model as Critic (\cref{pat:critic}) applies where the
check is incomplete.

\item[Evidence] In five live runs, a model drafted a rule from one policy sentence;
every run ended with a rule that passed all six hidden cases, three at the first
attempt, two after one repair of the same syntax slip. From the same faulty drafts, a
repair succeeded at the first request in 5 of 10 runs with the message ``expected
`;'\,'' and in 10 of 10 with a message naming the fix. Across 36 drafts for 12
policies, messages that name the fix raised repairs within two attempts from 8 of 15 to
14 of 15. In a checking study, static checks over rules found every seeded structural
fault at every size at no model cost, while model reviews of code and prompts found an
unreachable rule in only 2 or 3 of 10 artefacts at 64 rules.
\end{patternform}

\FloatBarrier
\subsection{Pattern 3: Model as Translator}
\label{pat:translator}

\begin{patternform}
\item[Intent] Put the model between a person and the rules, in both directions: the
person's words into a request the rules can answer, and the rules' answer and proof into
words for that person.

\item[Also known as] Natural-language interface, verbalisation, explanation, the model
calls the rules as a tool.

\item[Motivation] A clerk asks ``can we pay this invoice?'' and expects an answer in
words, with the reason. The rules hold the answer and its proof but speak in queries and
proof trees. The model speaks the clerk's language but should not decide. The pattern
gives the model the conversation and the rules the decision: the model calls the rules
as a tool, receives the verdict, the decisive quotes and the missing evidence, and
phrases them; the host checks that the reply uses the rules' decision word.

\item[Applicability] Use it when people ask in their own words, when answers must
come with reasons a non-expert can read, or when a model-based assistant needs a
verified source of decisions. Prefer fixed templates over a model where the audience
and the wording are known (sentences generated from a decision table need no model).

\item[Structure]\leavevmode
\begin{center}
\begin{tikzpicture}[bx]
\node[bxinst] (c1) at (0,0) {data:\\text};
\node[bxnote, below=0.6mm of c1] {question};
\node[bxproc, right=of c1] (c2) {infer};
\node[bxmodel, above=of c2] (c2m) {model:\\gen};
\node[bxnote, above=0.3mm of c2m] {language model};
\draw[bxflow] (c2m)--(c2);
\draw[bxflow] (c1)--(c2);
\node[bxinst, right=of c2] (c3) {symbol};
\node[bxnote, below=0.6mm of c3] {request};
\draw[bxflow] (c2)--(c3);
\node[bxproc, right=of c3] (c4) {infer:\\deduce};
\node[bxmodel, above=of c4] (c4m) {model:\\sem};
\node[bxnote, above=0.3mm of c4m] {rules};
\draw[bxflow] (c4m)--(c4);
\draw[bxflow] (c3)--(c4);
\node[bxinst, right=of c4] (c5) {symbol};
\node[bxnote, below=0.6mm of c5] {decision, proof};
\draw[bxflow] (c4)--(c5);
\node[bxproc, right=of c5] (c6) {infer};
\node[bxmodel, above=of c6] (c6m) {model:\\gen};
\node[bxnote, above=0.3mm of c6m] {language model};
\draw[bxflow] (c6m)--(c6);
\draw[bxflow] (c5)--(c6);
\node[bxinst, right=of c6] (c7) {data:\\text};
\node[bxnote, below=0.6mm of c7] {reply};
\draw[bxflow] (c6)--(c7);
\node[bxproc, right=of c7] (c8) {trans-\\form};
\node[bxnote, below=0.6mm of c8] {decision word};
\draw[bxflow] (c7)--(c8);
\node[bxinst, right=of c8] (c9) {symbol};
\node[bxnote, below=0.6mm of c9] {reply accepted};
\draw[bxflow] (c8)--(c9);
\draw[bxreq] (c4.south) .. controls +(0,-10mm) and +(0,-10mm) .. node[pos=0.5, below, inner sep=1pt, font=\tiny]{request} (c6.south);
\end{tikzpicture}
\end{center}

\item[Participants] \emph{Person}; \emph{Model} (translates); \emph{Rules} (answer,
prove, name what is missing: \texttt{query}, \texttt{explain}, \texttt{why\_not},
\texttt{preview}, \texttt{evidence\_needs}); \emph{Host} (checks the reply against the
answer).

\item[Collaborations] The model turns the question into a call of a read-only operation
from a closed set, or, where free queries are admitted, into a candidate query that the
rules validate like any draft (\cref{pat:author}) before it runs; the rules answer with a
proof; the model phrases the answer; a check binds the phrasing to
the answer (decision word, cited proof nodes). In the other direction the rules can be
offered read-only to any model client as a server.

\item[Consequences] The person gets an answer in words with a proof behind it. A
mistranslated question gives a correct answer to the wrong question, so a
back-translation shown to the person is the main safeguard. Prose that adds facts the proof lacks must be rejected, which requires the phrasing to
cite the proof. The two forms of request differ in what can be checked. A call of a closed
operation set keeps invariant~1 without qualification and lets a rule refuse a request
with a reason that names the field. A free query is more expressive, but a check can only
establish its form, its vocabulary and that it is read-only; it inherits the consequences
of Model as Author, and a query that validates can still ask the wrong question.

\item[Implementation] Prefer a closed set of read-only operations whose arguments are
identifiers; the host builds the executable query from the call. Admit a free query only
through \texttt{validate} (syntax, allowed classes and slots, no assertions) and record it
as a candidate with its receipt. Keep the model's view small: the verdict, the decisive
quotes and the missing evidence, without the documents. Offer the rules as read-only tools with a hash
of the files that answered. Bind every sentence of an explanation to a proof node, or
use templates.

In rules, the model's first call turns the question into a request, the request runs
only when every field names a member of a closed list, a refusal names the field, and the
rules derive the decision. The second call names the decision word as \texttt{mustSay},
which the model does not see; the host records whether the reply contains it
(\texttt{saysDecision}), and only such a reply is accepted. The test is a substring test
that ignores case: ``not admitted'' contains ``admitted''. Where one decision word
contains another, ask for the decision word in a field of its own as well and compare it
exactly. The program answers
admission questions; the operations \emph{requirements} and \emph{missing} are recognised
and refused with a reason:
\begin{lstlisting}[style=fxprog,basicstyle=\footnotesize\ttfamily,language=framex]
// Pattern 3, Model as Translator: a question becomes a checked request, the rules'
// decision becomes a sentence that must state it.
world open.
closed class Operation. closed class Programme. closed class Applicant.
closed slot admitted.

"admission":Operation. "requirements":Operation. "missing":Operation.
"mbi":Programme. "mgt":Programme.
"nora":Applicant. "tomas":Applicant.
"nora"[admitted -> "mbi"].
q1:Question[text -> "Hi, I am Nora. Am I admitted to the Master in Business Innovation?"].

translatePrompt:Prompt[text -> "Map the question to one call. Answer as JSON {\"op\": admission, requirements or missing, \"programme\": mbi or mgt, \"applicant\": the lowercase first name}."].
phrasePrompt:Prompt[text -> "Tell the person the decision in one friendly sentence that uses the decision word. Answer as JSON {\"text\": ...}."].

// From words to a request.
external translate(?Q) using _llm with translatePrompt[question -> ?T]
  <- ?Q:Question[text -> ?T].

?Q:Request[op -> ?O; programme -> ?P; applicant -> ?A]
  <- ?R[tool#name -> "_llm"; tool#arg1 -> ?Q; result("op") -> ?O;
        result("programme") -> ?P; result("applicant") -> ?A].

// The request runs only if every field names something the rules know.
?Q[runs -> true] <- ?Q:Request[op -> ?O; programme -> ?P; applicant -> ?A]
    AND ?O:Operation AND ?P:Programme AND ?A:Applicant.

?Q[refused -> "unknown operation"] <- ?Q:Request[op -> ?O] AND NOT ?O:Operation.
?Q[refused -> "unknown programme"] <- ?Q:Request[programme -> ?P] AND NOT ?P:Programme.
?Q[refused -> "unknown applicant"] <- ?Q:Request[applicant -> ?A] AND NOT ?A:Applicant.
?Q[refused -> "this program answers admission questions only"]
  <- ?Q:Request[runs -> true; op -> ?O] AND ?O != "admission".

// The rules decide.
?Q[decision -> "admitted"]
  <- ?Q:Request[runs -> true; op -> "admission"; programme -> ?P; applicant -> ?A]
     AND ?A[admitted -> ?P].

?Q[decision -> "not admitted"]
  <- ?Q:Request[runs -> true; op -> "admission"; programme -> ?P; applicant -> ?A]
     AND NOT ?A[admitted -> ?P].

// From the decision to words; the reply counts only if it says the decision word.
external phrase(?Q) using _llm with phrasePrompt[decision -> ?D; mustSay -> ?D]
  <- ?Q:Request[decision -> ?D].

?Q[reply -> ?Text]
  <- ?Q:Request AND ?X[tool#name -> "_llm"; tool#arg1 -> ?Q;
        result("saysDecision") -> true; result("text") -> ?Text].

?Q[replyRefused -> "the reply does not state the decision"]
  <- ?Q:Request AND ?X[tool#name -> "_llm"; tool#arg1 -> ?Q;
        result("saysDecision") -> false].

?- ?Q:Request[decision -> ?D].
?- ?Q[reply -> ?Text].
?- ?Q[refused -> ?Why].
\end{lstlisting}
\Cref{fig:rg-translator} draws the program as a rule graph.

\item[Library form] With \texttt{translator.fx} (\cref{sec:libraries}) the program declares
the closed list of each answer field and keeps its decision rules. The library reads the
fields, refuses a request with the name of the field that lies outside its list, and
phrases the decision under \texttt{mustSay}. Decision rules read only valid fields
(\texttt{tr\#valid}). A rule that asks whether \emph{all} fields are valid negates over the
receipts, and the reply request depends on the decision; FrameX refuses that as a negative
cycle, so the check is spelled out per field:
\begin{lstlisting}[style=fxprog,basicstyle=\footnotesize\ttfamily,language=framex]
// Pattern 3 in library form: load translator.fx first.
world open.
closed slot admitted.

translatePrompt:Prompt[text -> "Map the question to one call. Answer as JSON {\"op\": admission, requirements or missing, \"programme\": mbi or mgt, \"applicant\": the lowercase first name}."].
phrasePrompt:Prompt[text -> "Tell the person the decision in one friendly sentence that uses the decision word. Answer as JSON {\"text\": ...}."].
tr#phrasing[tr#prompt -> phrasePrompt].

"op"[tr#allows -> "admission"].
"programme"[tr#allows -> "mbi"; tr#allows -> "mgt"].
"applicant"[tr#allows -> "nora"; tr#allows -> "tomas"].
"nora"[admitted -> "mbi"].

q1:tr#Question[tr#prompt -> translatePrompt;
    tr#text -> "Hi, I am Nora. Am I admitted to the Master in Business Innovation?"].

// The rules decide; this part stays with the program.
?Q[tr#decision -> "admitted"]
  <- ?Q:tr#Question[tr#valid("op") -> "admission"; tr#valid("programme") -> ?P;
        tr#valid("applicant") -> ?A] AND ?A[admitted -> ?P].
?Q[tr#decision -> "not admitted"]
  <- ?Q:tr#Question[tr#valid("op") -> "admission"; tr#valid("programme") -> ?P;
        tr#valid("applicant") -> ?A] AND NOT ?A[admitted -> ?P].

?- ?Q[tr#decision -> ?D].
?- ?Q[tr#reply -> ?Text].
?- ?Q[tr#refused -> ?F].
\end{lstlisting}

\item[Known uses] The assignment-review chat model calls a reviewer tool that runs
the rules; FrameX as a read-only server for model clients; a research agent that keeps
the model's explanation beside a checked conclusion over a knowledge graph; sentences
generated from decision-table rows without a model.
\knownlit{translator}

\item[Related patterns] Model as Sensor (\cref{pat:sensor}) translates
documents where this pattern translates questions. Rules Steer the Model (\cref{pat:steer}) adds a
flow to the conversation.

\item[Variants] \emph{Read-Back Check.} An independent reader that sees only the written
reply reads its facts back, and rules compare them with the facts the reply was meant to
state: the decision, the amounts, the dates. It is the translator's check applied to the
output side, and it catches a changed value that a decision word alone does not. The
reader must differ from the writer (another model or another prompt), or it repeats the
writer's misreading. Reported in several Hack Apertus submissions (Sturmglocke, Lightpath,
SONA with entailment in both directions); not built in FrameX. In one of
them, a 70B second reader cut instructions with a changed meaning from 4.6\,\% to
0.2\,\% of 2,078 (result file committed in the submission).

\item[Evidence] On a clean five-document case, a model alone answered the
assignment-review question correctly in 12 of 12 runs at lower cost and time than the
hybrid (0.28 against 0.72 cents per case). On easy cases the translator earns its cost
through the proof, the stated missing evidence and the fixed decision word.

\begin{figure}[hp]
\centering
\includegraphics[width=\linewidth,height=0.82\textheight,keepaspectratio]{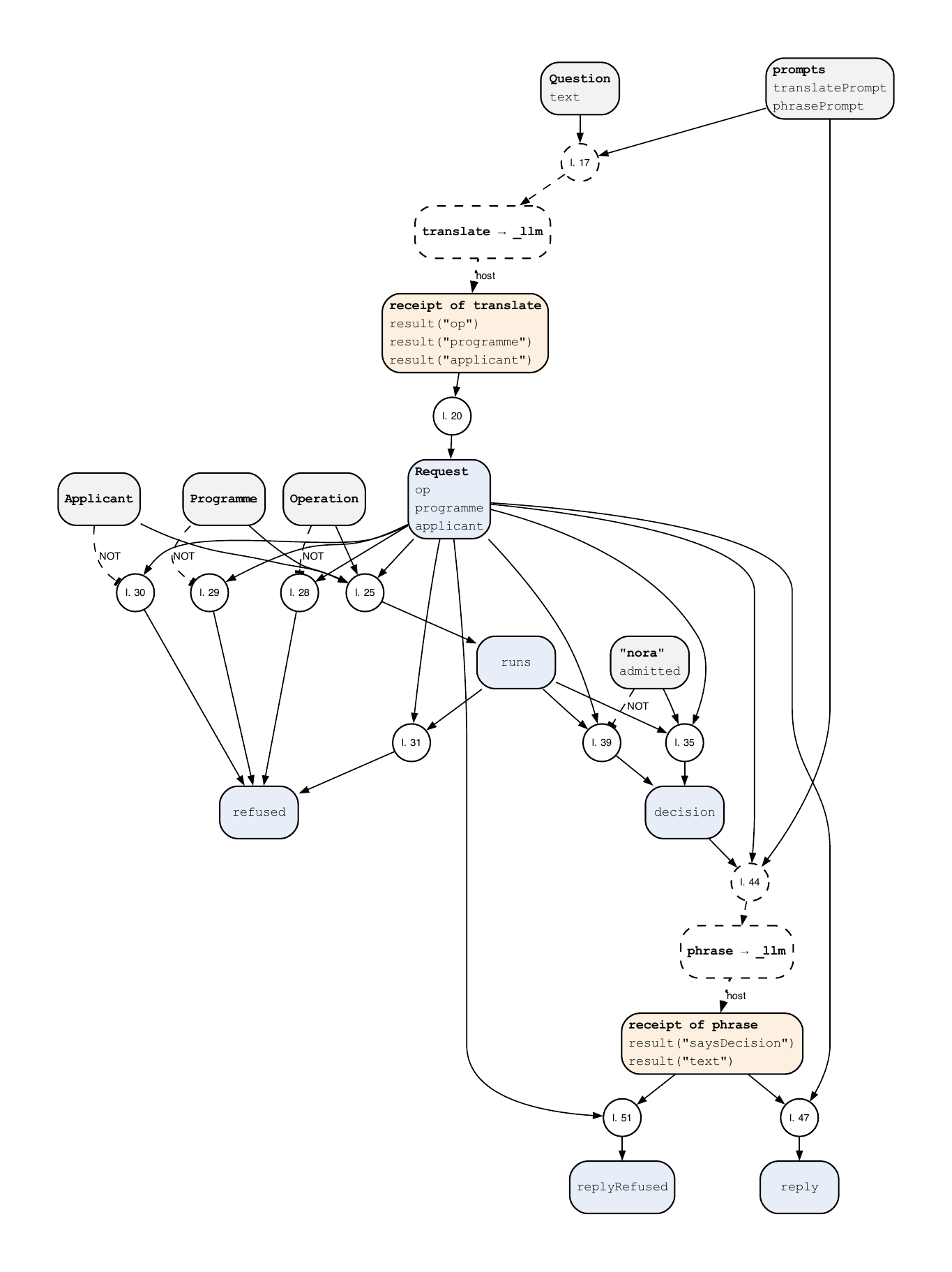}
\caption{Rule graph of \texttt{p03-translator.fx}, generated from the program. A circle is one rule; ``l.~23'' means that the rule starts in line 23 of the listing. Legend in \cref{tab:rglegend}.}
\label{fig:rg-translator}
\end{figure}

\end{patternform}

\FloatBarrier
\subsection{Pattern 12: Model Selects, Rules Write}
\label{pat:select}

\begin{patternform}
\item[Intent] Let the model choose and phrase, but never state content: it returns
identifiers of passages that the host numbered, and words around values it never sees.
Every passage and every value that reaches a person comes from facts, word for word.

\item[Also known as] Model points, code speaks; model phrases, code fills; extractive
answer by identifier; slot filling; template generation with a model.

\item[Motivation] An applicant asks which documents the Master in Business Innovation
requires and what the application costs. The answer must quote the regulation exactly
and state the fee exactly. A model that writes the answer can paraphrase a requirement
into a different one or round a fee, and a check afterwards (a quote that must occur, a
decision word that must appear) catches only what it looks for. The pattern moves the
check before the writing: the model sees the passages under identifiers and answers with
identifiers, and it phrases a sentence around a gap that the rules fill. A requirement
the model invents has no identifier, and a fee it never saw it cannot change.

\item[Applicability] Use it when
\begin{itemize}
\item the content of an answer exists as facts or source text (regulations, a catalogue,
computed amounts, dates) and only its selection or wording needs a model;
\item a wrong word in the content is costly, so a reply must be correct by construction
rather than by check;
\item the source can be split into passages that a person reads on their own.
\end{itemize}
Do not use it when the answer must combine or summarise what no single passage says;
then use Model as Translator (\cref{pat:translator}) with its check of the reply.

\item[Structure]\leavevmode
\begin{center}
\begin{tikzpicture}[bx]
\node[bxinst] (s1) at (0,0) {symbol};
\node[bxnote, below=0.6mm of s1] {numbered passages};
\node[bxproc, right=of s1] (s2) {infer};
\node[bxmodel, above=of s2] (s2m) {model:\\gen};
\node[bxnote, above=0.3mm of s2m] {language model};
\draw[bxflow] (s2m)--(s2);
\draw[bxflow] (s1)--(s2);
\node[bxinst, right=of s2] (s3) {data:\\text};
\node[bxnote, below=0.6mm of s3] {identifiers, words};
\draw[bxflow] (s2)--(s3);
\node[bxproc, right=of s3] (s4) {trans-\\form};
\node[bxnote, below=0.6mm of s4] {receipt};
\draw[bxflow] (s3)--(s4);
\node[bxinst, right=of s4] (s5) {symbol};
\node[bxnote, below=0.6mm of s5] {proposed};
\draw[bxflow] (s4)--(s5);
\node[bxproc, right=of s5] (s6) {infer:\\deduce};
\node[bxmodel, above=of s6] (s6m) {model:\\sem};
\node[bxnote, above=0.3mm of s6m] {rules, source, values};
\draw[bxflow] (s6m)--(s6);
\draw[bxflow] (s5)--(s6);
\node[bxinst, right=of s6] (s7) {data:\\text};
\node[bxnote, below=0.6mm of s7] {passages, value verbatim};
\draw[bxflow] (s6)--(s7);
\draw[bxreq] (s6.south) .. controls +(0,-15mm) and +(0,-15mm) .. node[pos=0.5, below, inner sep=1pt, font=\tiny]{request} (s2.south);
\end{tikzpicture}
\end{center}

\item[Participants] \emph{Host}: splits the source into passages and numbers them,
calls the model, stores the reply as a receipt. \emph{Model}: chooses identifiers from
the list it was given, and phrases words around a gap. \emph{Rules}: request the
selection and the phrasing, accept an identifier only if it names a passage of the
source, and write the answer from the passages and the values. \emph{Person}: reads
passages and values as they stand in the source.

\item[Collaborations] A rule requests a selection for a question with the numbered
passages; the receipt holds identifiers; a rule shows the passage of each identifier
that belongs to the source and refuses any other with its reason. A second request asks
for words before and after a value without the value; a rule inserts the value between
them, unless the words state a number of their own.

\item[Consequences] No sentence of the model reaches the person as content: an invented
passage has no identifier and is refused, a fee is never retyped. What the model can
still get wrong is the choice: a passage that does not answer the question, or one that
is missing. That is a fault of relevance, not of truth, and a person sees it in the
source's own words. Answers are as fluent as the source; text that combines several
passages is out of reach. The cost is one call per selection and per phrasing.

\item[Implementation] Number the passages in the host, where the source is split; the
model sees \texttt{a1: \ldots}, the rules know \texttt{"a1"} as a \texttt{Passage} with
its source and text. Declare the class closed, so that an identifier outside it is
refused rather than left \emph{unknown}. For values, the request carries what the value
is about but never the value; the rules join the words and the value with
\texttt{concat}, and refuse words that contain a digit, so that a number the model made
up cannot sit beside the right one. The program answers one question from three
passages and states one fee:
\begin{lstlisting}[style=fxprog,basicstyle=\footnotesize\ttfamily,language=framex]
// Pattern 12, Model Selects, Rules Write: the model returns identifiers and phrasing;
// every passage and every value shown to the person comes from facts.
world open.
closed class Passage. closed class Digit. closed slot hasDigit.

// The source, split into passages by the host, and the same passages numbered for the model.
"a1":Passage[source -> reg; text -> "Applicants hold a recognised Bachelor degree."].
"a2":Passage[source -> reg; text -> "Applicants prove English at level C1."].
"a3":Passage[source -> reg; text -> "The programme starts in the autumn semester."].
reg[numbered -> "a1: Applicants hold a recognised Bachelor degree. a2: Applicants prove English at level C1. a3: The programme starts in the autumn semester."].

// A question about the source, and a value an answer must state.
q1:Question[source -> reg; text -> "Which documents do I need for the MBI?"].
fee1:Amount[question -> q1; value -> "CHF 250"; about -> "the application fee"].

selectPrompt:Prompt[text -> "Choose the passages that answer the question. Answer as JSON {\"rows\": [{\"passage\": an identifier from the list}]}."].
phrasePrompt:Prompt[text -> "Write one sentence that tells the applicant the amount named in about; the system inserts the amount. Answer as JSON {\"before\": the text before the amount, \"after\": the text after it}."].

// Request 1: the model sees the numbered passages and returns identifiers only.
external select(?Q) using _llm with selectPrompt[question -> ?T; passages -> ?N]
  <- ?Q:Question[source -> ?S; text -> ?T] AND ?S[numbered -> ?N].

// Accept: an identifier shows its own passage, word for word.
?Q[shows(?I) -> ?Text]
  <- ?R[tool#name -> "_llm"; tool#arg1 -> ?Q; row(?I, "passage") -> ?P]
     AND ?Q:Question[source -> ?S] AND ?P:Passage[source -> ?S; text -> ?Text].

?Q[refused(?I) -> "not a passage of this source"]
  <- ?R[tool#name -> "_llm"; tool#arg1 -> ?Q; row(?I, "passage") -> ?P]
     AND ?Q:Question AND NOT ?P:Passage.

// Request 2: the model phrases around the value; the request does not carry the value.
external phrase(?V) using _llm with phrasePrompt[about -> ?A]
  <- ?V:Amount[about -> ?A].

"0":Digit. "1":Digit. "2":Digit. "3":Digit. "4":Digit.
"5":Digit. "6":Digit. "7":Digit. "8":Digit. "9":Digit.

?R[hasDigit -> true]
  <- ?R[tool#name -> "_llm"; tool#arg1 -> ?V; result("before") -> ?B]
     AND ?V:Amount AND ?D:Digit
     AND ?N := occurrences(?B, ?D) AND ?N > 0.
?R[hasDigit -> true]
  <- ?R[tool#name -> "_llm"; tool#arg1 -> ?V; result("after") -> ?F]
     AND ?V:Amount AND ?D:Digit
     AND ?N := occurrences(?F, ?D) AND ?N > 0.

// Write: the rules put the value between the model's words.
?Q[answer -> ?Text]
  <- ?R[tool#name -> "_llm"; tool#arg1 -> ?V; result("before") -> ?B; result("after") -> ?F]
     AND ?V:Amount[question -> ?Q; value -> ?X] AND NOT ?R[hasDigit -> true]
     AND ?Text := concat(concat(?B, ?X), ?F).

?V[rejected -> "the phrasing states a number of its own"]
  <- ?R[tool#name -> "_llm"; tool#arg1 -> ?V; hasDigit -> true] AND ?V:Amount.

// A failed call shows nothing and keeps the host's reason.
?X[notAnswered -> ?E]
  <- ?R[tool#name -> "_llm"; tool#arg1 -> ?X; tool#status -> "failed"; error -> ?E].

?- q1[shows(?I) -> ?Text].
?- q1[refused(?I) -> ?Why].
?- q1[answer -> ?A].
?- fee1[rejected -> ?Why].
\end{lstlisting}
\Cref{fig:rg-select} draws the program as a rule graph. The two requests have their own
subjects, the question and the amount, so that each receipt is read only by its own
rules. Where the serving stack decodes under a JSON schema, an enumeration of the
passage identifiers makes an identifier outside the list impossible to produce; the
rule of line~28 stays as the check that does not depend on the stack. There is no
library form yet.

\item[Known uses] In FrameX, the program above; decision-table rows become sentences from
templates without any model, the limit case of this pattern. In the submissions to the Hack Apertus hackathon (October 2026), as
described in the jury assessments, models chose passage or line identifiers that the
application then quoted (QuietLetter, Beneficia, Comfort, StimmFolgen, Fedlex, ClearDesk,
a meditation retrieval system) or wrote with placeholders that code filled with numbers
and terms (Apertus QA, Bon·Appétit, Sturmglocke, Velum).
\knownlit{select}

\item[Related patterns] Model as Sensor (\cref{pat:sensor}) reads a source and checks
quotes afterwards; this pattern makes the quote the only possible answer. Model as
Translator (\cref{pat:translator}) checks a written reply for the decision word; here
the reply cannot contain the decision. Deriving the options first under Model Proposes,
Rules Dispose (\cref{pat:dispose}) is the same move for actions.

\item[Evidence] The program is checked against recorded receipts only%
; FrameX has no measurement of its own. Figures from the hackathon are taken from result files committed in the submissions' repositories at a named commit; we did not rerun them, and figures that appear only in a README, report or pitch are left out. In an alert
generator for civil protection, Apertus v1.5 8B filled code-built slots with a locked
official hazard term; the official term and all facts appeared verbatim in 99\,\% of 120
alerts, against 45\,\% when the same model translated freely (70B: 98\,\% against
71\,\%). There the model still saw the facts and code checked them afterwards, a weaker
form of the pattern. In a beneficial-ownership tool, 8B in Q4\_K\_M on a CPU, citing line
identifiers that code verified, got 14 of 14 cases right, with no held-out set. Selection
does not close every gap: in court-decision anonymisation, 8B reached a recall of 0.764
against 0.894 for 70B on 537 decisions, and in a statistics question-answering system a
rule baseline without any model matched 70B on the main set.

\begin{figure}[hp]
\centering
\includegraphics[width=\linewidth,height=0.82\textheight,keepaspectratio]{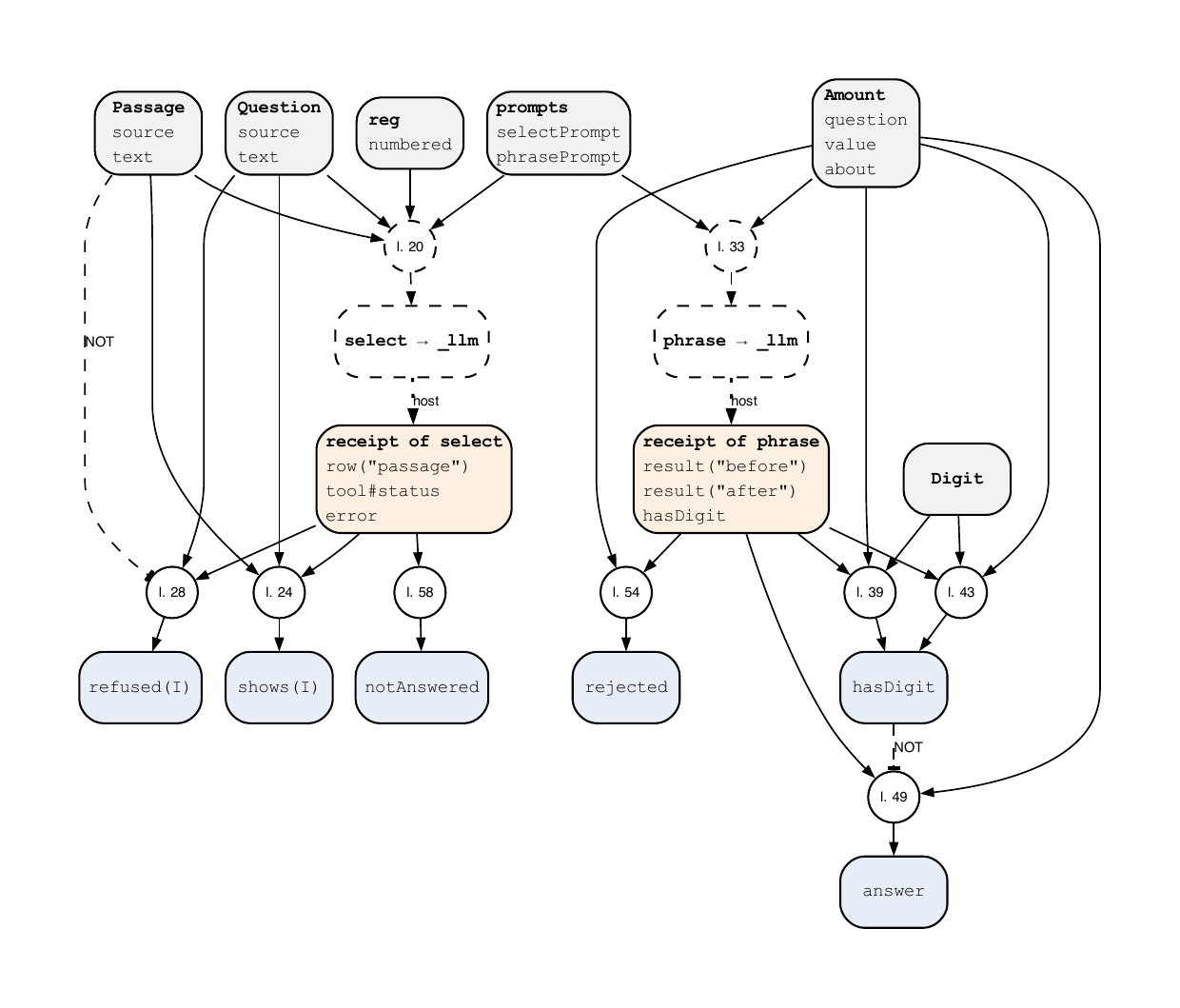}
\caption{Rule graph of \texttt{p12-select.fx}, generated from the program. A circle is one rule; ``l.~23'' means that the rule starts in line 23 of the listing. Legend in \cref{tab:rglegend}.}
\label{fig:rg-select}

\end{figure}

\end{patternform}

\section{Patterns of control: who decides}
\label{sec:control}

\FloatBarrier
\subsection{Pattern 4: Model Proposes, Rules Dispose}
\label{pat:dispose}

\begin{patternform}
\item[Intent] Let the model propose an answer, an action or a choice, and let a rule
admit it, refuse it with a reason, or narrow what may be proposed.

\item[Also known as] Guardrail, action gating, policy enforcement point, bounded recommendation.

\item[Motivation] An agent proposes to write a file, a model recommends a portfolio
weight, an assistant proposes to send a refund. Each proposal may be reasonable, but
some must never happen: a write outside the task, a weight change below a materiality
threshold, a refund above an authority limit. The pattern keeps the model's initiative
and puts every effect behind a rule that can say no and say why.

\item[Applicability] Use it when the model acts (tool calls, messages, transactions),
when the space of admissible outcomes is known even if the best one is not, or when a
proposal must rest on evidence available at a given time.

\item[Structure]\leavevmode
\begin{center}
\begin{tikzpicture}[bx]
\node[bxinst] (d1) at (0,0) {symbol};
\node[bxnote, below=0.6mm of d1] {task, outcome};
\node[bxproc, right=of d1] (d2) {infer};
\node[bxmodel, above=of d2] (d2m) {model:\\gen};
\node[bxnote, above=0.3mm of d2m] {language model};
\draw[bxflow] (d2m)--(d2);
\draw[bxflow] (d1)--(d2);
\node[bxinst, right=of d2] (d3) {symbol};
\node[bxnote, below=0.6mm of d3] {proposal};
\draw[bxflow] (d2)--(d3);
\node[bxproc, right=of d3] (d4) {infer:\\deduce};
\node[bxmodel, above=of d4] (d4m) {model:\\sem};
\node[bxnote, above=0.3mm of d4m] {grants, limits};
\draw[bxflow] (d4m)--(d4);
\draw[bxflow] (d3)--(d4);
\node[bxinst, right=of d4] (d5) {symbol};
\node[bxnote, below=0.6mm of d5] {admitted or refused};
\draw[bxflow] (d4)--(d5);
\node[bxproc, right=of d5] (d6) {act};
\draw[bxflow] (d5)--(d6);
\node[bxinst, right=of d6] (d7) {symbol};
\node[bxnote, below=0.6mm of d7] {evidence};
\draw[bxflow] (d6)--(d7);
\draw[bxflow] (d5.south) -- ++(0,-6mm) -- node[below, font=\tiny]{outcome of the step, until finish or the bound} ([yshift=-6mm]d1.south -| d1.west) -- ++(-1.2mm,0) |- (d1.west);
\draw[bxreq] (d4.north) to[bend right=22] node[above, inner sep=1pt, font=\tiny]{request} (d2.north east);
\end{tikzpicture}
\end{center}

\item[Participants] \emph{Model} (proposes one action or value); \emph{Rules}
(capabilities, preconditions, limits, evidence requirements); \emph{Host} (executes
only what the rules admit); \emph{Person} (approves what the rules require).

\item[Collaborations] The proposal arrives as a receipt; rules derive admission or a
refusal with its reason; a refusal returns to the model, which replans; the step
bound stops a task whose proposals keep being refused. A task is complete when a rule derives it from
recorded evidence; a model saying it is done is one more proposal.

\item[Consequences] Effects become auditable and bounded; the model may still waste
turns on refused proposals. Conflicts between constraints are reported to a person, and
no priority order settles them silently. The pattern bounds what can happen; whether an
admitted proposal is the best one stays open.

\item[Implementation] Derive the options first where they are few (configurations,
offers in a legal band) and let the model choose among them; a rule checks that the
choice is one of them. Export the closed set of allowed values as the structured-output
schema of the call, and still check the receipt.

Each step is a new request \texttt{propose(?T, ?N)} whose
receipt becomes a proposal. A proposal is admitted only for an action the task grants;
any other is refused with a reason. An admitted look-up runs as a query on the rules
themselves (\texttt{\_query}), and the next step carries the outcome of the last one,
the refusal or the looked-up result. The task is complete when an admitted
\emph{finish} rests on evidence, and it stops at the step bound otherwise:
\begin{lstlisting}[style=fxprog,basicstyle=\footnotesize\ttfamily,language=framex]
// Pattern 4, Model Proposes, Rules Dispose: one action per step, admitted or refused.
world open.
closed slot grants.

// The requirements, and a task with the actions it grants.
mbi[requires -> "recognised Bachelor"; requires -> "English C1"].
t1:Task[goal -> "Tell Nora which documents the Master in Business Innovation requires.";
    actions -> "lookUp: read the requirements; sendOffer: send Nora an admission offer; finish: answer Nora with what you found"; maxSteps -> 4].
t1[grants -> "lookUp"; grants -> "finish"].

proposePrompt:Prompt[text -> "Propose the next action for the goal. outcome says what became of your last action; once it answers the goal, choose finish. Answer as JSON {\"action\": one of the actions}."].

// Step 1: the model proposes one action; outcome is "none" before the first step.
external propose(?T, ?N) using _llm with proposePrompt[goal -> ?G; actions -> ?Acts;
        outcome -> "none"; tool#order -> 2]
  <- ?T:Task[goal -> ?G; actions -> ?Acts] AND ?N := 1.
?R:Proposal[task -> ?T; step -> ?N; action -> ?K]
  <- ?R[tool#name -> "_llm"; tool#arg1 -> ?T; tool#arg2 -> ?N;
        result("action") -> ?K].

// Dispose: an action is admitted only if the task grants it.
?A[allowed -> true] <- ?A:Proposal[task -> ?T; action -> ?K] AND ?T[grants -> ?K].
?A[refused -> "action not granted for this task"]
  <- ?A:Proposal[task -> ?T; action -> ?K] AND NOT ?T[grants -> ?K].

// An admitted lookUp runs as a query on the rules themselves.
external lookUp(?T) using _query[query -> "?- mbi[requires -> ?X].";
        tool#order -> 1]
  <- ?A:Proposal[task -> ?T; action -> "lookUp"; allowed -> true].

?T[evidence -> true]
  <- ?R[tool#name -> "_query"; tool#arg1 -> ?T; tool#status -> "succeeded"].

// The next step carries the outcome of the last one; maxSteps stops the loop.
?T[next(?M) -> ?W]
  <- ?A:Proposal[task -> ?T; step -> ?N; refused -> ?W]
     AND ?T[maxSteps -> ?Max] AND ?N < ?Max AND ?M := ?N + 1.

?T[next(?M) -> ?X]
  <- ?A:Proposal[task -> ?T; step -> ?N; action -> "lookUp"; allowed -> true]
     AND ?R[tool#name -> "_query"; tool#arg1 -> ?T; result("text") -> ?X]
     AND ?T[maxSteps -> ?Max] AND ?N < ?Max AND ?M := ?N + 1.

external propose(?T, ?N) using _llm with proposePrompt[goal -> ?G; actions -> ?Acts;
        outcome -> ?W; tool#order -> 2]
  <- ?T:Task[goal -> ?G; actions -> ?Acts; next(?N) -> ?W].

// Completion only on recorded evidence.
?T[verdict -> "completed"]
  <- ?A:Proposal[task -> ?T; action -> "finish"; allowed -> true]
     AND ?T[evidence -> true].

?T[verdict -> "stopped at the step bound"]
  <- ?A:Proposal[task -> ?T; step -> ?N; action -> ?K] AND ?T[maxSteps -> ?N]
     AND ?K != "finish".

?- ?A:Proposal[step -> ?N; action -> ?K].
?- ?A[refused -> ?Why].
?- t1[verdict -> ?V].
\end{lstlisting}
\Cref{fig:rg-dispose} draws the program as a rule graph.

\item[Library form] With \texttt{agent.fx} (\cref{sec:libraries}) the program states the
task, its prompt and what it grants, and keeps what each granted action does: here the
query behind \texttt{lookUp}, its outcome for the next step, and the evidence that
completion rests on. Step loop, admission, refusal, bound and verdict come from the library:
\begin{lstlisting}[style=fxprog,basicstyle=\footnotesize\ttfamily,language=framex]
// Pattern 4 in library form: load agent.fx first.
world open.

mbi[requires -> "recognised Bachelor"; requires -> "English C1"].
proposePrompt:Prompt[text -> "Propose the next action for the goal. outcome says what became of your last action; once it answers the goal, choose finish. Answer as JSON {\"action\": one of the actions}."].

t1:ag#Task[ag#maxSteps -> 4; ag#prompt -> proposePrompt;
    ag#goal -> "Tell Nora which documents the Master in Business Innovation requires.";
    ag#actions -> "lookUp: read the requirements; sendOffer: send Nora an admission offer; finish: answer Nora with what you found"].
t1[ag#grants -> "lookUp"; ag#grants -> "finish"].

// What the action lookUp does, and what it reports; this stays with the program.
external lookUp(?T) using _query[query -> "?- mbi[requires -> ?X]."; tool#order -> 1]
  <- ?A:ag#Proposal[ag#task -> ?T; ag#action -> "lookUp"; ag#allowed -> true].
?T[ag#outcome(?N) -> ?X]
  <- ?A:ag#Proposal[ag#task -> ?T; ag#step -> ?N; ag#action -> "lookUp"; ag#allowed -> true]
     AND ?R[tool#name -> "_query"; tool#arg1 -> ?T; result("text") -> ?X].
?T[ag#evidence -> true]
  <- ?R[tool#name -> "_query"; tool#arg1 -> ?T; tool#status -> "succeeded"].

?- ?A:ag#Proposal[ag#step -> ?N; ag#action -> ?K].
?- ?A[ag#refused -> ?Why].
?- t1[ag#verdict -> ?V].
\end{lstlisting}

\item[Known uses] FrameX-Agents (one action at a time, admitted against the task's
capabilities, completion from evidence); allocation recommendation with point-in-time
evidence and a materiality gate; casework actions that need a native authorisation
proof.
\knownlit{dispose}

\item[Related patterns] Critique and Repair (\cref{pat:repair}) handles refused
proposals. Rules Steer the Model (\cref{pat:steer}) decides what the model is asked to
propose next.
\item[Variants] \emph{Simulator Disposes.} A simulation instead of a rule decides about
the proposal: the host runs the proposed plan in a simulator, and its receipt (reached,
collided, within budget) is what the rules accept or refuse. The simulator is an
external tool like any other, so the receipt keeps its version and inputs, and the rules
still make the decision. It is close to the preview in a copy of the session of Model as
Hypothesis Generator (\cref{pat:hypothesis}). Reported in Hack Apertus submissions
(Lightpath, Sturmglocke, Rescue Briefings, Forester); not built in FrameX. In one of
them, simulated violations were 16.4\,\% when the model chose the path and 10.5\,\% when a
rule engine did, over three seeds (result file committed in the submission).

\item[Evidence] One demonstration run%
. No measurement isolates
this pattern yet. It runs in every action of the
agent framework and of the allocation application (\cref{tab:compositions}). The
consequences above describe how those systems are designed and have not been measured.

\begin{figure}[hp]
\centering
\includegraphics[width=\linewidth,height=0.82\textheight,keepaspectratio]{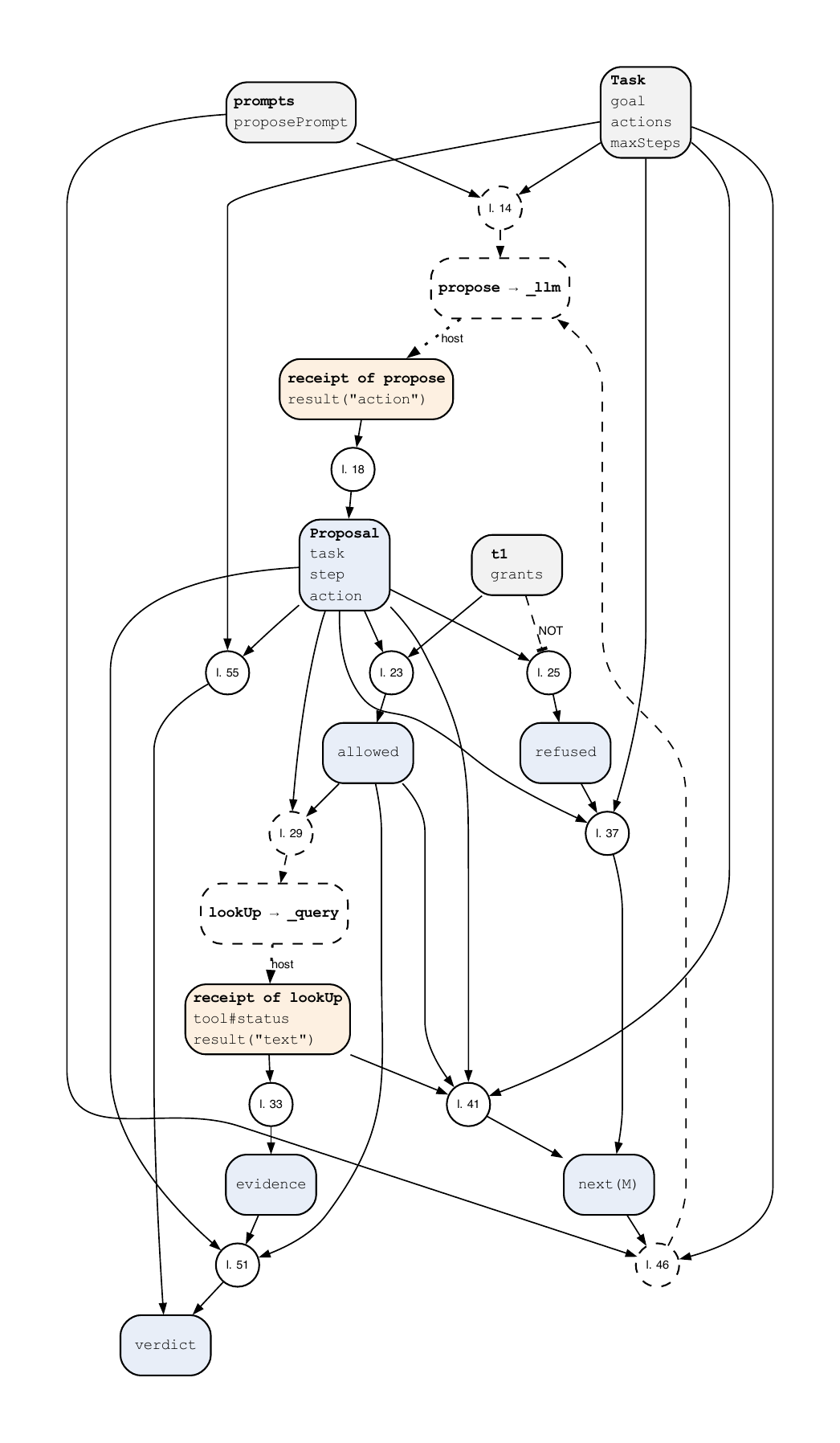}
\caption{Rule graph of \texttt{p04-dispose.fx}, generated from the program. A circle is one rule; ``l.~23'' means that the rule starts in line 23 of the listing. Legend in \cref{tab:rglegend}.}
\label{fig:rg-dispose}
\end{figure}

\end{patternform}

\FloatBarrier
\subsection{Pattern 5: Rules Steer the Model}
\label{pat:steer}

\begin{patternform}
\item[Intent] Let rules decide what the model is asked, when, with which context and
which tools. The model never decides the next step.

\item[Also known as] Dialogue flow, guided conversation, symbolic orchestration,
state machine with model leaves, rule-selected context.

\item[Motivation] A triage call must ask about emergency signs first, ask one thing at
a time, ask again when an answer is unclear, and hand over to a nurse when a sign is
present. A model asked to run the whole call from a long prompt skips questions as the
protocol grows. The pattern turns the protocol into rules: a flow derives the current
step, the model phrases its question, a judge decides whether the answer satisfies the
step, and rules decide branching, re-asking, hand-over and the end.

\item[Applicability] Use it when the process is known (a protocol, an interview, an
intake, an agent's phases), when the number of rules a model would have to apply
by itself grows with the case, or when what the model may see must be decided by
policy (permissions, point-in-time availability, jurisdiction).

\item[Structure]\leavevmode
\begin{center}
\begin{tikzpicture}[bx]
\node[bxinst] (e1) at (0,0) {symbol};
\node[bxnote, below=0.6mm of e1] {state};
\node[bxproc, right=of e1] (e2) {infer:\\deduce};
\node[bxmodel, above=of e2] (e2m) {model:\\sem};
\node[bxnote, above=0.3mm of e2m] {flow rules};
\draw[bxflow] (e2m)--(e2);
\draw[bxflow] (e1)--(e2);
\node[bxinst, right=of e2] (e3) {symbol};
\node[bxnote, below=0.6mm of e3] {step};
\draw[bxflow] (e2)--(e3);
\node[bxproc, right=of e3] (e4) {infer};
\node[bxmodel, above=of e4] (e4m) {model:\\gen};
\node[bxnote, above=0.3mm of e4m] {dialogue model};
\draw[bxflow] (e4m)--(e4);
\draw[bxflow] (e3)--(e4);
\node[bxinst, right=of e4] (e5) {data:\\text};
\node[bxnote, below=0.6mm of e5] {question};
\draw[bxflow] (e4)--(e5);
\node[bxactor, right=of e5] (e6) {};
\node[bxnote, below=0.6mm of e6] {person};
\draw[bxflow] (e5)--(e6);
\node[bxinst, right=of e6] (e7) {data:\\text};
\node[bxnote, below=0.6mm of e7] {answer};
\draw[bxflow] (e6)--(e7);
\node[bxproc, right=of e7] (e8) {infer};
\node[bxmodel, above=of e8] (e8m) {model:\\gen};
\node[bxnote, above=0.3mm of e8m] {judge};
\draw[bxflow] (e8m)--(e8);
\draw[bxflow] (e7)--(e8);
\node[bxinst, right=of e8] (e9) {symbol};
\node[bxnote, below=0.6mm of e9] {verdict};
\draw[bxflow] (e8)--(e9);
\draw[bxreq] (e2.south) .. controls +(0,-10mm) and +(0,-10mm) .. node[pos=0.5, below, inner sep=1pt, font=\tiny]{request} (e4.south);
\draw[bxflow] (e9.south) -- ++(0,-13mm) -- node[below, font=\tiny]{the verdict updates the state} ([yshift=-13mm]e1.south -| e1.west) -- ++(-1.2mm,0) |- (e1.west);
\end{tikzpicture}
\end{center}

\item[Participants] \emph{Rules} (flow steps, slots, branches, phase, permitted tools,
selected sources, missing evidence); \emph{Dialogue model} (phrases the current
question); \emph{Judge} (one criterion about one answer, a verdict from a fixed set);
\emph{Person} (answers, or takes over).

\item[Collaborations] Each turn: rules derive the step; the dialogue model asks; the
person answers; the judge returns \emph{answered}, \emph{not answered} or
\emph{unclear} with a reason; rules decide the next step. An unanswered question never
decides a branch.

\item[Consequences] The protocol is executed completely regardless of its length; the
model's work per call stays small, so a weaker model suffices. Every turn has a
receipt. The cost is one or more calls per turn, and a judge's verdict that contradicts
its own reason is a failure the rules must catch.

\item[Implementation] Ask one thing per question. Name verdict labels that cannot be
read as content (\emph{answered} instead of \emph{yes}). Let the judge read the earlier
answers it needs. Derive the missing evidence for a goal and ask for exactly
that. Select only the rules that apply or nearly apply to the case
as context.

A flow is data that the rules of FrameX's flow library read. The library is the file
\texttt{flow.fx} of FrameX, loaded before the program; the program itself contains only
facts:
\begin{lstlisting}[style=fxprog,basicstyle=\footnotesize\ttfamily,language=framex]
// Pattern 5, Rules Steer the Model: a dialogue flow written as facts. FrameX's flow
// library (python/framex/policies/flow.fx, loaded first) holds the rules that decide
// which step is active, request _turn (the dialogue model asks the question and returns
// the person's answer) and _judge (a judge model decides the step's criterion), ask
// again after a negative or unclear verdict, and repair a reply that breaks the format
// (Pattern 6).
world open.

advice:flow#Flow[flow#start -> programme].

// Ask which programme is meant; at most two attempts, then hand on to an officer.
programme:flow#Collect[flow#ask -> "Which Master do you mean?";
    flow#criterion -> "Does the person name Business Innovation or Management?";
    flow#slot -> programmeGiven; flow#maxAttempts -> 2;
    flow#ifSlot -> programmeGiven; flow#to -> language;
    flow#otherwise -> officer].

// Ask for the language certificate; at most three attempts (the default).
language:flow#Collect[flow#ask -> "Which English certificate do you hold?";
    flow#criterion -> "Does the person name an English certificate at level C1?";
    flow#slot -> c1Named;
    flow#ifSlot -> c1Named; flow#to -> check; flow#otherwise -> missing].

// Three ways to end.
check:flow#Say[flow#text -> "Thank you. We check your record."; flow#next -> flow#end].
missing:flow#Say[flow#text -> "An English certificate at level C1 is missing."; flow#next -> flow#end].
officer:flow#Say[flow#text -> "An admissions officer will contact you."; flow#next -> flow#end].

?- ?S:flow#Active.
?- workflow[status -> ?St].
\end{lstlisting}
\begin{figure}[p]
\centering
\includegraphics[width=\linewidth,height=0.82\textheight,keepaspectratio]{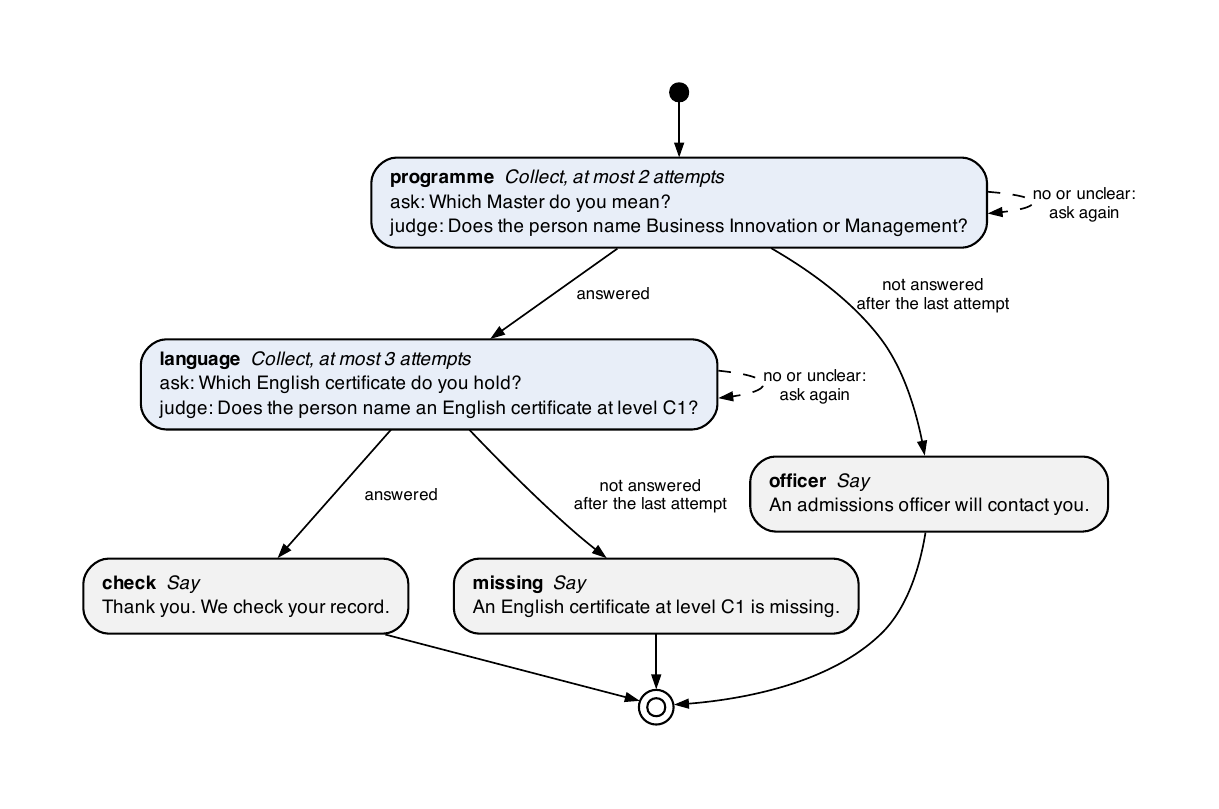}
\caption{Flow diagram of \texttt{p05-steer.fx}: the steps of the flow and the transitions the flow library derives from them. Legend in \cref{tab:rglegend}.}
\label{fig:rg-steer}
\end{figure}
\Cref{fig:rg-steer} draws the flow as a diagram.
A \texttt{flow\#Collect} step asks \texttt{flow\#ask}, lets a judge decide
\texttt{flow\#criterion} about the answer and records the outcome in the slot named by
\texttt{flow\#slot}. \texttt{flow\#ifSlot}, \texttt{flow\#to} and
\texttt{flow\#otherwise} branch on a slot; \texttt{flow\#maxAttempts} bounds the
attempts (three by default); a \texttt{flow\#Say} step shows a text, and
\texttt{flow\#end} ends the flow. The library derives which step holds the token
(\texttt{flow\#Active}) and which attempt is due (\texttt{flow\#askNow}), and it
issues two model calls per attempt. \texttt{advice:flow\#Flow[flow\#start -> programme]}
declares the flow and its first step. A rejected attempt (\texttt{flow\#rejected}) is
asked again while attempts remain below the step's limit (\texttt{flow\#limit}, from
\texttt{flow\#maxAttempts}). Four of the library's rules, shown only to make the chain
visible:
\begin{lstlisting}[language=framex]
// From FrameX's flow.fx (library; a program does not repeat these).
external turn(?S, ?A) using _turn[question -> ?Q; attempt -> ?A; tool#order -> 1]
    <- ?S[flow#askNow -> ?A; flow#ask -> ?Q].

?S[flow#answer(?A) -> ?Text]
    <- ?R[tool#arg1 -> ?S; tool#arg2 -> ?A; result("answer") -> ?Text].

external judge(?S, ?A) using _judge[criterion -> ?C; question -> ?Q;
        answer -> ?Text; tool#order -> 2]
    <- ?S[flow#answer(?A) -> ?Text; flow#criterion -> ?C; flow#ask -> ?Q].

?S[flow#askNow -> ?B]
    <- ?S[flow#rejected(?A) -> true; flow#limit -> ?M] AND ?B := ?A + 1 AND ?B <=?M.
\end{lstlisting}
\texttt{\_turn} lets the dialogue model phrase the question, shows it to the person and
returns the answer; \texttt{\_judge} gives the judge model the criterion and that answer.
For a Collect step the judge answers with the labels \emph{answered}, \emph{not
answered} or \emph{unclear}; the host writes them as \texttt{"yes"}, \texttt{"no"} and
\texttt{"unclear"} into \texttt{result("verdict")}, which the library reads, and keeps the
label itself in \texttt{result("label")}. In the recorded run the first answer was ``the
Master''; the receipts read (without the line \texttt{tool\#completed -> true} that the host also
writes):
\begin{lstlisting}[language=framex]
`turn(programme, 1)`[tool#status -> "succeeded"; result("answer") -> "the Master"].

`judge(programme, 1)`[tool#status -> "succeeded"; result("verdict") -> "no";
    result("label") -> "not answered"; result("reason") ->
    "The answer \"the Master\" does not mention Business Innovation or Management."].
\end{lstlisting}
A \emph{no} or \emph{unclear} verdict rejects the attempt, so the library asks the
programme question a second time, and the dialogue model rephrases it with the judge's
reason. The person answers at the terminal or, for a run that can be repeated, from a
file of answers%
.

\item[Library form] Pattern~5 is written in library form from the start. FrameX's flow
library \texttt{flow.fx} holds the rules, and the program above states facts only.

\item[Known uses] Interview, suitability and triage flows; customer support with
hand-over; FrameX-Agents phases, grants and context; point-in-time retrieval in the
allocation application; the evidence-needs search for the questions that settle a goal.
\knownlit{steer}

\item[Related patterns] Model as Sensor (\cref{pat:sensor}) is the judge's role;
Critique and Repair (\cref{pat:repair}) repairs malformed turns; Division by Competence
(\cref{pat:competence}) sends \emph{unclear} to a person.

\item[Evidence] With synthetic intake protocols of 4 to 96 questions, the hybrid
decided all 770 runs correctly; a model applying the same rules in the conversation
fell to 39 of 60 at 64 questions and 31 of 60 at 96, with 45 of its 50 errors unsafe
(McNemar $p<0.001$). Without reasoning, the same model decided 10 of 30 conversations
as a pure model and 30 of 30 in the hybrid. On a ladder of six local models, no model
up to 70 billion parameters reached the criterion as a pure model, while a 9-billion
model sufficed for the hybrid in single decisions. Cost favours neither side: with
cached input priced, the two cost about the same at 64 and 96 questions. In a
telephone-triage head-to-head on 14 simulated callers, a model alone was first more
accurate (274 of 280 against 243 of 280). The hybrid's errors had one cause, found in its
recorded verdicts and reasons and corrected in one place in the judge; afterwards it was
right in 278 of 280 runs and in 199 of 200 on unseen callers, level with the model alone
within noise, at lower cost per call and with every level traced to recorded verdicts.
\end{patternform}

\FloatBarrier
\subsection{Pattern 6: Critique and Repair}
\label{pat:repair}

\begin{patternform}
\item[Intent] When a model's reply fails a check, classify the failure by rule, state
it, and ask again with the problem and the faulty reply in the prompt, a bounded
number of times; stop on repetition.

\item[Also known as] Generate, verify, repair; validated retry; feedback loop with an
external critic.

\item[Motivation] At temperature zero a model gives the same reply to the same input;
asking again without changing the input does not help, and a model asked to check
itself without external feedback often makes things worse. The rules know what is
wrong (a key misspelled, a quote not found, an action not allowed, rows missing) and
can say so.

\item[Applicability] Use it whenever another pattern checks a reply: format errors,
ungrounded statements, refused proposals, incomplete extractions, failed regression
cases of a draft.

\item[Structure]\leavevmode
\begin{center}
\begin{tikzpicture}[bx]
\node[bxinst] (f1) at (0,0) {data:\\text};
\node[bxnote, below=0.6mm of f1] {reply};
\node[bxproc, right=of f1] (f2) {trans-\\form};
\node[bxnote, below=0.6mm of f2] {host check};
\draw[bxflow] (f1)--(f2);
\node[bxinst, right=of f2] (f3) {symbol};
\node[bxnote, below=0.6mm of f3] {failure kind};
\draw[bxflow] (f2)--(f3);
\node[bxproc, right=of f3] (f4) {infer:\\deduce};
\node[bxmodel, above=of f4] (f4m) {model:\\sem};
\node[bxnote, above=0.3mm of f4m] {rules};
\draw[bxflow] (f4m)--(f4);
\draw[bxflow] (f3)--(f4);
\node[bxinst, right=of f4] (f5) {symbol};
\node[bxnote, below=0.6mm of f5] {problem, or stop};
\draw[bxflow] (f4)--(f5);
\node[bxproc, right=of f5] (f6) {infer};
\node[bxmodel, above=of f6] (f6m) {model:\\gen};
\node[bxnote, above=0.3mm of f6m] {same model};
\draw[bxflow] (f6m)--(f6);
\draw[bxflow] (f5)--(f6);
\node[bxinst, right=of f6] (f7) {data:\\text};
\node[bxnote, below=0.6mm of f7] {new reply};
\draw[bxflow] (f6)--(f7);
\draw[bxreq] (f4.south) .. controls +(0,-10mm) and +(0,-10mm) .. node[pos=0.5, below, inner sep=1pt, font=\tiny]{request} (f6.south);
\draw[bxflow] (f7.south) -- ++(0,-13mm) -- node[below, font=\tiny]{checked again, a bounded number of times} ([yshift=-13mm]f1.south -| f1.west) -- ++(-1.2mm,0) |- (f1.west);
\end{tikzpicture}
\end{center}

\item[Participants] \emph{Validator} in the host (measures format and occurrence);
\emph{Rules} (classify, decide, bound); \emph{Model}; \emph{Person} or a stronger model
(escalation).

\item[Collaborations] The rules classify a failure into one of six kinds and respond
per kind:
\begin{center}\small
\begin{tabular}{@{}lp{0.6\linewidth}@{}}
\toprule
Failure & Response \\
\midrule
wrong format & ask again with the problem stated, up to a bound; a repeated failure stops with its reason \\
not grounded & reject the statement with its reason \\
not allowed & refuse with a reason the model receives on its next turn \\
incomplete & derive what is missing; the answer becomes \emph{unknown} \\
failed call & stop with the reason; never treat it as an \emph{unclear} answer \\
plausible but wrong & not detectable from the reply; needs grounding, a second reader or a person \\
\bottomrule
\end{tabular}
\end{center}
Each repair is a new request with its own identity and receipt; both replies stay
recorded. The first five kinds are results the rules can tell apart, as invariant~5
requires. The sixth is the class that no check can detect; we list it so that a passed
check is never read as protection against it.

\item[Consequences] Repairs succeed when the message names the fix. The bound keeps
cost predictable. Silent correction is excluded: mapping ``ja'' to \emph{yes} is
guessing and is refused; a correction is accepted without a call only when trusted
facts entail it, as a derived fact with its own proof.

\item[Implementation] For a plain model call, the repair is part of the program. The
program of Model as Author (\cref{pat:author}) contains it: every failure keeps the
checker's message, and a new request \texttt{draft(?P, ?M)} with the next number hands
the model its rule and that message. It stops at \texttt{maxAttempts}, and earlier when
the same message comes back twice (\texttt{stalled}):
\begin{lstlisting}[style=fxprog,language=framex,firstline=45,lastline=61]
// Pattern 6: ask again with the rule and the problem in the request, as a new
// request with the next number. Stop at the bound or when the message repeats.
?P[newProblem(1) -> true] <- ?P[failed(1) -> ?F].

?P[newProblem(?N) -> true]
  <- ?P[failed(?N) -> ?F] AND ?K := ?N - 1 AND ?P[failed(?K) -> ?G] AND ?F != ?G.

external draft(?P, ?M) using _llm with repairPrompt[policy -> ?T; vocabulary -> ?V;
        previousRule -> ?Old; problem -> ?F; tool#order -> 2]
  <- ?P:Policy[text -> ?T; vocabulary -> ?V; maxAttempts -> ?Max;
        rule(?N) -> ?Old; failed(?N) -> ?F; newProblem(?N) -> true]
     AND ?N < ?Max AND ?M := ?N + 1.

?P[stalled -> ?N]
  <- ?P[failed(?N) -> ?F] AND ?K := ?N - 1 AND ?P[failed(?K) -> ?F].

?P[gaveUp -> ?F] <- ?P:Policy[maxAttempts -> ?N; failed(?N) -> ?F].
\end{lstlisting}
In the recorded run the first draft lacked its final full stop and was rejected before
evaluation; the repair request carried the message ``expected `.'\,'', and the second
draft passed all cases%
.

In a dialogue flow the library does the same for the dialogue model. The host validates
each reply of \texttt{\_turn} against the agreed format and writes what it found into
the receipt; a rule of the library reads a format error and requests
\texttt{\_repairTurn}, which calls the same model again with the question, the problem
and the faulty reply:
\begin{lstlisting}[language=framex]
// From FrameX's flow.fx (library).
external repair(?S, ?A) using _repairTurn[question -> ?Q; attempt -> ?A;
        problem -> ?E; received -> ?X; tool#order -> 1]
    <- ?R[tool#name -> "_turn"; tool#arg1 -> ?S; tool#arg2 -> ?A;
          result("formatError") -> ?E; result("received") -> ?X]
    AND ?S[flow#ask -> ?Q].
\end{lstlisting}
The receipt that triggers it reads, for example,
\texttt{result("formatError") -> "missing key: answer"}. A second broken reply ends the
flow with \texttt{workflow[status -> "interrupted"]} and its reason.

\item[Library form] The repair loop is part of two libraries: \texttt{author.fx} for drafts
checked by a tool (\cref{pat:author}) and FrameX's \texttt{flow.fx} for dialogue turns
(\cref{pat:steer}). A program that loads one of them gets the bounded repair without
writing a rule.

\item[Known uses] Format repair of dialogue turns and judges; refused agent actions
and premature finishes with their failed checks; completeness checks that turn a
partial extraction into \emph{unknown}; the rule-drafting loop with the checker's
message.
\knownlit{repair}

\item[Related patterns] It completes Model as Sensor, Model as Author and Model Proposes, Rules Dispose. Model as
Critic (\cref{pat:critic}) is its reverse: there the model criticises the rules.

\item[Evidence] See Model as Author: naming the fix raised first-request repairs from
5 of 10 to 10 of 10. In the literature, self-correction without an external signal lowered
accuracy, and the gains reported for self-refinement rested on oracle stopping signals
\parencite{huang2024cannot}; back-prompting from a sound validator raised executable plans
from about 12\,\% to 82\,\% in one planning domain \parencite{kambhampati2024llmmodulo}.
The rules play the part of that signal and of that validator.
\end{patternform}

\FloatBarrier
\subsection{Pattern 7: Division by Competence}
\label{pat:competence}

\begin{patternform}
\item[Intent] Let the rules decide the cases they cover, send what they leave open to
a model or a person, and turn recurring answers into rules over time.

\item[Also known as] Selective delegation, escalation cascade, tiered autonomy, rule
distillation, human in the loop.

\item[Motivation] Most cases of a claims process are routine and fully covered by
rules; a few are ambiguous. Sending every case to a model costs money and gives up
proofs where they were available; sending none leaves the ambiguous cases undecided.
The pattern uses the rules' own three-valued result as the gate: \emph{true} and
\emph{false} are decided, \emph{unknown}, \emph{unclear} and conflicts are handed on.

\item[Applicability] Use it when coverage by rules is high but incomplete, when the number
of model calls should follow the number of ambiguous cases, and when a person must decide
the residue.

\item[Structure]\leavevmode
\begin{center}
\begin{tikzpicture}[bx]
\node[bxinst] (g1) at (0,0) {symbol};
\node[bxnote, below=0.6mm of g1] {case};
\node[bxproc, right=of g1] (g2) {infer:\\deduce};
\node[bxmodel, above=of g2] (g2m) {model:\\sem};
\node[bxnote, above=0.3mm of g2m] {rules};
\draw[bxflow] (g2m)--(g2);
\draw[bxflow] (g1)--(g2);
\node[bxinst, right=of g2] (g3) {symbol};
\node[bxnote, below=0.6mm of g3] {unknown or conflict};
\draw[bxflow] (g2)--(g3);
\node[bxproc, right=of g3] (g4) {infer};
\node[bxmodel, above=of g4] (g4m) {model:\\gen};
\node[bxnote, above=0.3mm of g4m] {language model};
\draw[bxflow] (g4m)--(g4);
\draw[bxflow] (g3)--(g4);
\node[bxinst, right=of g4] (g5) {data:\\text};
\node[bxnote, below=0.6mm of g5] {assessment};
\draw[bxflow] (g4)--(g5);
\node[bxactor, right=of g5] (g6) {};
\node[bxnote, below=0.6mm of g6] {officer};
\draw[bxflow] (g5)--(g6);
\node[bxinst, right=of g6] (g7) {symbol};
\node[bxnote, below=0.6mm of g7] {decision};
\draw[bxflow] (g6)--(g7);
\draw[bxreq] (g2.north) to[bend left=25] node[above, inner sep=1pt, font=\tiny]{request} (g4.north west);
\node[bxinst, below=9mm of g3] (g3b) {symbol}; \node[bxnote, below=0.6mm of g3b] {decided by the rules};
\draw[bxflow] (g2.south) |- (g3b.west);
\end{tikzpicture}
\end{center}

\item[Participants] \emph{Rules} (decide, mark the residue); \emph{Model} (assesses
the residue, as a recorded assessment); \emph{Person}; \emph{Counter} (counts recurring
model decisions).

\item[Collaborations] Rules decide; a case left \emph{unknown} or in conflict becomes
a request; the model's verdict is an assessment that travels with the case to a
person and decides nothing; recurring verdicts become candidate rules for Model as Author.

\item[Consequences] The rule part grows and the model part shrinks over time. The gate
must never be the model's own confidence; use \emph{unknown}, conflict, or a measured
precision per model and task.

\item[Implementation] Gate on the rules' own result. Under an open world a case counts as decided only when the rules derive a
decision. Every other case goes to the model for an assessment and, with that
assessment, to a person:
\begin{lstlisting}[style=fxprog,basicstyle=\footnotesize\ttfamily,language=framex]
// Pattern 7, Division by Competence: the rules decide what they cover; the rest is
// assessed by the model and handed on to a person, with the assessment as a note.
world open.
closed slot decided. closed slot admitted. closed slot notAdmitted.

// Three files; the rules have decided two of them.
nora:Applicant[file -> "Bachelor HSG; English C1 certificate on file."].
tomas:Applicant[file -> "Bachelor ETH; English C1 recorded as failed."].
yara:Applicant[file -> "Bachelor University of Bologna; IELTS 7.0."].
nora[admitted -> mbi]. tomas[notAdmitted -> mbi].

assessPrompt:Prompt[text -> "Assess this admission file for the MBI. Answer as JSON {\"verdict\": likely or unlikely or unclear, \"reason\": ...}."].

?A[decided -> true] <- ?A[admitted -> ?P].
?A[decided -> true] <- ?A[notAdmitted -> ?P].

// Only an undecided file goes to the model.
external assess(?A) using _llm with assessPrompt[file -> ?T]
  <- ?A:Applicant[file -> ?T] AND NOT ?A[decided -> true].

?A[assessment -> ?V]
  <- ?R[tool#name -> "_llm"; tool#arg1 -> ?A; result("verdict") -> ?V].

?A[assessmentReason -> ?W]
  <- ?R[tool#name -> "_llm"; tool#arg1 -> ?A; result("reason") -> ?W].

// Every undecided file goes to an officer. The model's assessment travels with it as
// a note and decides nothing; a failed call is noted as such.
?A[handOn -> officer] <- ?A:Applicant AND NOT ?A[decided -> true].

?A[note -> ?W] <- ?A[handOn -> officer; assessmentReason -> ?W].

?A[note -> "no assessment: the model call failed"]
  <- ?A[handOn -> officer] AND ?R[tool#name -> "_llm"; tool#arg1 -> ?A;
        tool#status -> "failed"].

?- ?A[handOn -> ?W].
?- ?A[assessment -> ?V; note -> ?N].
\end{lstlisting}
\Cref{fig:rg-competence} draws the program as a rule graph.
The hand-over is a derived fact, \texttt{handOn -> officer}. Showing it to a person is the
host's task; FrameX's tool \texttt{\_ask} puts a question to a person as written and
returns the answer as a receipt.

\item[Library form] With \texttt{competence.fx} (\cref{sec:libraries}) the program states
the cases and keeps its rules for what it decides (\texttt{cp\#decided}). The library
assesses every other case and hands it on with the model's reason as a note:
\begin{lstlisting}[style=fxprog,basicstyle=\footnotesize\ttfamily,language=framex]
// Pattern 7 in library form: load competence.fx first.
world open.

assessPrompt:Prompt[text -> "Assess this admission file for the MBI. Answer as JSON {\"verdict\": likely or unlikely or unclear, \"reason\": ...}."].

nora:cp#Case[cp#prompt -> assessPrompt; cp#file -> "Bachelor HSG; English C1 certificate on file."].
tomas:cp#Case[cp#prompt -> assessPrompt; cp#file -> "Bachelor ETH; English C1 recorded as failed."].
yara:cp#Case[cp#prompt -> assessPrompt; cp#file -> "Bachelor University of Bologna; IELTS 7.0."].
nora[admitted -> mbi]. tomas[notAdmitted -> mbi].

// What the rules decide; this stays with the program.
?A[cp#decided -> true] <- ?A[admitted -> ?P].
?A[cp#decided -> true] <- ?A[notAdmitted -> ?P].

?- ?C[cp#handOn -> true; cp#assessment -> ?V; cp#note -> ?N].
\end{lstlisting}

\item[Known uses] Three-way branches in triage that send \emph{unclear} to a person;
\emph{awaiting user} in FrameX-Agents; required human review for evidence-gated
actions in casework.
\knownlit{competence}

\item[Related patterns] Model as Author (\cref{pat:author}) for distillation; Rules
Steer the Model (\cref{pat:steer}) for the hand-over.
\item[Evidence] One demonstration run%
. No measurement isolates
this pattern yet. The triage flows of
\cref{pat:steer} use it to send \emph{unclear} to a person; a measurement should report
the share of cases handed on, the accuracy of those decided by rules, and the cost per
decided case.

\begin{figure}[hp]
\centering
\includegraphics[width=\linewidth,height=0.82\textheight,keepaspectratio]{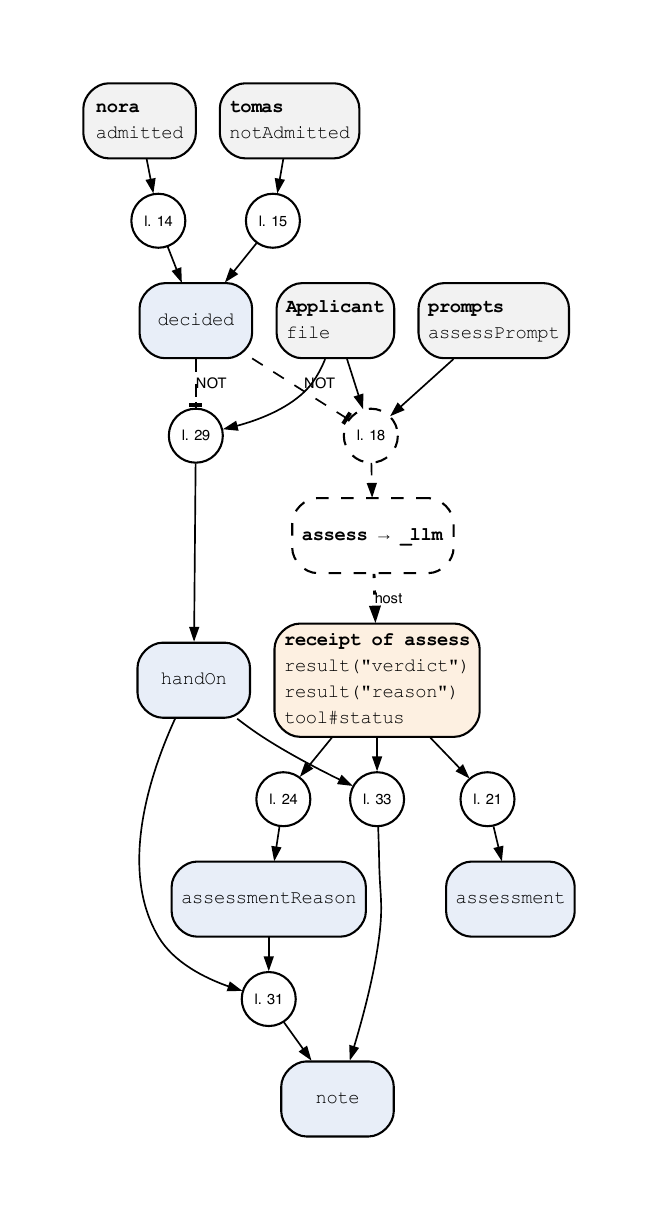}
\caption{Rule graph of \texttt{p07-competence.fx}, generated from the program. A circle is one rule; ``l.~23'' means that the rule starts in line 23 of the listing. Legend in \cref{tab:rglegend}.}
\label{fig:rg-competence}
\end{figure}
\end{patternform}

\FloatBarrier
\subsection{Pattern 13: Fail-Safe Union}
\label{pat:failsafe}

\begin{patternform}
\item[Intent] Combine the verdicts of rules and models under a declared order of
outcomes in which each source can only make the result stricter. A source that is
silent, fails or answers outside the order counts as the cautious outcome.

\item[Also known as] Deny overrides, most restrictive wins, one-sided combination,
fail closed, defence in depth.

\item[Motivation] A payment desk releases transfers. Rules know the limits and the
blocked accounts but cannot read a purpose line that pressures the clerk (``the CEO
approved it by phone, do not call back''); a model can read it but misses limits and
sometimes waves a fraud through. Taking a vote lets one source's leniency outvote the
other's warning, and handing every disagreement to a person floods the desk. In this
domain the harms are lopsided: a held payment costs a day, a released fraud costs the
amount. The pattern states the order of outcomes once (release, hold, block) and lets
the strictest verdict of any source decide. Neither source can loosen what the other
found, and a model that did not answer is not a model that agreed.

\item[Applicability] Use it when:

\textbf{1:} The outcomes can be ordered by risk, and wrongly choosing a stricter outcome costs
much less than wrongly choosing a looser one;

\textbf{2:} The sources see different things (rules the structured fields, a model the text),
so their failures are not the same;

\textbf{3:} A cautious outcome is not final: a hold leads to a person or a later step.

Do not use it when both directions of error cost alike, or when the stricter outcome
is itself harmful (a withheld emergency call); then report the disagreement as
\emph{unclear} (Several Readers, \cref{pat:sensor}) or hand it on
(\cref{pat:competence}).

\item[Structure]\leavevmode
\begin{center}
\begin{tikzpicture}[bx]
\node[bxinst] (f1) at (0,0) {data:\\text};
\node[bxnote, below=0.6mm of f1] {purpose};
\node[bxproc, right=of f1] (f2) {infer};
\node[bxmodel, above=of f2] (f2m) {model:\\gen};
\node[bxnote, above=0.3mm of f2m] {language model};
\draw[bxflow] (f2m)--(f2);
\draw[bxflow] (f1)--(f2);
\node[bxinst, right=of f2] (f3) {data:\\text};
\node[bxnote, below=0.6mm of f3] {reply};
\draw[bxflow] (f2)--(f3);
\node[bxproc, right=of f3] (f4) {trans-\\form};
\node[bxnote, below=0.6mm of f4] {receipt};
\draw[bxflow] (f3)--(f4);
\node[bxinst, right=of f4] (f5) {symbol};
\node[bxnote, below=0.6mm of f5] {verdict of the model};
\draw[bxflow] (f4)--(f5);
\node[bxinst, below=13mm of f3] (g1) {symbol};
\node[bxnote, below=0.6mm of g1] {payment};
\node[bxproc, below=13mm of f4] (g2) {infer:\\deduce};
\node[bxmodel, below=4mm of g2] (g2m) {model:\\sem};
\node[bxnote, below=0.3mm of g2m] {rules};
\draw[bxflow] (g2m)--(g2);
\draw[bxflow] (g1)--(g2);
\node[bxinst, below=13mm of f5] (g3) {symbol};
\node[bxnote, below=0.6mm of g3] {verdict of the rules};
\draw[bxflow] (g2)--(g3);
\node[bxproc, right=10mm of f5, yshift=-9mm] (u1) {infer:\\deduce};
\node[bxmodel, below=4mm of u1] (u1m) {model:\\sem};
\node[bxnote, below=0.3mm of u1m] {declared order};
\draw[bxflow] (u1m)--(u1);
\draw[bxflow] (f5)--(u1);
\draw[bxflow] (g3)--(u1);
\node[bxinst, right=of u1] (u2) {symbol};
\node[bxnote, below=0.6mm of u2] {strictest verdict, sources named};
\draw[bxflow] (u1)--(u2);
\draw[bxreq] (u1.north) .. controls +(0,16mm) and +(4mm,10mm) .. node[pos=0.45, above, inner sep=1pt, font=\tiny]{request} (f2.north east);
\end{tikzpicture}
\end{center}

\item[Participants] \emph{Rules}: give their own verdict from the structured fields and
combine all verdicts. \emph{Model}: gives a verdict from the text, one of the ordered
outcomes. \emph{Host}: calls the model and records a failed call as such. \emph{Order}:
the outcomes with their rank, declared as facts. \emph{Person}: takes up what is held.

\item[Collaborations] Each source derives a verdict for each case. A source without a
usable verdict (no receipt, a failed call, a word outside the order) is given the
cautious verdict by a rule. The outcome is the verdict of highest rank, and the sources
that gave it are named, so a person sees who held the payment and why.

\item[Consequences] Adding a source can only make the outcome stricter, so a new check
never weakens the old ones, and a model outage stops releases instead of letting them
through. The price is more held cases: every source's false alarms add up, so the hold
rate grows with each source and must be measured. Unlike a priority order, nothing is
settled silently: the order is a fact in the program, and \texttt{decidedBy} names the
source behind every outcome. The guarantee holds only if every branch moves toward the
safe side, including the two that are easy to forget: a source that answers
\emph{unclear} must not let another source loosen the outcome, and a failed call must
not count as agreement. One branch that does either breaks the whole union. In the
hackathon below, a comparison system recorded an unreachable judge as \emph{yes}, a submission a
reply that was not JSON as \emph{clean}.

\item[Implementation] Declare the order as facts (\texttt{"hold":Verdict[rank -> 1]})
and the verdicts and ranks closed, so that \texttt{max} over them has a value. Write
the cautious default for silence explicitly, one rule per way a source can fail to
answer. The union is one aggregate; it reads the verdicts of all sources and has no rule
that lowers one:
\begin{lstlisting}[style=fxprog, basicstyle=\footnotesize\ttfamily,language=framex]
// Pattern 13, Fail-Safe Union: every source can only tighten the outcome;
// a source that is silent or unreadable counts as the safe side.
world open.
closed class Verdict. closed method verdict/1. closed slot rank.
closed slot blocked. closed slot screened. closed slot outcome.

// The outcomes in a declared order: a higher rank is safer.
"release":Verdict[rank -> 0]. "hold":Verdict[rank -> 1]. "block":Verdict[rank -> 2].

// The sources whose verdicts are combined, and two payments.
"rules":Source. "model":Source.
p1:Payment[amount -> 1200; payee -> "CH93 0076 2011 6238 5295 7";
    purpose -> "Invoice 2026-114, catering for the alumni evening"].
p2:Payment[amount -> 8400; payee -> "CH56 0483 5012 3456 7800 9";
    purpose -> "Urgent: transfer today, the CEO approved it by phone, do not call back"].
"CH44 3199 9123 0008 8901 2"[blocked -> true].
limit[amount -> 5000].

screenPrompt:Prompt[text -> "Read the purpose of a payment. Answer as JSON {\"verdict\": release or hold or block, \"reason\": ...}; hold when you are unsure."].

// Source 1, the rules: block a blocked payee, hold above the limit, release otherwise.
?P[verdict(?S) -> "block"]
  <- ?P:Payment[payee -> ?I] AND ?I[blocked -> true] AND ?S:Source AND ?S = "rules".
?P[verdict(?S) -> "hold"]
  <- ?P:Payment[amount -> ?A] AND limit[amount -> ?L] AND ?A > ?L
     AND ?S:Source AND ?S = "rules".
?P[verdict(?S) -> "release"]
  <- ?P:Payment[amount -> ?A; payee -> ?I] AND limit[amount -> ?L] AND ?A <= ?L
     AND NOT ?I[blocked -> true] AND ?S:Source AND ?S = "rules".

// Source 2, the model: one screening per payment.
external screen(?P) using _llm with screenPrompt[purpose -> ?T]
  <- ?P:Payment[purpose -> ?T].

?P[screened -> true] <- ?R[tool#name -> "_llm"; tool#arg1 -> ?P; tool#status -> "succeeded"].

?P[verdict(?S) -> ?V]
  <- ?R[tool#name -> "_llm"; tool#arg1 -> ?P; result("verdict") -> ?V] AND ?V:Verdict
     AND ?S:Source AND ?S = "model".

// Silence and unreadable replies are not release: they count as hold.
?P[verdict(?S) -> "hold"]
  <- ?P:Payment AND NOT ?P[screened -> true] AND ?S:Source AND ?S = "model".
?P[verdict(?S) -> "hold"]
  <- ?R[tool#name -> "_llm"; tool#arg1 -> ?P; result("verdict") -> ?V] AND NOT ?V:Verdict
     AND ?S:Source AND ?S = "model".

// The union: the safest verdict of any source decides, and the sources that set it are named.
?P[outcome -> ?V]
  <- ?P:Payment AND ?M := max{?K | ?P[verdict(?S) -> ?W] AND ?W[rank -> ?K]}
     AND ?V:Verdict[rank -> ?M].
?P[decidedBy -> ?S] <- ?P[outcome -> ?V; verdict(?S) -> ?V].

?- ?P[outcome -> ?V].
?- ?P[decidedBy -> ?S].
\end{lstlisting}
\Cref{fig:rg-failsafe} draws the program as a rule graph. Before any receipt the program
holds both payments, because the model has not spoken (line~42). The rules give no
verdict for a payment without an amount; where inputs can be incomplete, add a cautious
default for the rule source as well. There is no library form yet.

\item[Known uses] In the submissions to the Hack Apertus hackathon (October 2026), as
described in the jury assessments, rules and a model could each only escalate in a
retrieval system for meditation texts, Inbrev, Schild, Mandate Desk and PoesIA.
In FrameX, the program above. A comparison system built for the hackathon looked like this pattern but is a
Silent Rule Override (\cref{sec:failures}).
\knownlit{failsafe}

\item[Related patterns] Several Readers, One Rule (\cref{pat:sensor}) makes a
disagreement \emph{unclear}; this pattern resolves it toward the safe side by a declared
order. Model Proposes, Rules Dispose (\cref{pat:dispose}) lets only the rules refuse;
here the model may refuse too, but never admit. Division by Competence
(\cref{pat:competence}) takes up what is held.

\item[Variants] \emph{Strict union.} A failed call counts as the cautious outcome, as in
the program above. \emph{Rule floor.} A failed call leaves only the rules' verdict, but
never less than it: the model can only add findings, and an outage falls back to the rules
instead of to the safe outcome. Use the floor when holding everything during an outage
costs more than the rules alone miss, and say in the outcome that the model was absent.

\item[Evidence] The program is checked against recorded receipts only%
; FrameX has no measurement of its own. Figures from the hackathon are taken from result files committed in the submissions' repositories at a named commit; we did not rerun them, and figures that appear only in a README, report or pitch are left out. In an
invoice-payment agent (Apertus v1.5 8B, 500 synthetic invoices, 300 of which must not be
paid), the model alone paid 178 of the 300, code checks alone 89, and the strict union
none (95\,\% CI 0--1.3\,\%); it held 7 of 200 legitimate invoices, against 2 for the model
alone. On a set run blind before later checks were added, the figures were 337, 174 and
29 of 600. In redaction of personal data (8B, 300 synthetic documents, 1,860 entities),
recall was 38.7\,\% for rules, 57.7\,\% for the model and 78.5\,\% for the union, and
81\,\% of documents still leaked at least one entity: the union is stricter than each
source, not safe by itself.

\begin{figure}[hp]
\centering
\includegraphics[width=\linewidth,height=0.82\textheight,keepaspectratio]{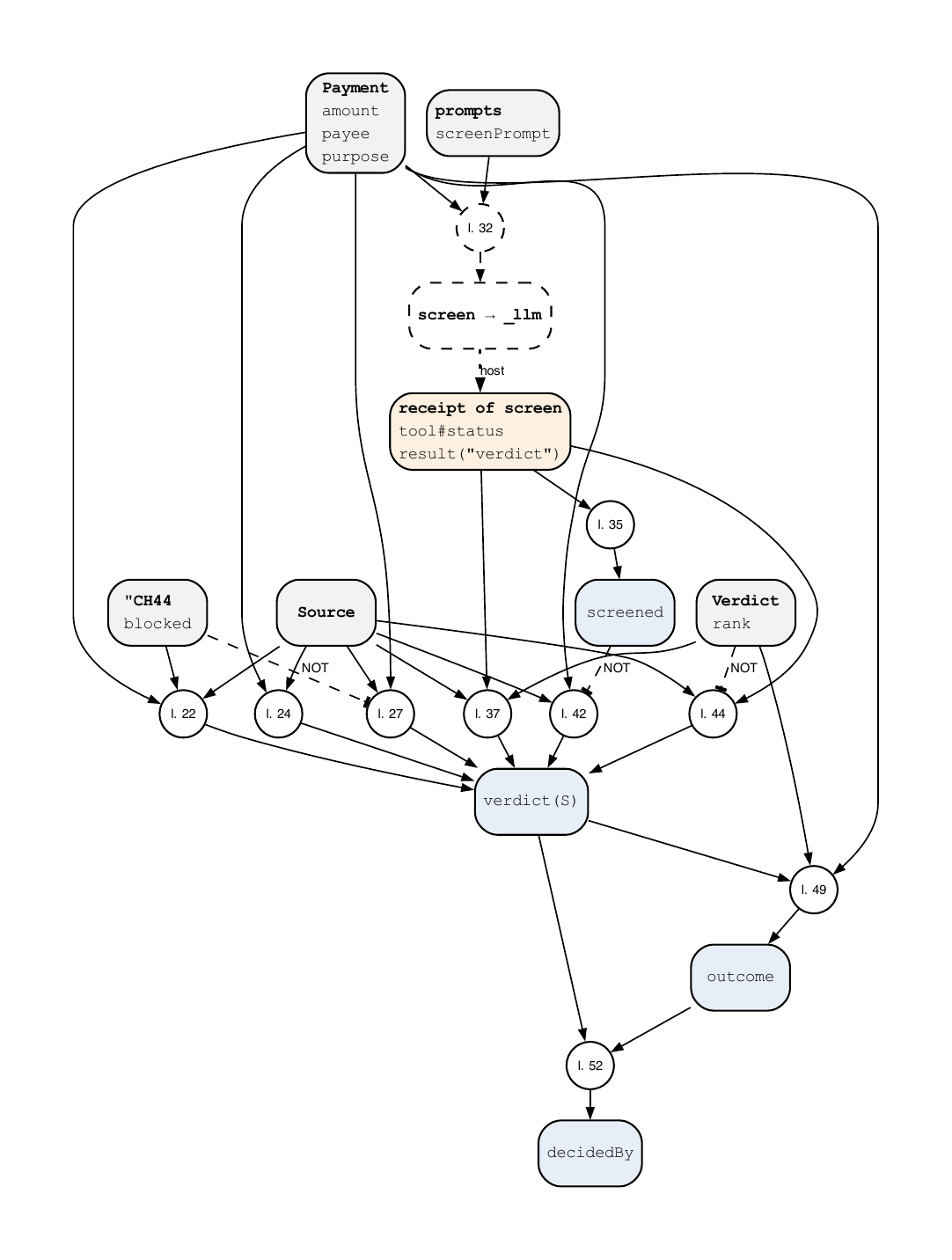}
\caption{Rule graph of \texttt{p13-failsafe.fx}, generated from the program. A circle is one rule; ``l.~23'' means that the rule starts in line 23 of the listing. Legend in \cref{tab:rglegend}.}
\label{fig:rg-failsafe}
\end{figure}
\end{patternform}

\section{Patterns in the reasoning itself}
\label{sec:reasoning}

\FloatBarrier
\subsection{Pattern 8: Model as Hypothesis Generator}
\label{pat:hypothesis}

\begin{patternform}
\item[Intent] Let the model propose what might be the case (a cause, a missing
document, an explanation) and let the rules test each proposal in a copy of the session
and ask for the evidence that decides between them. A hypothesis never becomes a fact.

\item[Also known as] Abduction with a model, diagnosis, candidate generation and
test.

\item[Motivation] A bank reconciliation shows a difference between the ledger and the
bank statement. A model suggests causes (a fee, a transposed amount, a payment booked
on the wrong day); each is plausible. The rules can test each: assumed in a copy, does
it explain the difference without violating a constraint? The surviving candidates
become questions, and only the answer decides.

\item[Applicability] Use it for diagnosis, reconciliation, investigation and
\emph{unknown} goals, where generating candidates needs world knowledge but testing
them needs exactness.

\item[Structure]\leavevmode
\begin{center}
\begin{tikzpicture}[bx]
\node[bxinst] (h1) at (0,0) {data:\\text};
\node[bxnote, below=0.6mm of h1] {observation};
\node[bxproc, right=of h1] (h2) {infer};
\node[bxmodel, above=of h2] (h2m) {model:\\gen};
\node[bxnote, above=0.3mm of h2m] {language model};
\draw[bxflow] (h2m)--(h2);
\draw[bxflow] (h1)--(h2);
\node[bxinst, right=of h2] (h3) {symbol};
\node[bxnote, below=0.6mm of h3] {candidates};
\draw[bxflow] (h2)--(h3);
\node[bxproc, right=of h3] (h4) {infer:\\deduce};
\node[bxmodel, above=of h4] (h4m) {model:\\sem};
\node[bxnote, above=0.3mm of h4m] {rules, in a copy};
\draw[bxflow] (h4m)--(h4);
\draw[bxflow] (h3)--(h4);
\node[bxinst, right=of h4] (h5) {symbol};
\node[bxnote, below=0.6mm of h5] {survivors};
\draw[bxflow] (h4)--(h5);
\node[bxactor, right=of h5] (h6) {};
\node[bxnote, below=0.6mm of h6] {person};
\draw[bxflow] (h5)--(h6);
\node[bxinst, right=of h6] (h7) {symbol};
\node[bxnote, below=0.6mm of h7] {deciding answer};
\draw[bxflow] (h6)--(h7);
\draw[bxreq] (h4.south) .. controls +(0,-10mm) and +(0,-10mm) .. node[pos=0.5, below, inner sep=1pt, font=\tiny]{request} (h2.south);
\end{tikzpicture}
\end{center}

\item[Participants] \emph{Model} (proposes candidates marked hypothetical);
\emph{Rules} (test each in a copy, derive evidence questions); \emph{Person} or a
source (answers the deciding question).

\item[Collaborations] Candidates arrive as receipts. In the full pattern each is previewed in a copy of the
session; rules keep those that derive every observation and violate nothing; the
evidence search names the one to four answers that would decide.

\item[Consequences] Hypotheses are cheap to generate and never pollute the knowledge
base. The search is bounded; an incomplete search is never a claim that no explanation
exists.

\item[Implementation] Ask the model for candidates as rows, keep each as a
hypothesis keyed by its row number, refute it by rule, and ask about the survivors only;
a refuted candidate never becomes a fact. The question itself goes to a person through a
flow (\cref{pat:steer}). From the bank reconciliation:
\pagebreak
\begin{lstlisting}[style=fxprog,basicstyle=\footnotesize\ttfamily,language=framex]
// Pattern 8, Model as Hypothesis Generator: the model proposes causes, the rules test
// each against what is known and ask about the survivors.
world open.
closed method refuted/1.

// The observation to explain: ledger and bank differ by 40 francs.
recon[difference -> 40;
    ledger -> "Ledger balance 12 340.00 CHF on 30 September; bank statement 12 300.00 CHF. Account fee 40.00 CHF charged monthly. Payment of 90.00 CHF to Muster AG booked on 30 September."].

proposePrompt:Prompt[text -> "List possible causes of the difference between ledger and bank statement. Answer as JSON {\"rows\": [{\"entry\": ..., \"amount\": a number}]}."].

// A request without arguments: the request object is propose itself.
external propose using _llm with proposePrompt[difference -> ?D; ledger -> ?L]
  <- recon[difference -> ?D; ledger -> ?L].

// Each row is a candidate cause, never a fact.
recon[cause(?I) -> ?E; amount(?I) -> ?A]
  <- propose[tool#status -> "succeeded"; row(?I, "entry") -> ?E;
        row(?I, "amount") -> ?A].

// Test: a cause whose amount differs from the difference cannot explain it.
recon[refuted(?I) -> true]
  <- recon[amount(?I) -> ?A; difference -> ?D] AND ?A != ?D.

recon[survives(?I) -> true]
  <- recon[cause(?I) -> ?E] AND NOT recon[refuted(?I) -> true].

// Ask: every surviving cause becomes a question for a person.
recon[askAbout(?I) -> ?E] <- recon[survives(?I) -> true; cause(?I) -> ?E].

?- recon[askAbout(?I) -> ?E].
?- recon[refuted(?I) -> true].
\end{lstlisting}
\Cref{fig:rg-hypothesis} draws the program as a rule graph.
This program tests every candidate with one rule, its amount. A fuller test assumes each
candidate in a copy of the session (\texttt{preview}) and keeps it only if it derives the
observation without violating a constraint, and the evidence search
(\texttt{\_evidence\_needs}) names the answers that would decide. The recorded run shows
why the test should be as strong as the domain allows: four of five candidates passed the
amount test, among them an ``unrecorded deposit'' whose direction contradicts the ledger.

\item[Library form] With \texttt{hypothesis.fx} (\cref{sec:libraries}) the program states
the observation, names the row field that a person is asked about (\texttt{hy\#label}) and
keeps its tests (\texttt{hy\#refuted}). The library keeps every field of every row as
\texttt{hy\#candidate(?I, field)}. It sends the observation as one text, so the model sees a
different request than in the program above:
\begin{lstlisting}[style=fxprog,basicstyle=\footnotesize\ttfamily,language=framex]
// Pattern 8 in library form: load hypothesis.fx first.
world open.

proposePrompt:Prompt[text -> "List possible causes of the difference between ledger and bank statement. Answer as JSON {\"rows\": [{\"entry\": ..., \"amount\": a number}]}."].

recon:hy#Observation[hy#prompt -> proposePrompt; hy#label -> "entry"; difference -> 40;
    hy#text -> "Ledger balance 12 340.00 CHF on 30 September; bank statement 12 300.00 CHF. Account fee 40.00 CHF charged monthly. Payment of 90.00 CHF to Muster AG booked on 30 September."].

// The test stays with the program: a cause whose amount differs cannot explain the difference.
recon[hy#refuted(?I) -> true]
  <- recon[hy#candidate(?I, "amount") -> ?A; difference -> ?D] AND ?A != ?D.

?- recon[hy#askAbout(?I) -> ?E].
\end{lstlisting}

\item[Known uses] Bank reconciliation with model-proposed causes; the evidence-needs
search over declared open questions; the candidate checker's \texttt{facts} mode.
\knownlit{hypothesis}

\item[Related patterns] Critique and Repair (\cref{pat:repair}): a repair suggestion
is a hypothesis that is a correction. Rules Steer the Model (\cref{pat:steer}) asks the
deciding question.
\item[Evidence] One demonstration run%
; no measurement yet. The
bank reconciliation of \cref{tab:compositions} runs the pattern.

\begin{figure}[hp]
\centering
\includegraphics[width=\linewidth,height=0.82\textheight,keepaspectratio]{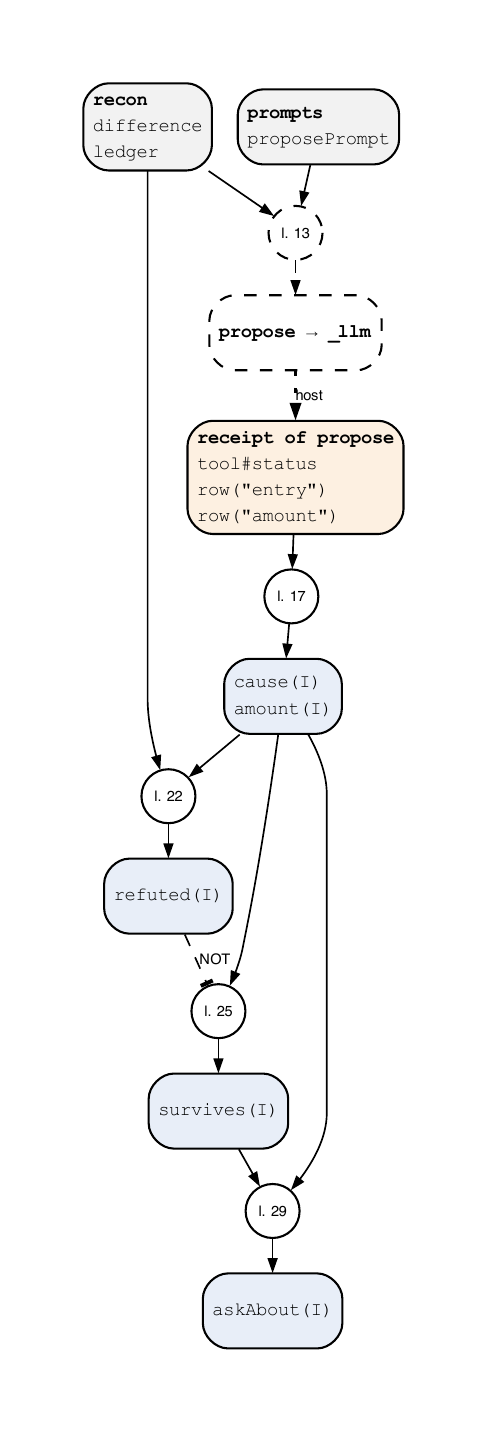}
\caption{Rule graph of \texttt{p08-hypothesis.fx}, generated from the program. A circle is one rule; ``l.~23'' means that the rule starts in line 23 of the listing. Legend in \cref{tab:rglegend}.}
\label{fig:rg-hypothesis}
\end{figure}
\end{patternform}

\FloatBarrier
\subsection{Pattern 9: Model as Critic of the Rules}
\label{pat:critic}

\begin{patternform}
\item[Intent] Let the model review what the rules hold or conclude (a policy change,
a rule against its source text, a surprising decision); its critique opens a review and
never changes a rule or a conclusion.

\item[Also known as] Policy critic, plausibility review, two-channel cross-check.

\item[Motivation] A policy change passes every regression case and still changes the
decision for a case nobody expected. The change's author wrote a rationale; a model can
read the changed decisions against it and flag what looks inconsistent. Its remarks
may be wrong, so they open a review that a person disposes of.

\item[Applicability] Use it where checks are incomplete: semantic faults, fidelity of
a rule to the text it cites, plausibility of changed decisions before a release.

\item[Structure]\leavevmode
\begin{center}
\begin{tikzpicture}[bx]
\node[bxinst] (i1) at (0,0) {data:\\text};
\node[bxnote, below=0.6mm of i1] {rationale, review};
\node[bxproc, right=of i1] (i2) {infer};
\node[bxmodel, above=of i2] (i2m) {model:\\gen};
\node[bxnote, above=0.3mm of i2m] {critic model};
\draw[bxflow] (i2m)--(i2);
\draw[bxflow] (i1)--(i2);
\node[bxinst, right=of i2] (i3) {data:\\text};
\node[bxnote, below=0.6mm of i3] {remarks};
\draw[bxflow] (i2)--(i3);
\node[bxproc, right=of i3] (i4) {trans-\\form};
\node[bxnote, below=0.6mm of i4] {quote check};
\draw[bxflow] (i3)--(i4);
\node[bxinst, right=of i4] (i5) {symbol};
\node[bxnote, below=0.6mm of i5] {grounded remarks};
\draw[bxflow] (i4)--(i5);
\node[bxactor, right=of i5] (i6) {};
\node[bxnote, below=0.6mm of i6] {reviewer};
\draw[bxflow] (i5)--(i6);
\node[bxinst, right=of i6] (i7) {symbol};
\node[bxnote, below=0.6mm of i7] {disposed};
\draw[bxflow] (i6)--(i7);
\node[bxproc, right=of i7] (i8) {infer:\\deduce};
\node[bxmodel, above=of i8] (i8m) {model:\\sem};
\node[bxnote, above=0.3mm of i8m] {release rules};
\draw[bxflow] (i8m)--(i8);
\draw[bxflow] (i7)--(i8);
\node[bxinst, right=of i8] (i9) {symbol};
\node[bxnote, below=0.6mm of i9] {approved};
\draw[bxflow] (i8)--(i9);
\draw[bxreq] (i8.south) .. controls +(0,-10mm) and +(0,-10mm) .. node[pos=0.5, below, inner sep=1pt, font=\tiny]{request} (i2.south);
\end{tikzpicture}
\end{center}

\item[Participants] \emph{Model} (critic); \emph{Rules} (keep a remark only if it
names a case and quotes the rationale verbatim; flag declared invariants without a
model; gate approval); \emph{Reviewer} (disposes of every finding).

\item[Collaborations] A sealed before-and-after review of the policy is the critic's
input; grounded remarks become findings; approval is derived only after every finding
is disposed of. A corrected rule proposed by the critic is checked like any draft.

\item[Consequences] The critic adds recall where checks are incomplete; its false
remarks cost reviewer time, so grounding requirements filter them. It never weakens a
complete check.

\item[Implementation] The model receives the rationale and the sealed review and answers
with rows, as in the Sensor, with the rationale as \texttt{groundIn}, the text its quotes
must come from. A remark counts only if its quote was found and it names a changed
decision. Approval needs a completed critique and a disposal of every remark; a failed
call blocks it with the host's reason:
\begin{lstlisting}[style=fxprog,basicstyle=\footnotesize\ttfamily,language=framex]
// Pattern 9, Model as Critic of the Rules: grounded remarks open a review; approval
// needs a completed critique and a reviewer's disposal of every remark.
world open.
closed class ChangedDecision. closed method disposed/1. closed slot hasOpen.
closed slot critiqued.

// A rule change with its rationale and the decisions it changes.
"c7":ChangedDecision. "c9":ChangedDecision.
chg:Change[reviewed -> true;
    rationale -> "The amendment lowers the language requirement for applicants with a Bachelor taught in English. Applicants with other degrees keep the C1 requirement.";
    review -> "c7: Bachelor taught in German, no C1; before not admitted, after admitted. c9: Bachelor taught in English, no C1; before not admitted, after admitted."].

critiquePrompt:Prompt[text -> "Read the changed decisions against the rationale. Report one row {case, quote, remark} per decision that contradicts the rationale; quote the rationale verbatim. Answer as JSON {\"rows\": [...]}."].

// The quote of every remark must occur in the rationale.
external critique(?C) using _llm with critiquePrompt[rationale -> ?Ra;
        review -> ?Rv; groundIn -> ?Ra]
  <- ?C:Change[rationale -> ?Ra; review -> ?Rv].

// A remark counts only with a found quote and a case that the change really changed.
?C[remark(?I) -> ?K]
  <- ?R[tool#name -> "_llm"; tool#arg1 -> ?C; row(?I, "case") -> ?K;
        row(?I, "quoteFound") -> true] AND ?K:ChangedDecision.

?C[critiqued -> true]
  <- ?R[tool#name -> "_llm"; tool#arg1 -> ?C; tool#status -> "succeeded"].

?C[blocked -> ?E]
  <- ?R[tool#name -> "_llm"; tool#arg1 -> ?C; tool#status -> "failed"; error -> ?E].

// A reviewer disposes of a remark by stating chg[disposed(1) -> true].
?C[hasOpen -> true] <- ?C[remark(?I) -> ?K] AND NOT ?C[disposed(?I) -> true].

?C[approved -> true]
  <- ?C:Change[reviewed -> true; critiqued -> true] AND NOT ?C[hasOpen -> true].

?- chg[remark(?I) -> ?K].
?- chg[approved -> ?A].
\end{lstlisting}
\Cref{fig:rg-critic} draws the program as a rule graph.

\item[Library form] With \texttt{critic.fx} (\cref{sec:libraries}) the program states the
change, its rationale, the review and the cases it changed; a reviewer's disposal is the
fact \texttt{chg[cr\#disposed(1) -> true]}:
\begin{lstlisting}[style=fxprog,basicstyle=\footnotesize\ttfamily,language=framex]
// Pattern 9 in library form: load critic.fx first.
world open.

critiquePrompt:Prompt[text -> "Read the changed decisions against the rationale. Report one row {case, quote, remark} per decision that contradicts the rationale; quote the rationale verbatim. Answer as JSON {\"rows\": [...]}."].

chg:cr#Change[cr#reviewed -> true; cr#prompt -> critiquePrompt;
    cr#changed -> "c7"; cr#changed -> "c9";
    cr#rationale -> "The amendment lowers the language requirement for applicants with a Bachelor taught in English. Applicants with other degrees keep the C1 requirement.";
    cr#review -> "c7: Bachelor taught in German, no C1; before not admitted, after admitted. c9: Bachelor taught in English, no C1; before not admitted, after admitted."].

?- chg[cr#remark(?I) -> ?K].
?- chg[cr#approved -> ?A].
\end{lstlisting}

\pagebreak
\item[Known uses] The policy critic in a university administration case with seeded
defects.
\knownlit{critic}

\item[Related patterns] Reverse of Critique and Repair (\cref{pat:repair}); input to
Model as Author (\cref{pat:author}) when it proposes a correction.

\item[Evidence] In a pilot study with \texttt{gpt-oss:20b}, 20 changes over 10 rule
domains were each reviewed three times. In the 10 changes with one seeded decision that
contradicts the rationale, the critic found it in 29 of 30 reviews; in the 10 clean
changes it made no remark in 29 of 30. All 34 of its remarks quoted the rationale
verbatim, so the quote check removed nothing. The changes were short and each defect was
stated plainly; harder defects and longer reviews are not yet measured. In the checking
study, model reviews found contradictory rules and unknown facts almost always, but a
missing rule in only 3 of 10 rule reviews at 64 rules and in none of the code or prompt
reviews, so a critic belongs on top of the checks.

\begin{figure}[hp]
\centering
\includegraphics[width=\linewidth,height=0.82\textheight,keepaspectratio]{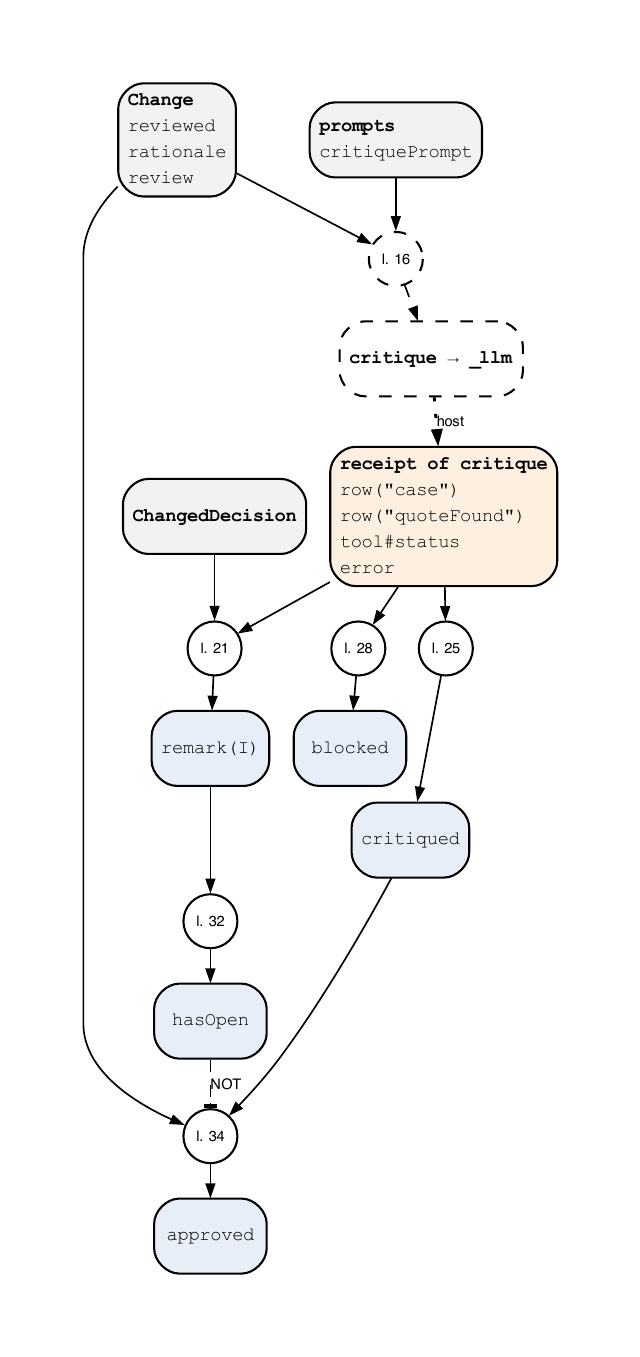}
\caption{Rule graph of \texttt{p09-critic.fx}, generated from the program. A circle is one rule; ``l.~23'' means that the rule starts in line 23 of the listing. Legend in \cref{tab:rglegend}.}
\label{fig:rg-critic}
\end{figure}
\end{patternform}

\FloatBarrier
\subsection{Pattern 10: Rules Teach the Model}
\label{pat:teach}

\begin{patternform}
\item[Intent] Let the rules supply labels, rule-selected worked examples and grades,
with proofs, so that knowledge flows into the model's context or weights; the rules stay
in production as the independent check.

\item[Also known as] Rules as labels and graders, rule-selected examples, rule-informed
preference data.

\item[Motivation] A model that must answer admission questions can be shown worked
cases. Which cases, and whether its answers are right, the rules know: they label
generated cases with decisions and proofs, select the examples closest to the current
case, and grade the answers.

\item[Applicability] Use it when a model must work where the rules are not called (a
fast path, an edge device) or must learn the vocabulary of a domain, and when training
or prompt examples need trustworthy labels.

\item[Structure]\leavevmode
\begin{center}
\begin{tikzpicture}[bx]
\node[bxinst] (j1) at (0,0) {symbol};
\node[bxnote, below=0.6mm of j1] {cases};
\node[bxproc, right=of j1] (j2) {infer:\\deduce};
\node[bxmodel, above=of j2] (j2m) {model:\\sem};
\node[bxnote, above=0.3mm of j2m] {rules};
\draw[bxflow] (j2m)--(j2);
\draw[bxflow] (j1)--(j2);
\node[bxinst, right=of j2] (j3) {symbol};
\node[bxnote, below=0.6mm of j3] {labels, proofs};
\draw[bxflow] (j2)--(j3);
\node[bxproc, right=of j3] (j4) {generate:\\train};
\draw[bxflow] (j3)--(j4);
\node[bxmodel, right=of j4] (j5) {model:\\gen};
\node[bxnote, below=0.6mm of j5] {taught model};
\draw[bxflow] (j4)--(j5);
\node[bxproc, right=of j5] (j6) {infer};
\draw[bxflow] (j5)--(j6);
\node[bxinst, right=of j6] (j7) {symbol};
\node[bxnote, below=0.6mm of j7] {answer};
\draw[bxflow] (j6)--(j7);
\node[bxproc, right=of j7] (j8) {infer:\\deduce};
\node[bxmodel, above=of j8] (j8m) {model:\\sem};
\node[bxnote, above=0.3mm of j8m] {rules};
\draw[bxflow] (j8m)--(j8);
\draw[bxflow] (j7)--(j8);
\node[bxinst, right=of j8] (j9) {symbol};
\node[bxnote, below=0.6mm of j9] {grade};
\draw[bxflow] (j8)--(j9);
\node[bxinst, above=of j4] (j4i) {data:\\text}; \draw[bxflow] (j4i)--(j4);
\end{tikzpicture}
\end{center}

\item[Participants] \emph{Rules} (label, select, grade); \emph{Model}; \emph{Trainer}
outside the rule engine.

\item[Collaborations] The rules label every generated case with its decision and the
missing requirements; the model answers each case; a rule grades the answer against the
label and records each gap the answer missed. Labels and grades, with their proofs, leave
the engine as training or evaluation data, and after training the same rules grade the
taught model.

\item[Consequences] The labels are as good as the rules and come with proofs; what the
model learned is never the authority.

\item[Implementation] Grade a model's answer against the rules' decision on generated
cases with complete evidence, and record each gap the answer missed. The model
answers each case; its receipt becomes the answer that the rules grade:
\lstinputlisting[style=fxprog,basicstyle=\footnotesize\ttfamily,language=framex]{programs/p10-teach.fx}
\Cref{fig:rg-teach} draws the program as a rule graph.
The training step of the Structure diagram happens outside the engine and is not part of
the program.

\item[Library form] With \texttt{teach.fx} (\cref{sec:libraries}) the program states the
cases and keeps its labelling rules (\texttt{te\#label}, \texttt{te\#gap}); the library asks
the model and grades each answer against the label:
\begin{lstlisting}[style=fxprog,basicstyle=\footnotesize\ttfamily,language=framex]
// Pattern 10 in library form: load teach.fx first.
world open.
closed slot hasBachelor. closed slot hasC1.

answerPrompt:Prompt[text -> "Requirements: a recognised Bachelor and English C1. Is the applicant admitted? Answer as JSON {\"decision\": admitted or not admitted, \"rows\": [{\"gap\": englishC1 or bachelor}] per missing requirement}."].

k1:te#Case[hasBachelor -> true; hasC1 -> true; te#prompt -> answerPrompt;
    te#description -> "Recognised Bachelor; English C1 certificate on file."].
k2:te#Case[hasBachelor -> true; te#prompt -> answerPrompt;
    te#description -> "Recognised Bachelor; no language record."].

// The labels come from the rules; this stays with the program.
?C[te#label -> "admitted"] <- ?C:te#Case[hasBachelor -> true; hasC1 -> true].
?C[te#label -> "not admitted"] <- ?C:te#Case AND NOT ?C[hasC1 -> true].
?C[te#label -> "not admitted"] <- ?C:te#Case AND NOT ?C[hasBachelor -> true].
?C[te#gap -> "englishC1"] <- ?C:te#Case AND NOT ?C[hasC1 -> true].
?C[te#gap -> "bachelor"] <- ?C:te#Case AND NOT ?C[hasBachelor -> true].

?- ?A:te#Answer[te#case -> ?C; te#decision -> ?D; te#right -> ?R].
?- ?A[te#missedGap -> ?G].
\end{lstlisting}

\item[Known uses] Labels, rule-selected examples and grading for 144 generated
admission cases (built, measures fixed, no model run yet); rules that govern a small
network's training loop.
\knownlit{teach}

\item[Related patterns] Rules Steer the Model (\cref{pat:steer}) for selecting context;
Model as Sensor (\cref{pat:sensor}) for the judge that may be taught.
\item[Evidence] One demonstration run%
; no measurement yet.

\begin{figure}[hp]
\centering
\includegraphics[width=\linewidth,height=0.82\textheight,keepaspectratio]{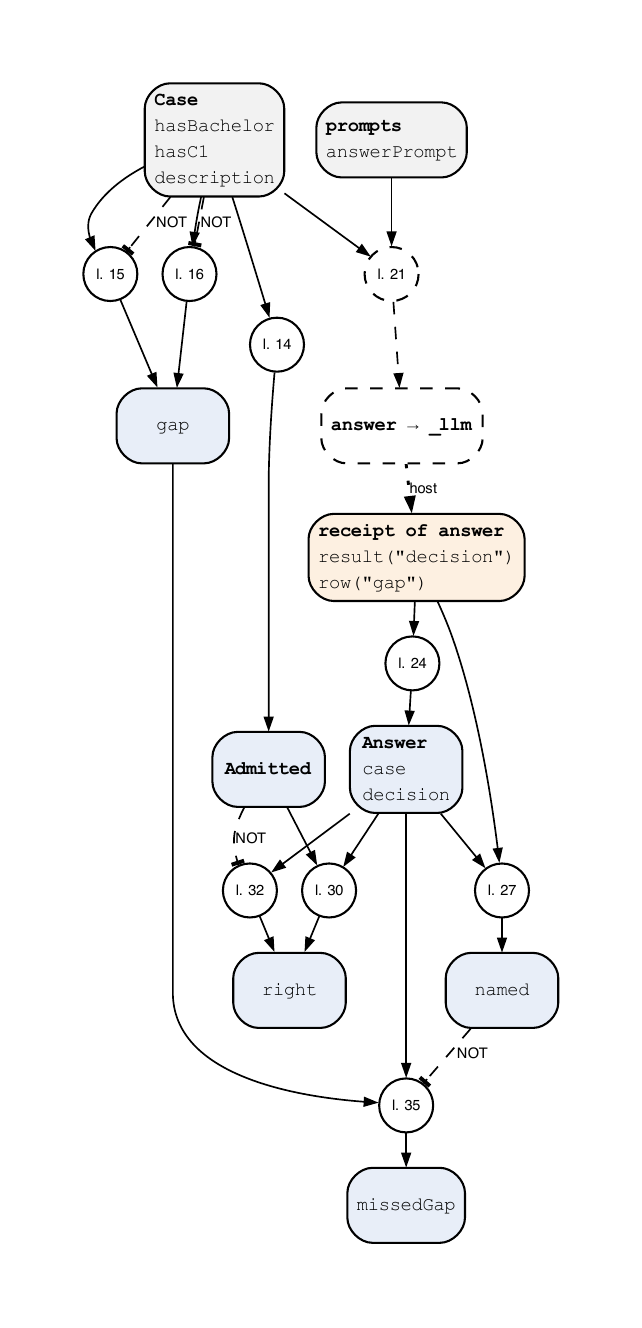}
\caption{Rule graph of \texttt{p10-teach.fx}, generated from the program. A circle is one rule; ``l.~23'' means that the rule starts in line 23 of the listing. Legend in \cref{tab:rglegend}.}
\label{fig:rg-teach}
\end{figure}
\end{patternform}

\section{Patterns over time}
\label{sec:time}

\FloatBarrier
\subsection{Pattern 11: Rule-Governed Memory}
\label{pat:memory}

\begin{patternform}
\item[Intent] Keep what an assistant learns across conversations as statements that
rules admit, supersede and withdraw, and let rules decide which
remembered statements enter the next prompt.

\item[Also known as] Agent memory, long-term memory, memory handler, belief revision
with a model as source.

\item[Motivation] An advisor learns in March that a student prefers German and in May
that English is fine; in June the student asks it to forget a message. A model that keeps
its own notes either keeps both values or overwrites one, cannot say which holds or why,
and its notes quietly become records: a grade mentioned in a chat sits next to the grade
on the transcript. A single reading (\cref{pat:sensor}) decides whether a statement holds
now; memory must decide whether it still holds later. The pattern keeps every remembered
statement with its utterance and quote, lets a slot policy decide which sources may
establish which facts, lets a newer admitted statement supersede an older one without
deleting it, and recomputes what
rests on an utterance when that utterance is forgotten. Rules choose what the model sees next, with a similarity score as one input, and the
model cites the statements it used.

\item[Applicability] Use it when
\begin{itemize}
\item an assistant serves the same person or case across sessions;
\item what is known changes over time, and the history of a value matters;
\item sources differ in what they may establish (a chat may set a preference, only a
certificate a qualification);
\item a request to forget must take effect on everything that rested on the forgotten
input.
\end{itemize}
For a single session a Sensor suffices. Retrieval by similarity over documents is the
task of a retriever outside the engine; its results enter as receipts like any other.

\item[Structure]\leavevmode
\begin{center}
\begin{tikzpicture}[bx]
\node[bxinst] (k1) at (0,0) {data:\\text};
\node[bxnote, below=0.6mm of k1] {utterance};
\node[bxproc, right=of k1] (k2) {infer};
\node[bxmodel, above=of k2] (k2m) {model:\\gen};
\node[bxnote, above=0.3mm of k2m] {language model};
\draw[bxflow] (k2m)--(k2);
\draw[bxflow] (k1)--(k2);
\node[bxinst, right=of k2] (k3) {data:\\text};
\node[bxnote, below=0.6mm of k3] {reply};
\draw[bxflow] (k2)--(k3);
\node[bxproc, right=of k3] (k4) {trans-\\form};
\node[bxnote, below=0.6mm of k4] {quote check};
\draw[bxflow] (k3)--(k4);
\node[bxinst, right=of k4] (k5) {symbol};
\node[bxnote, below=0.6mm of k5] {proposed};
\draw[bxflow] (k4)--(k5);
\node[bxproc, right=of k5] (k6) {infer:\\deduce};
\node[bxmodel, above=of k6] (k6m) {model:\\sem};
\node[bxnote, above=0.3mm of k6m] {slot policy};
\draw[bxflow] (k6m)--(k6);
\draw[bxflow] (k5)--(k6);
\node[bxinst, right=of k6] (k7) {symbol};
\node[bxnote, below=0.6mm of k7] {current};
\draw[bxflow] (k6)--(k7);
\node[bxproc, right=of k7] (k8) {infer};
\node[bxmodel, above=of k8] (k8m) {model:\\gen};
\node[bxnote, above=0.3mm of k8m] {answering model};
\draw[bxflow] (k8m)--(k8);
\draw[bxflow] (k7)--(k8);
\node[bxinst, right=of k8] (k9) {symbol};
\node[bxnote, below=0.6mm of k9] {cited ids};
\draw[bxflow] (k8)--(k9);
\draw[bxflow] (k9.south) -- ++(0,-6mm) -- node[below, font=\tiny]{citations checked by the rules: sent and current} ([yshift=-6mm]k6.south) -- (k6.south);
\draw[bxreq] (k6.north) to[bend left=25] node[above, inner sep=1pt, font=\tiny]{request} (k8.north west);
\draw[bxreq] (k6.north) to[bend right=25] node[above, inner sep=1pt, font=\tiny]{request} (k2.north east);
\end{tikzpicture}
\end{center}

\item[Participants] \emph{Model}: proposes statements with quotes from an utterance, and
later cites the statements it used. \emph{Rules}: slot policy, admission, supersession,
withdrawal, and the selection of statements for a prompt. \emph{Host}: stores
utterances and receipts, measures quotes, removes the content of a forgotten utterance.
\emph{Person}: states, corrects, asks to forget.

\item[Collaborations] Each utterance triggers a Sensor request; the proposed statements
return as receipts; rules admit them per slot; supersession follows the order of
admission, and the current value is a query. A request to forget marks the utterance; in
the next derivation every statement that rested on it loses its acceptance, and an older
statement can become current again. Before a model call, a steering rule
(\cref{pat:steer}) selects the current statements the step needs; the reply cites their
identifiers, and a rule checks that every cited statement is current.

\item[Consequences] Every remembered value has an utterance, a quote, a date and a
history. Forgetting is exact without manual clean-up, because acceptance is derived and
recomputed. A chat cannot establish what its source may not, and a derived fact such as
an admission is never written to memory. The cost is one Sensor call per utterance and a
store that grows, since only a request to forget removes content. A statement admitted
wrongly holds until it is superseded or forgotten, and a conflict between sources needs a
person (\cref{pat:competence}).

\item[Implementation] Each utterance triggers a Sensor request with the utterance as
\texttt{groundIn}. Row \texttt{?I} of the answer for utterance \texttt{?U} becomes one
statement object \texttt{st(?U, ?I)}, a function term in the rule head, kept only with a
found quote. For recall, the rules mark the statements one step may see with
\texttt{selectedFor}; the request names the step with \texttt{contextKey}, and the host
sends exactly those statements and records their identifiers in the receipt
(\texttt{tool\#sent}). The reply cites identifiers, and rules accept it only if every
cited statement was sent and is current. \texttt{tool\#order} makes every
\texttt{remember} run before the answer, so the selection is made over all statements:
\lstinputlisting[style=fxprog,basicstyle=\footnotesize\ttfamily,language=framex]{programs/p11-memory.fx}
\Cref{fig:rg-memory} draws the program as a rule graph.
Forgetting \texttt{u2} withdraws its statements, \texttt{st(u1, 1)} with
\emph{German} becomes current again, and \texttt{\_forgetAnswers} removes every
recorded model answer whose prompt or answer holds the utterance, so a replay cannot bring
it back. A similarity score from a retriever can be one more condition of the selection
rule.

\item[Library form] With the library \texttt{memory.fx} (\cref{sec:libraries}) loaded
first, the program states only what each kind of source may establish, the utterances,
what a step needs, and the question. Selection is a fact (\texttt{mem\#Step}), and a
request to forget is the fact \texttt{u2[mem\#forgotten -> true]}:
\lstinputlisting[style=fxprog,basicstyle=\footnotesize\ttfamily,language=framex]{programs/library/p11-memory-lib.fx}
The library derives the statements, their admission and supersession, the context of
each step, the answer and every refused citation, and it removes forgotten utterances
from the answer store.

\item[Known uses] The program above, which has both halves; DreamRAG lite among the
FrameX applications, the memory handler: proposal from an utterance, admission per slot,
supersession, conflict and withdrawal. Memory claims with quotes in FrameX-Agents.
\knownlit{memory}

\item[Related patterns] Storing is Model as Sensor (\cref{pat:sensor}) applied to
utterances over time; recall is Rules Steer the Model (\cref{pat:steer}); conflicts go to
Division by Competence (\cref{pat:competence}); a reply that cites a superseded statement
is repaired by Critique and Repair (\cref{pat:repair}).

\item[Evidence] One demonstration run%
; no measurement yet.

\begin{figure}[hp]
\centering
\includegraphics[width=\linewidth,height=0.82\textheight,keepaspectratio]{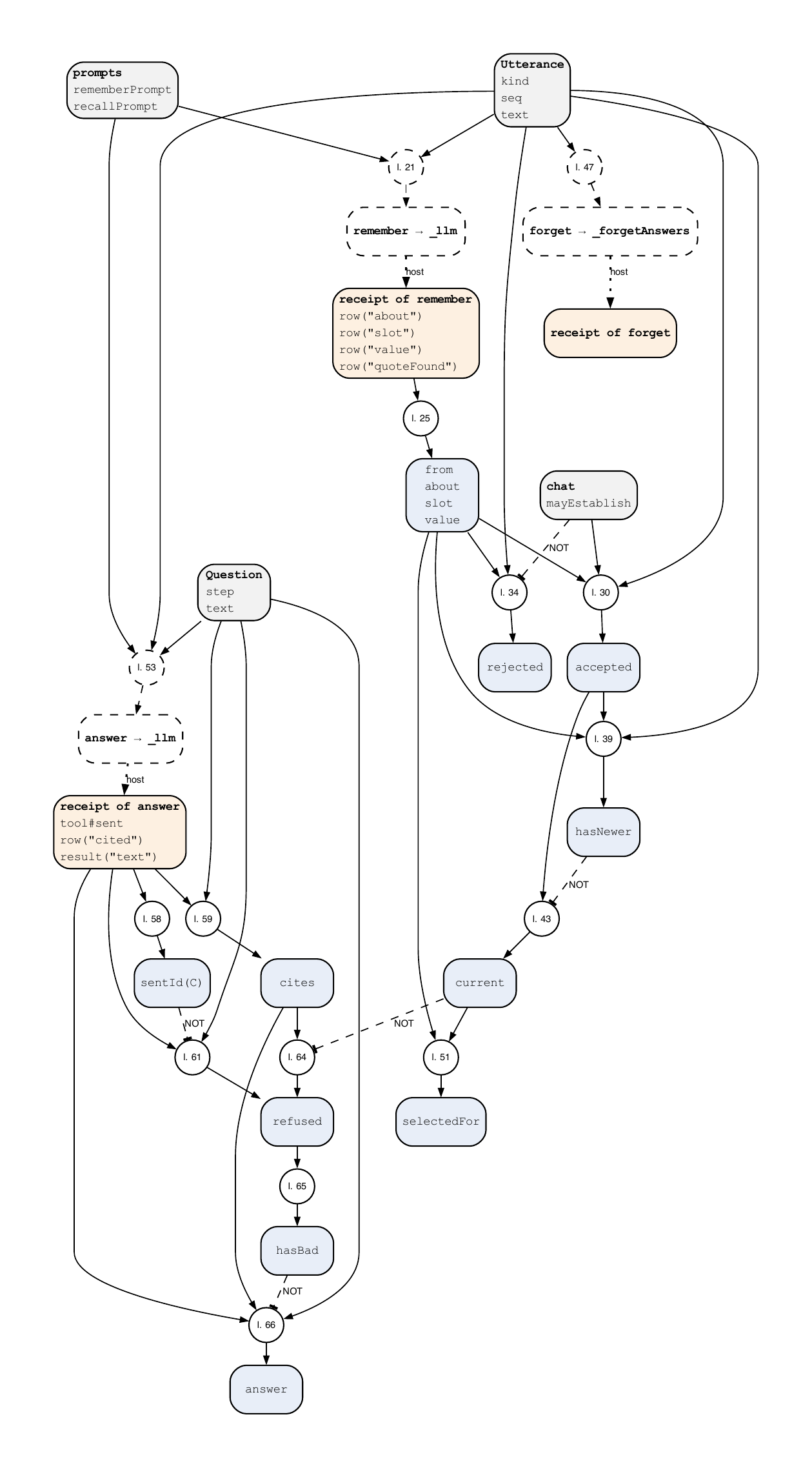}
\caption{Rule graph of \texttt{p11-memory.fx}, generated from the program. A circle is one rule; ``l.~23'' means that the rule starts in line 23 of the listing. Legend in \cref{tab:rglegend}.}
\label{fig:rg-memory}
\end{figure}
\end{patternform}
\FloatBarrier

\section{Compositions}
\label{sec:compositions}

Real systems combine patterns. \Cref{tab:compositions} lists the compositions in the
applications from which the catalogue was drawn. Three combinations recur: a flow that
steers a model and uses a judge as sensor, with repair and hand-over
(5\,+\,1\,+\,6\,+\,7); an agent whose proposals are admitted by rules that also steer
its context (4\,+\,5\,+\,6); and an authoring loop in which a checker's message drives
the repair (2\,+\,6).

Rule-Governed Memory (\cref{pat:memory}) is itself built from two of the other
patterns: storing is the Sensor, recall is Rules Steer the Model.

\begin{table}[ht]
\centering\small
\caption{Patterns in the applications from which the catalogue was drawn.}
\label{tab:compositions}
\begin{tabularx}{\linewidth}{@{}llX@{}}
\toprule
Application & Patterns & How \\
\midrule
Assignment review & 1, 3, 6 & quote-checked extraction; chat model calls the rules; completeness check \\
Interview, suitability, triage & 5, 1, 6, 7 & flow steers; judge as sensor; format repair; \emph{unclear} to a person \\
Campervan search & 1, 5 & one classification per listing, requested by rule \\
Allocation & 4, 5, 1 & bounded recommendation; point-in-time evidence; grader verdicts \\
Agent framework & 4, 5, 6, 1 & admitted actions; derived phase and context; stalled repair stops; memory claims \\
Personal memory (DreamRAG lite) & 11 (storing) & admission, supersession, conflict and withdrawal of memories as rules; this application has no recall by rules, the program of pattern 11 has \\
Rule drafting, decision tables & 2, 6 & model drafts; the checker and a bounded repair decide \\
Bank reconciliation & 8, 5 & model proposes causes; rules test them and ask for the deciding date \\
Policy critic & 9 & critic reads a sealed policy review; grounded findings gate approval \\
Model teaching & 10 & rules label cases, select examples and grade answers \\
\bottomrule
\end{tabularx}
\end{table}

\section{Failure patterns}
\label{sec:failures}

Five couplings recur in working systems and break an invariant of
\cref{sec:foundations}. Each looks like one of the patterns above, which is why it
survives review. We name them so that a design review can look for them; each entry says
what it looks like, what it breaks and which pattern replaces it. Most of them were
named in the jury assessments of the hackathon of \cref{pat:select,pat:failsafe}; the
first is the one that Critique and Repair already excludes by name.

\begin{description}[leftmargin=1.2em,font=\normalfont\bfseries\small]
\item[Silent Rule Override] Code outside the rules changes a verdict before the rules
see it: a regular expression in the host that turns a judge's \emph{no} into \emph{yes}, a
post-processing step that corrects a value. The outcome may be right, but the proof does
not contain the step that decided it, and nobody can tell the correction from the model.
It breaks invariant~3, and Critique and Repair (\cref{pat:repair}) excludes it by name.
In a comparison system built for the hackathon, checked in its code, a host function rewrote the judge's
reply before it became a receipt: \emph{yes} became \emph{no} when a pattern found the
answer vague, which only tightens; but \emph{unclear} was settled by the pattern in either
direction, and an unreachable model was recorded as \emph{yes}. Only a note in the reason
text was left, and the rules saw the corrected verdict alone. Instead, write the
correction as a rule that reads the receipt, so that
\texttt{explain} shows it; if the correction can only make the outcome stricter, it is a
source in a Fail-Safe Union (\cref{pat:failsafe}).
\item[Circular Evaluation] The cases or the gold labels that measure a coupling come from
the model, the prompt or the developer that is being measured, or the criteria are
chosen after the results are seen. The numbers then measure agreement with oneself. It
breaks the claim behind every \emph{Evidence} heading. Instead, fix the cases, the truths
and the criteria before the first model call, and derive labels by rules or by
construction: an injected fault, a value read from an exact source (Rules Teach the
Model, \cref{pat:teach}, and C3 in \cref{sec:related}).
\item[Self-Judge] The model that wrote an answer, or a model of the same family with the
same prompt, judges whether the answer is right. Its errors and its judgement fail
together, and self-correction without an external signal does not help
\parencite{huang2024cannot}. It breaks the independence that Several Readers, One Rule
(\cref{pat:sensor}) and the Read-Back Check (\cref{pat:translator}) require. Instead, let
a rule check what can be checked, and use a different model or a different prompt for
the rest (\cref{pat:repair,pat:critic}).
\item[Prose Policy] The policy that decides exists only as sentences in a prompt. Nothing
can test it, a change cannot be reviewed against cases, and an answer cannot show which
sentence it followed. It breaks invariant~3 and leaves the guide's first question
unanswered (\cref{sec:guide}). Instead, turn the policy into rules with Model as Author
(\cref{pat:author}) and let Rules Steer the Model (\cref{pat:steer}) put into the prompt
only what one step needs.
\item[Decorative Check] A check is computed and recorded, but nothing depends on it: a
\texttt{quoteFound} that no rule reads, a confidence that is logged, a validation whose
failure still lets the action through. It looks like evidence in a review and changes no
outcome. It breaks invariant~4. Instead, let an acceptance rule read every check, and
look for receipt fields without an outgoing arrow in the rule graph
(\cref{sec:rglegend}).
\end{description}

\section{Related catalogues}
\label{sec:related}

Two bodies of work describe couplings of learning and reasoning. Neuro-symbolic research
classifies how a learned and a symbolic component are combined, from Kautz's six types
\parencite{kautz2022third} to the boxology of \textcite{vanharmelen2019boxology}, its
modular taxonomy \parencite{vanbekkum2021modular} and its extension to language models
\parencite{deboer2025patterns}, and from the roadmap that unifies language models and
knowledge graphs \parencite{pan2024roadmap} to recent surveys
\parencite{garcez2023third,yang2025nesy,rani2026nesyllm}. Software engineering for
model-based systems collects architectural patterns for agents
\parencite{liu2025agentpatterns}, practitioners' workflow patterns
\parencite{schluntz2024agents}, guardrail toolkits \parencite{rebedea2023nemo} and design
patterns for machine-learning applications \parencite{washizaki2020sep4mla,washizaki2022sepml}.
Between them lie systems in which a model translates for a solver
\parencite{pan2023logiclm,olausson2023linc,ye2023satlm} or generates candidates for sound
critics \parencite{kambhampati2024llmmodulo}.

\subsection{Mapping}

\Cref{tab:mapping} places each pattern in these classifications. Kautz's types name who
calls whom; the boxology names the flow of data and symbols; the roadmap names which side
enhances which; the agent literature names the mechanism.

\begin{table}[ht]
\centering\small
\caption{The thirteen patterns in published classifications. ``--'' means no counterpart.}
\label{tab:mapping}
\begin{tabularx}{\linewidth}{@{}>{\raggedright\arraybackslash}p{2.6cm}>{\raggedright\arraybackslash}p{2.7cm}>{\raggedright\arraybackslash}p{2.8cm}>{\raggedright\arraybackslash}X>{\raggedright\arraybackslash}p{1.5cm}@{}}
\toprule
Pattern & Kautz type & Boxology, modular patterns & Language-model and agent literature & Status \\
\midrule
1 Model as Sensor & Neuro\,|\,Symbolic & learning with symbolic output (4.1) & KG construction \parencite{pan2024roadmap} & known \\
2 Model as Author & -- & engineered model (1c) & model acquisition \parencite{kambhampati2024llmmodulo} & known in part \\
3 Model as Translator & Symbolic[Neuro] & LLM-augmented KG \parencite{deboer2025patterns} & Logic-LM, LINC, SatLM & known \\
4 Model Proposes, Rules Dispose & Symbolic[Neuro] & -- & guardrails, rule safeguards, hard critics & known \\
5 Rules Steer the Model & Symbolic[Neuro] & meta-reasoning for control (4.8), training only & routing, dialogue rails & new as stated \\
6 Critique and Repair & -- & -- & solver feedback, back-prompting & known \\
7 Division by Competence & inverse of Neuro[Symbolic] & -- & routing; decompiler & new in part \\
8 Model as Hypothesis Generator & -- & -- & abduction and hypothesis search & new in part \\
9 Model as Critic of the Rules & -- & -- & soft critics review candidates only & new \\
10 Rules Teach the Model & Neuro:Symbolic\-$\to$\allowbreak Neuro & informed learning (4.4) & Constitutional AI, RLAIF & known \\
11 Rule-Governed Memory & Symbolic[Neuro] & -- & memory component \parencite{liu2025agentpatterns}, Reflexion & new as stated \\
12 Model Selects, Rules Write & Symbolic[Neuro] & -- & template-based generation \parencite{reiter2000nlg}, restricted retrieval answers & known in part \\
13 Fail-Safe Union & Symbolic[Neuro] & -- & deny-overrides \parencite{oasis2013xacml}, layered guardrails & new as stated \\
\bottomrule
\end{tabularx}
\end{table}

\subsection{The patterns in boxology notation}\label{sec:boxology}

The boxology began with \textcite{vanharmelen2019boxology}: ovals for processes, either
deductive (KR) or inductive (ML), and boxes for the symbols or data that processes consume
and produce. \Textcite{vanbekkum2021modular} refined it into a vocabulary of four kinds of
node, instances (rectangles, data or symbols), models (hexagons, statistical or
semantic), processes (ovals: generate, transform, infer) and actors (triangles), and into
elementary patterns that compose. \Textcite{deboer2025patterns} added the generative model,
with an elementary pattern in which a model infers new data from data; the earlier
patterns let a model infer only symbols or models, which cannot describe a language model
that writes text.

Every pattern in this catalogue carries its structure in this notation, with
\emph{model:gen} for the language model and \emph{model:sem} for the rule base, and one
mark of ours, the dashed request arrow (\cref{sec:moves} and the legend in the catalogue
form).

The drawings show what the notation captures and what it leaves out.
\begin{itemize}[leftmargin=*]
\item In the Sensor (\cref{pat:sensor}) the model's reply is text, that is data; it becomes symbols only
through a transformation, which is our receipt, and the rules then infer which symbols
count. The notation shows this shape. It cannot show that the first symbol box holds
proposals and the second accepted facts, refusals with their reason and \emph{unknown}
values; the labels below the boxes add that.
\item In Model as Author (\cref{pat:author}) the generative model produces a semantic model, the draft
rule, which an inference with the checker and the cases evaluates; an expert adopts it
through the elementary pattern in which an actor engineers a model.
\item In Rules Steer the Model (\cref{pat:steer}) the rules derive the step, one generative model
phrases the question, a person answers and a second generative model judges the answer.
The closest named pattern is meta-reasoning for control, a symbolic process that directs a
learned one; here it does so step by step through a conversation.
\item In Rules Teach the Model (\cref{pat:teach}) the labels derived by the rules enter the training of the
model, the shape of informed learning with prior knowledge.
\item Model Proposes, Rules Dispose (\cref{pat:dispose}), Critique and Repair
(\cref{pat:repair}) and Rule-Governed Memory (\cref{pat:memory}) are loops; the notation
draws the arrow back, but the bound, the order of the steps and the supersession of one
value by a later one are rules that the drawing cannot show.
\end{itemize}
Three things the boxes and arrows cannot say are the subject of this catalogue: the
direction of control, which the dashed arrows add; the status of a symbol (proposed,
accepted, refused, \emph{unknown}) with its date and source; and repetition over time, a
retry, a next step or a later supersession. \Textcite{deboer2025patterns} name the last
gap themselves, and they observe that the notation often yields several drawings of the
same system. The four elementary moves of \cref{sec:moves} supply all three: a request
carries control, a receipt carries date and source, and accept and ask carry the status.

\subsection{What the catalogue adds}

\paragraph{Direction.} Most published systems let the model drive and call a solver or a
tool: Kautz's Neuro[Symbolic], ReAct \parencite{yao2023react}, Toolformer
\parencite{schick2023toolformer}, Logic-LM and LLM-Modulo. All thirteen patterns here are
Symbolic[Neuro]: the rules issue requests and the model is a subroutine whose answers
return as data. We chose this direction deliberately, for accountability. The agenda of
\textcite{garcez2023third} names trust and accountability as aims of tight integration,
and loose coupling with explicit provenance reaches both without tight integration.

\paragraph{Time and iteration.} \Textcite{deboer2025patterns} state that temporal,
recurring or iterative aspects ``have not yet been taken into account and can currently
not be visualised adequately'' in the boxology, and that the same system can be drawn in
several ways. Patterns 5, 6 and 8 are iterative by nature, and the dated receipt gives each
iteration an identity, an order and a record. Pattern 11 makes time its subject:
supersession, withdrawal and the current value as a query are rules over that order.

\paragraph{Three values and receipts.} None of the sources above distinguishes a failed call, a
malformed reply, an \emph{unclear} judgement and a logical \emph{unknown}, and none of them makes
every model answer a replayable observation. These are the invariants of
\cref{sec:foundations}, and they are what lets patterns 6 and 7 route by result and leave the
model's confidence aside.

\paragraph{Patterns without precedent.} Rules that derive every step of a conversation from
their state (5) are closest to programmable dialogue rails \parencite{rebedea2023nemo},
whose flows are written by hand. Recurring model answers that become rules (7)
appear only as offline rule extraction. Asking for the evidence that decides between
surviving hypotheses (8) is absent from abduction with language models
\parencite{shi2023lamp,wang2024hypothesis}. A model that reviews the rules themselves (9)
appears in none of the catalogues we read; in all of them the model is the one being
checked. The measurements of \cref{sec:guide} go beyond the validation the catalogues report,
which is coverage of literature \parencite{vanharmelen2019boxology}, redrawing published
systems \parencite{deboer2025patterns}, or a systematic review with known uses
\parencite{liu2025agentpatterns}.

\subsection{What the catalogue lacks}

Six published patterns have no counterpart, or only an unnamed one; C1 is now
named as a variant of the Sensor and C5 has become pattern 11. We rank them by how
often they appear and how well they fit the invariants.

\begin{enumerate}[leftmargin=*,label=C\arabic*.]
\item \textbf{Several answers, one rule.} Ask several models or samples and let rules
adjudicate: agreement, majority with a count, \emph{unclear} on disagreement. Voting and
debate are agent patterns \parencite{liu2025agentpatterns}, parallelisation by voting is a
workflow \parencite{schluntz2024agents}, and self-consistency beat multi-agent debate at
the same number of answers \parencite{huang2024cannot}. Three-valued logic suits it: a
disagreement is simply \emph{unclear}. It is now named as a variant of the Sensor (\cref{pat:sensor}); with its own
measurements it may become a pattern of its own.
\item \textbf{Gate and route before the model.} Rules decide whether a request may reach a
model at all and which model or prompt it goes to: input rails \parencite{rebedea2023nemo},
routing \parencite{schluntz2024agents}. It is implicit in the triage flow, which sends only
the careful judgements of emergency signs to the stronger model; it should be named as a
variant of pattern 5. Hackathon submissions add a stricter form, a \emph{sovereign
gatekeeper}: a local model and rules decide what may reach an external model,
pseudonymise it on the way out and translate it back on the way in, and send nothing
when the local side fails. It makes the rules the guard of the data flow between two
models, and it is a candidate for a pattern of its own once it is built and measured.
\item \textbf{Rules evaluate the model.} Rules reconstruct or audit what a model concluded:
rational reconstruction \parencite{vanbekkum2021modular}, knowledge graphs for
interpretability \parencite{pan2024roadmap}, the agent evaluator
\parencite{liu2025agentpatterns}. The break-even and capability studies are this pattern run
offline; online it is the two-channel disagreement listed under pattern 9. It belongs with
pattern 10 as grading, or as a pattern of its own. Its labels are cheapest where the truth
holds by construction: a fault the code injected, a value read from an exact oracle such
as a barcode or a simulator; several hackathon submissions measured this way, which also
avoids Circular Evaluation (\cref{sec:failures}).
\item \textbf{Pooled critics.} Several critics (rules, a model, a person) behind one
controller that merges their findings into the next request
\parencite{kambhampati2024llmmodulo}. A variant of pattern 6.
\item \textbf{Rule-governed memory.} Retrieval and episodic memory admitted, superseded and
cited by rules \parencite{liu2025agentpatterns,shinn2023reflexion}. It is now Rule-Governed Memory, pattern 11 (\cref{pat:memory}), built from the Sensor
for storing and Rules Steer the Model for recall; DreamRAG lite implements the storing
half, the program of pattern 11 both halves.
\item \textbf{Specification elicitation and the person as participant.} The model helps a
person state a goal or a specification, and rules check that it is complete
\parencite{kambhampati2024llmmodulo,liu2025agentpatterns}; human feedback as a pattern of its
own \parencite{witschel2020human}. Here the person is a participant in every pattern; the
elicitation is the idea of interviewing an expert under pattern 2.
\end{enumerate}

Three families are outside the catalogue on purpose: the model as driver that calls tools
(ReAct, Toolformer, Neuro[Symbolic]); rules compiled into the network, Kautz's
Neuro$_\text{Symbolic}$ and the tight integration of \textcite{garcez2023third}; and training
itself.

\subsection{What the catalogue learns from them}

\begin{itemize}[leftmargin=*]
\item \textbf{Correctness comes from the sound check.} In LLM-Modulo, soundness is
inherited from the hard critics and completeness depends on the model; a model alone
produced executable plans in about 12\,\% of cases, and back-prompting from a sound
validator raised this to 82\,\% in one domain \parencite{kambhampati2024llmmodulo}. The same
statement holds for patterns 4 and 8.
\item \textbf{Self-correction needs an external signal.} Without external feedback,
self-correction lowered accuracy, and earlier gains rested on oracle stopping signals
\parencite{huang2024cannot}. The rules are that signal in pattern 6; citing self-refinement
\parencite{madaan2023selfrefine,shinn2023reflexion} as precedent requires this caveat.
\item \textbf{Judges have known biases, and rules can mitigate them.} A strong judge agreed
with people about as often as people agreed with each other, but gave the same verdict when
two answers were swapped only about two thirds of the time; references and examples reduced
the failures \parencite{zheng2023judge}. A rule can ask twice in both orders and require
agreement, and supply the reference.
\item \textbf{Complementary failures justify division.} A solver-backed method and
chain-of-thought failed on different cases \parencite{olausson2023linc}, which is the
condition under which pattern 7 pays.
\item \textbf{Safeguards cover known unknowns.} A rule safeguard around a model covers only the
failures its authors foresaw \parencite{washizaki2020sep4mla}; pair pattern 4
with 8 or 9 where unforeseen failures matter.
\item \textbf{Form and selection.} Agent patterns are written with context, problem, forces,
solution, consequences tied to quality attributes, known uses and related patterns, and come
with a decision model that is not empirically validated \parencite{liu2025agentpatterns};
practitioners advise to start with the simplest option and to use evaluator loops only where
the criteria are clear \parencite{schluntz2024agents}. The modular taxonomy separates
elementary from composed patterns \parencite{vanbekkum2021modular}; the thirteen patterns here are compositions of four elementary moves (request, receipt, accept, ask), made explicit in \cref{sec:moves}.
\end{itemize}

\section{Choosing a pattern}
\label{sec:guide}

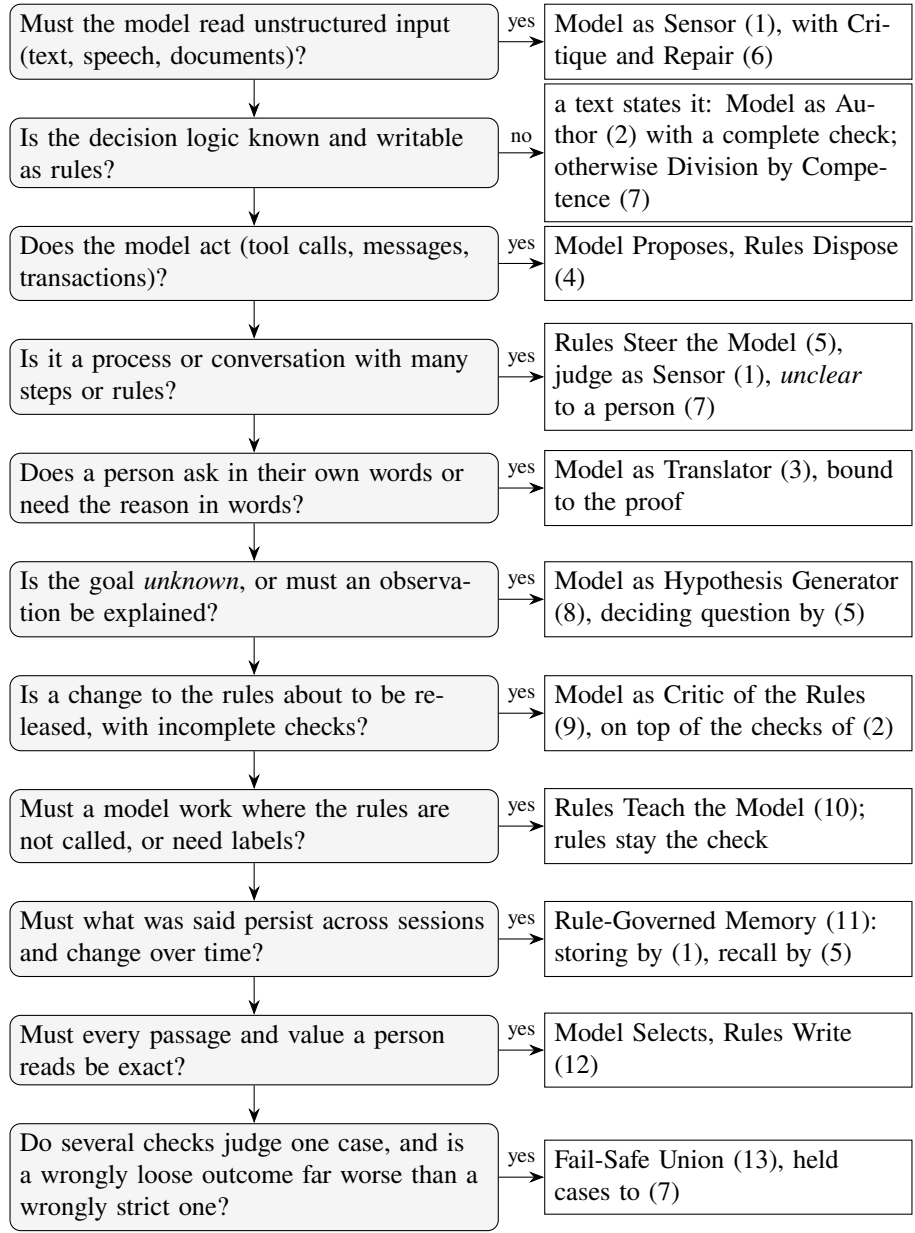
\begin{figure}[ht]
\centering
\begin{tikzpicture}[
  node distance=5mm and 6mm, font=\small,
  q/.style={draw, rounded corners, align=left, text width=6.2cm, fill=black!4},
  p/.style={draw, align=left, text width=4.6cm, fill=white},
  ->, >={Stealth[length=2mm]}]
\node[q] (q1) {Must the model read unstructured input (text, speech, documents)?};
\node[p, right=of q1] (p1) {Model as Sensor (1), with Critique and Repair (6)};
\node[q, below=of q1] (q2) {Is the decision logic known and writable as rules?};
\node[p, right=of q2] (p2) {a text states it: Model as Author (2) with a complete check; otherwise Division by Competence (7)};
\node[q, below=of q2] (q3) {Does the model act (tool calls, messages, transactions)?};
\node[p, right=of q3] (p3) {Model Proposes, Rules Dispose (4)};
\node[q, below=of q3] (q4) {Is it a process or conversation with many steps or rules?};
\node[p, right=of q4] (p4) {Rules Steer the Model (5), judge as Sensor (1), \emph{unclear} to a person (7)};
\node[q, below=of q4] (q5) {Does a person ask in their own words or need the reason in words?};
\node[p, right=of q5] (p5) {Model as Translator (3), bound to the proof};
\node[q, below=of q5] (q6) {Is the goal \emph{unknown}, or must an observation be explained?};
\node[p, right=of q6] (p6) {Model as Hypothesis Generator (8), deciding question by (5)};
\node[q, below=of q6] (q7) {Is a change to the rules about to be released, with incomplete checks?};
\node[p, right=of q7] (p7) {Model as Critic of the Rules (9), on top of the checks of (2)};
\node[q, below=of q7] (q8) {Must a model work where the rules are not called, or need labels?};
\node[p, right=of q8] (p8) {Rules Teach the Model (10); rules stay the check};
\node[q, below=of q8] (q9) {Must what was said persist across sessions and change over time?};
\node[p, right=of q9] (p9) {Rule-Governed Memory (11): storing by (1), recall by (5)};
\node[q, below=of q9] (q10) {Must every passage and value a person reads be exact?};
\node[p, right=of q10] (p10) {Model Selects, Rules Write (12)};
\node[q, below=of q10] (q11) {Do several checks judge one case, and is a wrongly loose outcome far worse than a wrongly strict one?};
\node[p, right=of q11] (p11) {Fail-Safe Union (13), held cases to (7)};
\foreach \i in {1,3,4,5,6,7,8,9,10,11} {\draw (q\i) -- node[above,font=\scriptsize]{yes} (p\i);}
\draw (q2) -- node[above,font=\scriptsize]{no} (p2);
\foreach \i/\j in {1/2,2/3,3/4,4/5,5/6,6/7,7/8,8/9,9/10,10/11} {\draw (q\i) -- (q\j);}
\end{tikzpicture}
\caption{Decision guide. Answer every question; the patterns of all answers that lead
right combine. Critique and Repair (6) accompanies every pattern whose reply is checked.}
\label{fig:guide}
\end{figure}

A pattern is chosen by the problem it solves and by the forces of the context. For
couplings the forces are measurable: how many rules a model would have to apply by
itself, whether a reply can be checked, whether the model acts, who must be able to
audit the result, and which model is available. This section first asks whether a
hybrid is needed at all, then leads from the problem to the patterns, and closes with
the measured results the guide rests on. Of the ways to select a pattern that
\textcite[sec.~1.7]{gamma1994design} list, two fit couplings: scanning the intents
(\cref{tab:overview}) and starting from the problem one has (\cref{fig:guide,tab:guide}).
Following practitioners' advice, the guide prefers the simplest coupling that meets the
requirement \parencite{schluntz2024agents}.

\subsection{First question: is a hybrid needed?}

A model alone was the better or equal choice in four measured situations, and the guide
says so first.
\begin{itemize}[leftmargin=*]
\item \textbf{Few rules.} In conversations of up to 48 questions, a reasoning model
applying the rules in its prompt was not measurably less accurate than the hybrid (58 or
more of 60 against 60 of 60); the accuracy break-even lay at 64 questions.
\item \textbf{Easy documents, a strong model.} On a clean five-document review, a model
alone decided 12 of 12 runs correctly, at 0.28 instead of 0.72 cents and in a third of
the time.
\item \textbf{Cost alone.} Once a provider's discount on cached input is priced, the
hybrid's cost advantage in conversations disappears; at 64 and 96 questions the two cost
about the same.
\item \textbf{A sensor that cannot quote.} An 8-billion-parameter open model extracted
worse in the hybrid than alone (45\,\% against 60\,\%), because a strict quote rule
refused correct values.
\end{itemize}
A hybrid is needed when at least one of the following holds: the decision must come with
a proof or an audit trail; the model acts and some actions must never happen; the rules
are many or the process is long (from about 64 rules applied in one conversation in our
runs); the answer aggregates over many objects (lists, counts, totals); a weaker, local
or cheaper model must do the work; or the rules change and the change must be checked
before it takes effect.

\subsection{From problem to pattern}

The questions are not exclusive: a triage line answers \emph{yes} to the first, third
and fourth and becomes a composition of 1, 4, 5, 6 and 7. \Cref{tab:guide} gives the
same guide as problems, with the force that decides and the caution that the
measurements add.

\begin{longtable}{@{}>{\raggedright\arraybackslash}p{3.6cm}>{\raggedright\arraybackslash}p{1.6cm}>{\raggedright\arraybackslash}p{4.6cm}>{\raggedright\arraybackslash}p{4.9cm}@{}}
\caption{Problems, patterns, deciding forces and cautions.}\label{tab:guide}\\
\toprule
Problem & Patterns & Deciding force & Caution from the measurements \\
\midrule\endfirsthead
\toprule
Problem & Patterns & Deciding force & Caution from the measurements \\
\midrule\endhead
Facts are in documents or speech & 1, 6 & a reading can be checked (quote, closed list) & fit the acceptance rule to the reader: a strict quote rule hurt a small model \\
Answer aggregates over many objects & 1, then rules & lists and counts are exact in rules & a model alone failed exhaustive lists at 100 agreements; extraction cost is paid once \\
A policy text must become rules & 2, 6 & a complete check exists (syntax, cell language, cases) & messages that name the fix doubled first-request repairs \\
A regulation changed & 2, 9 & regression cases decide; a person adopts & a draft outside the cell language is refused with its reason \\
People ask in their own words & 3 & the answer needs a proof & on easy cases a model alone was as accurate and cheaper \\
An agent acts & 4, 5, 6 & some effects must never happen & completion only from recorded evidence \\
A long protocol or conversation & 5, 1, 6, 7 & rules per conversation (break-even at 64) & a judge's verdict labels must not be readable as content \\
A weak or local model must suffice & 5 or 1 & the hybrid needs less capability & on our ladder a 9B model sufficed for the hybrid; no model up to 70B did as a pure model \\
Some cases are ambiguous & 7 & three-valued result as gate & never gate by the model's own confidence \\
Something must be explained & 8, 5 & generation needs world knowledge, testing needs exactness & incomplete search is never a ``no explanation'' \\
A soft criterion must be judged & 1, candidate C1 (\cref{sec:related}) & no closed check exists & judges are biased by answer order; ask in both orders and require agreement \parencite{zheng2023judge} \\
Semantic faults matter & 9 & checks are incomplete & model reviews lost semantic faults as the rule base grew; a model critic never replaces a sound check \parencite{huang2024cannot} \\
A model must learn the domain & 10 & labels with proofs & no measured model run yet \\
What a person said must be remembered & 11, 1, 5 & sources differ in what they may establish; requests to forget & no measurement yet; similarity retrieval stays outside the rules \\
An answer must quote a source or state values exactly & 12 & the content exists as passages or facts & hackathon figures only; the model can still choose the wrong passage \\
Several checks, harms of unequal weight & 13, 7 & a wrongly loose outcome costs far more than a wrongly strict one & hackathon figures only; false alarms of every source add up \\
\bottomrule
\end{longtable}

\subsection{What the guide rests on}

The cautions come from pre-registered studies in which designs and criteria were fixed
before the runs.
\begin{description}[leftmargin=1.2em,font=\normalfont\bfseries\small]
\item[Break-even.] Synthetic intake protocols of 4 to 96 questions; the hybrid decided
770 of 770 runs correctly; a model applying the rules itself fell to 39 of 60 at 64
questions and 31 of 60 at 96, with 45 of its 50 errors unsafe.
\item[Capability.] Without reasoning, a model decided 10 of 30 conversations as a pure
model and 30 of 30 in the hybrid. On a ladder of six local models from 1 to 70 billion
parameters, the hybrid's floor in single decisions was a 9-billion-parameter model;
the pure model had no floor on the ladder, and for it reasoning mattered more than size.
\pagebreak
\item[Checking.] With one seeded fault per artefact, static checks over rules found every
structural fault at every size at no model cost; model reviews of code and prompts lost
unreachable rules and missing rules as size grew.
\item[Authoring.] Five of five drafting runs ended with a rule passing all hidden cases;
error messages that name the fix decided how many repairs were needed.
\item[Aggregation.] At 100 agreements, a model alone produced no exact exhaustive list at
low effort; the hybrid answered 10 of 10.
\item[Decomposition.] With an 8-billion-parameter open model, classifying into 77 intents
by first choosing a group and then a label, as rules direct, gained 6 points over one
prompt with all labels; putting the groups into one prompt instead collapsed.
\end{description}

The literature adds three cautions that the guide follows: correctness is inherited from
the sound check \parencite{kambhampati2024llmmodulo}; a model's
self-correction without an external signal does not help \parencite{huang2024cannot}; and a
rule safeguard covers the failures its authors foresaw \parencite{washizaki2020sep4mla}.

\section{Limits}
\label{sec:limits}

We drew the catalogue from one engine and its applications and then checked it against
the literature. It covers every coupling we have seen so far around a symbolic, offline
core, and that does not prove it complete. Patterns 12 and 13 come from hackathon
submissions; their figures are taken from result files committed in those
repositories, and we have not rerun them. Every pattern up to 11 has a program that ran against a model; the
programs of patterns 12 and 13 are checked against recorded receipts only. Patterns 4, 7,
8, 10, 11, 12 and 13 have no FrameX measurement of their own yet, and the
measurement of pattern 9 is a pilot on easy cases. Most measurements use synthetic
protocols and one model family; the capability ladder mixes model families and reasoning modes. Couplings that
need a different kind of core (constraint solving and plan search, probabilistic
inference over model scores, differentiable logic inside the model, training itself) are
outside the catalogue. The rules can check what such a core produces, as they check any
external tool.

\appendix

\section{Boxology legend}\label{sec:legend}

\Cref{tab:legend} lists every element of the Structure diagrams. The node kinds and
their labels follow \textcite{vanbekkum2021modular}, the generative model follows
\textcite{deboer2025patterns}; the request arrow and the notes are ours.

\begin{longtable}{@{}>{\centering\arraybackslash}m{2.6cm}>{\raggedright\arraybackslash}m{3.4cm}>{\raggedright\arraybackslash}m{8.6cm}@{}}
\caption{Elements of the boxology diagrams.}\label{tab:legend}\\
\toprule
Symbol & Name & Meaning in this catalogue \\
\midrule\endfirsthead
\toprule
Symbol & Name & Meaning in this catalogue \\
\midrule\endhead
\tikz[bx]\node[bxinst]{data:\\text}; & instance: data & Model-free input or output, here text: a document, a question, a model's reply before it is read as facts. \\[2pt]
\tikz[bx]\node[bxinst]{symbol}; & instance: symbol & Model-based structure: facts, requests, proposals, verdicts. A note below the box gives its status, such as \emph{proposed} or \emph{accepted}. \\[2pt]
\tikz[bx]\node[bxmodel]{model:\\gen}; & model: generative (the LLM) & A language model that infers new data from data \parencite{deboer2025patterns}; the note names its role (dialogue model, judge, critic). \\[2pt]
\tikz[bx]\node[bxmodel]{model:\\sem}; & model: semantic (the FrameX rule base) & A rule base: the FrameX rules, a flow, a slot policy, a checker with its cases. \\[2pt]
\tikz[bx]\node[bxproc]{infer}; & process: infer & Applying a model to an instance. With a generative model above it, one model call. \\[2pt]
\tikz[bx]\node[bxproc]{infer:\\deduce}; & process: infer, deduce & Applying the rules: deriving, accepting, refusing, deciding. \\[2pt]
\tikz[bx]\node[bxproc]{trans-\\form}; & process: transform & Turning one kind of instance into another. Here mostly the receipt: the host turns a reply into symbols and adds its measurements (quote found, decision word). \\[2pt]
\tikz[bx]\node[bxproc]{generate:\\train}; & process: generate, train & Building a model from instances (pattern 10). \\[2pt]
\tikz[bx]\node[bxproc]{generate:\\engineer}; & process: generate, engineer & A person builds or adopts a model, such as a rule (pattern 2). \\[2pt]
\tikz[bx]\node[bxactor, label=below:{\tiny person}]{}; & actor & A person who answers, decides or reviews: applicant, expert, officer, reviewer. \\[2pt]
\tikz[bx]\draw[bxflow](0,0)--(9mm,0); & flow & What a process consumes or produces; from a model to a process, the model that the process applies. \\[2pt]
\tikz[bx]\draw[bxreq](0,0)--(9mm,0); & request (ours) & Control: the inference with the rules issues the call that the process at the arrow's head carries out. It marks the external requests of \cref{sec:coupling}. \\[2pt]
{\tiny\itshape\color{black!65} note} & note (ours) & The role of a node in the pattern, written below or above it. \\
\bottomrule
\end{longtable}

\section{Pattern libraries}\label{sec:libraries}

A library holds the rules of a pattern, and a program that loads it states facts and only
the rules that belong to its domain, in the way the flow library of FrameX serves
pattern~5. The companion folder has one library per pattern up to 11 (\texttt{library/}); patterns 12 and 13 have none yet. Each library comes
with the program of its pattern in library form, tests, and a recorded run. Load the
library before the program:
\begin{lstlisting}
$FRAMEX_HOME/target/release/framex test library/sensor.fx library/p01-sensor-lib.fx \
    library/tests/p01-sensor-lib.test.fx
python3 -m framex.tools library/memory.fx library/p11-memory-lib.fx
python3 -m framex.tools --adapter author_adapter.py library/author.fx library/p02-author-lib.fx
\end{lstlisting}
Every name a library defines carries its prefix (\texttt{sensor\#}, \texttt{author\#},
\texttt{tr\#}, \texttt{ag\#}, \texttt{cp\#}, \texttt{hy\#}, \texttt{cr\#}, \texttt{te\#},
\texttt{mem\#}), so it cannot collide with the program's own names. What stays with the
program is what only the domain knows: the decision of pattern~3, what the actions of
pattern~4 do, which cases pattern~7 decides, the tests of pattern~8, the labels of
pattern~10. The request fields keep fixed names; a library meant for other sources would
make them parameters. The libraries are prototypes in the companion folder and are meant to
move into FrameX beside \texttt{flow.fx}.

\subsection*{\texttt{sensor.fx}: pattern 1, Model as Sensor}
\begin{lstlisting}[style=fxprog,language=framex]
// Sensor library (pattern 1, Model as Sensor). Load it before the program.
// The program states facts only:
//   ?D:sensor#Source[sensor#text -> "..."; sensor#kind -> ?K; sensor#prompt -> ?P]
//   ?K[sensor#allows -> "slot"]          what this kind of source may state
//   ?P:Prompt[text -> "..."]             asks for rows {slot, value, quote}
// The library derives, for every source ?D and answer row ?I:
//   ?D[sensor#accepted(?I) -> ?V]        quoted, and a statement this kind may make
//   ?D[sensor#rejected(?I) -> "reason"]  with the reason
//   ?D[sensor#notRead -> "error"]        the call failed; nothing proposed
closed slot sensor#allows.

// Request: one call per source; the host checks every quote against the text.
external sense(?D) using _llm[prompt -> ?P; document -> ?T; groundIn -> ?T]
  <- ?D:sensor#Source[sensor#text -> ?T; sensor#prompt -> ?Pr] AND ?Pr[text -> ?P].

// Receipt: each answer row is a proposed statement about the source.
?D[sensor#slot(?I) -> ?S; sensor#value(?I) -> ?V; sensor#quoteFound(?I) -> ?Q]
  <- ?D:sensor#Source AND ?R[tool#name -> "_llm"; tool#arg1 -> ?D;
        row(?I, "slot") -> ?S; row(?I, "value") -> ?V; row(?I, "quoteFound") -> ?Q].

// Accept and reject.
?D[sensor#accepted(?I) -> ?V]
  <- ?D:sensor#Source[sensor#kind -> ?K; sensor#slot(?I) -> ?S; sensor#value(?I) -> ?V;
        sensor#quoteFound(?I) -> true] AND ?K[sensor#allows -> ?S].

?D[sensor#rejected(?I) -> "quote not found in the source"]
  <- ?D:sensor#Source[sensor#quoteFound(?I) -> false].

?D[sensor#rejected(?I) -> "not a statement this kind of source can make"]
  <- ?D:sensor#Source[sensor#kind -> ?K; sensor#slot(?I) -> ?S]
     AND NOT ?K[sensor#allows -> ?S].

// A failed call proposes nothing.
?D[sensor#notRead -> ?E]
  <- ?D:sensor#Source AND ?R[tool#name -> "_llm"; tool#arg1 -> ?D;
        tool#status -> "failed"; error -> ?E].
\end{lstlisting}

\subsection*{\texttt{author.fx}: pattern 2, Model as Author, with Critique and Repair}
\begin{lstlisting}[style=fxprog,language=framex]
// Author library (pattern 2, Model as Author, with pattern 6, Critique and Repair).
// Load it before the program; the host needs the adapter for _checkCandidate.
// The program states facts only:
//   ?P:author#Policy[author#text -> "..."; author#vocabulary -> "...";
//       author#draftPrompt -> ?D; author#repairPrompt -> ?R; author#maxAttempts -> N]
// The library derives author#rule(N), author#failed(N) with the checker's message,
// author#adoptable (a recommendation for a person), author#stalled and author#gaveUp.

// Attempt 1: the model drafts.
external draft(?P, 1) using _llm[prompt -> ?Pt; policy -> ?T; vocabulary -> ?V;
        tool#order -> 2]
  <- ?P:author#Policy[author#text -> ?T; author#vocabulary -> ?V; author#draftPrompt -> ?D]
     AND ?D[text -> ?Pt].

?P[author#rule(?N) -> ?T]
  <- ?P:author#Policy AND ?R[tool#name -> "_llm"; tool#arg1 -> ?P; tool#arg2 -> ?N;
        result("rule") -> ?T].

// Every draft is checked; the cases stay hidden from the model.
external check(?P, ?N) using _checkCandidate[candidate -> ?T; tool#order -> 1]
  <- ?P:author#Policy[author#rule(?N) -> ?T].

?P[author#passed(?N) -> true]
  <- ?P:author#Policy AND ?R[tool#name -> "_checkCandidate"; tool#arg1 -> ?P; tool#arg2 -> ?N;
        result("status") -> "checked"; result("regressions") -> false;
        result("failedAfter") -> 0].

?P[author#adoptable -> ?T] <- ?P:author#Policy[author#passed(?N) -> true; author#rule(?N) -> ?T].

?P[author#failed(?N) -> ?F]
  <- ?P:author#Policy AND ?R[tool#name -> "_checkCandidate"; tool#arg1 -> ?P; tool#arg2 -> ?N;
        result("status") -> ?St; result("feedback") -> ?F] AND ?St != "checked".

?P[author#failed(?N) -> ?F]
  <- ?P:author#Policy AND ?R[tool#name -> "_checkCandidate"; tool#arg1 -> ?P; tool#arg2 -> ?N;
        result("regressions") -> true; result("feedback") -> ?F].

?P[author#failed(?N) -> ?F]
  <- ?P:author#Policy AND ?R[tool#name -> "_checkCandidate"; tool#arg1 -> ?P; tool#arg2 -> ?N;
        result("failedAfter") -> ?K; result("feedback") -> ?F] AND ?K > 0.

// Repair (pattern 6): ask again with the rule and the problem, as a new request.
?P[author#newProblem(1) -> true] <- ?P:author#Policy[author#failed(1) -> ?F].

?P[author#newProblem(?N) -> true]
  <- ?P:author#Policy[author#failed(?N) -> ?F] AND ?K := ?N - 1
     AND ?P[author#failed(?K) -> ?G] AND ?F != ?G.

external draft(?P, ?M) using _llm[prompt -> ?Pt; policy -> ?T; vocabulary -> ?V;
        previousRule -> ?Old; problem -> ?F; tool#order -> 2]
  <- ?P:author#Policy[author#text -> ?T; author#vocabulary -> ?V; author#maxAttempts -> ?Max;
        author#repairPrompt -> ?Rp; author#rule(?N) -> ?Old; author#failed(?N) -> ?F;
        author#newProblem(?N) -> true]
     AND ?Rp[text -> ?Pt] AND ?N < ?Max AND ?M := ?N + 1.

?P[author#stalled -> ?N]
  <- ?P:author#Policy[author#failed(?N) -> ?F] AND ?K := ?N - 1 AND ?P[author#failed(?K) -> ?F].

?P[author#gaveUp -> ?F] <- ?P:author#Policy[author#maxAttempts -> ?N; author#failed(?N) -> ?F].
\end{lstlisting}

\subsection*{\texttt{translator.fx}: pattern 3, Model as Translator}
\begin{lstlisting}[style=fxprog,language=framex]
// Translator library (pattern 3, Model as Translator). Load it before the program.
// The program states:
//   ?Q:tr#Question[tr#text -> "..."; tr#prompt -> ?P]      the person's words
//   "field"[tr#allows -> "value"]                         the closed list of each field
//   tr#phrasing[tr#prompt -> ?P]                           how a decision is put into words
// and its own decision rules over valid fields only:
//   ?Q[tr#decision -> "word"] <- ?Q:tr#Question[tr#valid("f") -> ?V; ...] AND ...
// The library derives tr#field("f"), tr#valid("f"), tr#refused with the field,
// tr#reply or tr#replyRefused.
closed slot tr#allows. closed slot tr#hasBad.

external translate(?Q) using _llm[prompt -> ?P; question -> ?T]
  <- ?Q:tr#Question[tr#text -> ?T; tr#prompt -> ?Pr] AND ?Pr[text -> ?P].

// Every answer field the program declares, read from the receipt.
?Q[tr#field(?F) -> ?V]
  <- ?Q:tr#Question AND ?F[tr#allows -> ?X]
     AND ?R[tool#name -> "_llm"; tool#arg1 -> ?Q; result(?F) -> ?V].

// A field inside its closed list is valid; decision rules read only valid fields.
?Q[tr#valid(?F) -> ?V] <- ?Q:tr#Question[tr#field(?F) -> ?V] AND ?F[tr#allows -> ?V].

// A field outside its closed list refuses the request and names the field.
?Q[tr#bad(?F) -> true] <- ?Q:tr#Question[tr#field(?F) -> ?V] AND NOT ?F[tr#allows -> ?V].
?Q[tr#refused -> ?F] <- ?Q:tr#Question[tr#bad(?F) -> true].
?Q[tr#hasBad -> true] <- ?Q:tr#Question[tr#bad(?F) -> true].

// For display only: every declared field valid. Decision rules must not read it, because
// it negates over receipts, and the reply request depends on the decision.
?Q[tr#runs -> true]
  <- ?Q:tr#Question AND ?R[tool#name -> "_llm"; tool#arg1 -> ?Q; result(?F) -> ?V]
     AND NOT ?Q[tr#hasBad -> true].

// From the decision to words; the reply counts only if it says the decision word.
external phrase(?Q) using _llm[prompt -> ?P; decision -> ?D; mustSay -> ?D]
  <- ?Q:tr#Question[tr#decision -> ?D] AND tr#phrasing[tr#prompt -> ?Pr] AND ?Pr[text -> ?P].

?Q[tr#reply -> ?Text]
  <- ?Q:tr#Question AND ?X[tool#name -> "_llm"; tool#arg1 -> ?Q;
        result("saysDecision") -> true; result("text") -> ?Text].

?Q[tr#replyRefused -> "the reply does not state the decision"]
  <- ?Q:tr#Question AND ?X[tool#name -> "_llm"; tool#arg1 -> ?Q;
        result("saysDecision") -> false].
\end{lstlisting}

\subsection*{\texttt{agent.fx}: pattern 4, Model Proposes, Rules Dispose}
\begin{lstlisting}[style=fxprog,language=framex]
// Agent library (pattern 4, Model Proposes, Rules Dispose). Load it before the program.
// The program states:
//   ?T:ag#Task[ag#goal -> "..."; ag#actions -> "..."; ag#maxSteps -> N; ag#prompt -> ?P]
//   ?T[ag#grants -> "action"]                           what the task may do
// and, for each granted action, the rules that carry it out and report:
//   ?T[ag#outcome(?N) -> "..."]                         what became of step ?N
//   ?T[ag#evidence -> true]                             what completion rests on
// The library derives the proposals, admission or refusal with a reason, the next step,
// and ?T[ag#verdict -> "completed"] or "stopped at the step bound".
closed slot ag#grants.

external propose(?T, ?N) using _llm[prompt -> ?P; goal -> ?G; actions -> ?Acts;
        outcome -> "none"; tool#order -> 2]
  <- ?T:ag#Task[ag#goal -> ?G; ag#actions -> ?Acts; ag#prompt -> ?Pr] AND ?Pr[text -> ?P]
     AND ?N := 1.

?R:ag#Proposal[ag#task -> ?T; ag#step -> ?N; ag#action -> ?K]
  <- ?T:ag#Task AND ?R[tool#name -> "_llm"; tool#arg1 -> ?T; tool#arg2 -> ?N;
        result("action") -> ?K].

?A[ag#allowed -> true]
  <- ?A:ag#Proposal[ag#task -> ?T; ag#action -> ?K] AND ?T[ag#grants -> ?K].

?A[ag#refused -> "action not granted for this task"]
  <- ?A:ag#Proposal[ag#task -> ?T; ag#action -> ?K] AND NOT ?T[ag#grants -> ?K].

// The next step carries the outcome of the last one; maxSteps stops the loop.
?T[ag#next(?M) -> ?W]
  <- ?A:ag#Proposal[ag#task -> ?T; ag#step -> ?N; ag#refused -> ?W]
     AND ?T[ag#maxSteps -> ?Max] AND ?N < ?Max AND ?M := ?N + 1.

?T[ag#next(?M) -> ?X]
  <- ?A:ag#Proposal[ag#task -> ?T; ag#step -> ?N; ag#allowed -> true]
     AND ?T[ag#outcome(?N) -> ?X; ag#maxSteps -> ?Max] AND ?N < ?Max AND ?M := ?N + 1.

external propose(?T, ?N) using _llm[prompt -> ?P; goal -> ?G; actions -> ?Acts;
        outcome -> ?W; tool#order -> 2]
  <- ?T:ag#Task[ag#goal -> ?G; ag#actions -> ?Acts; ag#prompt -> ?Pr; ag#next(?N) -> ?W]
     AND ?Pr[text -> ?P].

// Completion only on recorded evidence.
?T[ag#verdict -> "completed"]
  <- ?A:ag#Proposal[ag#task -> ?T; ag#action -> "finish"; ag#allowed -> true]
     AND ?T[ag#evidence -> true].

?T[ag#verdict -> "stopped at the step bound"]
  <- ?A:ag#Proposal[ag#task -> ?T; ag#step -> ?N; ag#action -> ?K] AND ?T[ag#maxSteps -> ?N]
     AND ?K != "finish".
\end{lstlisting}

\subsection*{\texttt{competence.fx}: pattern 7, Division by Competence}
\begin{lstlisting}[style=fxprog,language=framex]
// Competence library (pattern 7, Division by Competence). Load it before the program.
// The program states the cases and its own rules for what it decides:
//   ?C:cp#Case[cp#file -> "..."; cp#prompt -> ?P]
//   ?C[cp#decided -> true] <- ...                       the cases the rules cover
// The library sends every other case to the model for an assessment and hands it on to a
// person, ?C[cp#handOn -> true], with the assessment as ?C[cp#note -> "..."].
closed slot cp#decided.

external assess(?C) using _llm[prompt -> ?P; file -> ?T]
  <- ?C:cp#Case[cp#file -> ?T; cp#prompt -> ?Pr] AND ?Pr[text -> ?P]
     AND NOT ?C[cp#decided -> true].

?C[cp#assessment -> ?V]
  <- ?C:cp#Case AND ?R[tool#name -> "_llm"; tool#arg1 -> ?C; result("verdict") -> ?V].

?C[cp#handOn -> true] <- ?C:cp#Case AND NOT ?C[cp#decided -> true].

?C[cp#note -> ?W]
  <- ?C:cp#Case[cp#handOn -> true] AND ?R[tool#name -> "_llm"; tool#arg1 -> ?C;
        result("reason") -> ?W].

?C[cp#note -> "no assessment: the model call failed"]
  <- ?C:cp#Case[cp#handOn -> true] AND ?R[tool#name -> "_llm"; tool#arg1 -> ?C;
        tool#status -> "failed"].
\end{lstlisting}

\subsection*{\texttt{hypothesis.fx}: pattern 8, Model as Hypothesis Generator}
\begin{lstlisting}[style=fxprog,language=framex]
// Hypothesis library (pattern 8, Model as Hypothesis Generator). Load it before the program.
// The program states the observation and its own tests:
//   ?O:hy#Observation[hy#text -> "..."; hy#prompt -> ?P; hy#label -> "key"]
//   ?O[hy#refuted(?I) -> true] <- ... hy#candidate(?I, "key") ...   what rules out a candidate
// The prompt asks for rows, one per candidate; hy#label names the row field that a
// person is asked about. The library derives the candidates, the survivors and
// ?O[hy#askAbout(?I) -> "..."]. A candidate never becomes a fact.
closed method hy#refuted/1.

external propose(?O) using _llm[prompt -> ?P; observation -> ?T]
  <- ?O:hy#Observation[hy#text -> ?T; hy#prompt -> ?Pr] AND ?Pr[text -> ?P].

// Every field of every row is kept, keyed by row and field name.
?O[hy#candidate(?I, ?K) -> ?V]
  <- ?O:hy#Observation AND ?R[tool#name -> "_llm"; tool#arg1 -> ?O; tool#status -> "succeeded";
        row(?I, ?K) -> ?V].

?O[hy#survives(?I) -> true]
  <- ?O:hy#Observation[hy#label -> ?L; hy#candidate(?I, ?L) -> ?E]
     AND NOT ?O[hy#refuted(?I) -> true].

?O[hy#askAbout(?I) -> ?E]
  <- ?O:hy#Observation[hy#label -> ?L; hy#survives(?I) -> true; hy#candidate(?I, ?L) -> ?E].
\end{lstlisting}

\subsection*{\texttt{critic.fx}: pattern 9, Model as Critic of the Rules}
\begin{lstlisting}[style=fxprog,language=framex]
// Critic library (pattern 9, Model as Critic of the Rules). Load it before the program.
// The program states the change:
//   ?C:cr#Change[cr#rationale -> "..."; cr#review -> "..."; cr#prompt -> ?P;
//       cr#reviewed -> true; cr#changed -> "case"]
//   ?C[cr#disposed(?I) -> true]          a reviewer has dealt with remark ?I
// The library derives grounded remarks cr#remark(?I), cr#blocked for a failed call, and
// cr#approved once the critique ran and every remark is disposed of.
closed method cr#disposed/1. closed slot cr#hasOpen. closed slot cr#critiqued.

external critique(?C) using _llm[prompt -> ?P; rationale -> ?Ra; review -> ?Rv; groundIn -> ?Ra]
  <- ?C:cr#Change[cr#rationale -> ?Ra; cr#review -> ?Rv; cr#prompt -> ?Pr] AND ?Pr[text -> ?P].

// A remark counts only with a found quote and a case the change really changed.
?C[cr#remark(?I) -> ?K]
  <- ?C:cr#Change[cr#changed -> ?K] AND ?R[tool#name -> "_llm"; tool#arg1 -> ?C;
        row(?I, "case") -> ?K; row(?I, "quoteFound") -> true].

?C[cr#critiqued -> true]
  <- ?C:cr#Change AND ?R[tool#name -> "_llm"; tool#arg1 -> ?C; tool#status -> "succeeded"].

?C[cr#blocked -> ?E]
  <- ?C:cr#Change AND ?R[tool#name -> "_llm"; tool#arg1 -> ?C; tool#status -> "failed";
        error -> ?E].

?C[cr#hasOpen -> true] <- ?C:cr#Change[cr#remark(?I) -> ?K] AND NOT ?C[cr#disposed(?I) -> true].

?C[cr#approved -> true]
  <- ?C:cr#Change[cr#reviewed -> true; cr#critiqued -> true] AND NOT ?C[cr#hasOpen -> true].
\end{lstlisting}

\subsection*{\texttt{teach.fx}: pattern 10, Rules Teach the Model}
\begin{lstlisting}[style=fxprog,language=framex]
// Teach library (pattern 10, Rules Teach the Model). Load it before the program.
// The program states the cases and its own labelling rules:
//   ?C:te#Case[te#description -> "..."; te#prompt -> ?P]
//   ?C[te#label -> "decision"] <- ...    ?C[te#gap -> "requirement"] <- ...
// The prompt asks for {"decision": ..., "rows": [{"gap": ...}]}. The library derives one
// te#Answer per case and grades it: te#right, te#missedGap.
closed slot te#named.

external answer(?C) using _llm[prompt -> ?P; description -> ?T]
  <- ?C:te#Case[te#description -> ?T; te#prompt -> ?Pr] AND ?Pr[text -> ?P].

?R:te#Answer[te#case -> ?C; te#decision -> ?D]
  <- ?C:te#Case AND ?R[tool#name -> "_llm"; tool#arg1 -> ?C; result("decision") -> ?D].

?R[te#named -> ?G] <- ?R:te#Answer AND ?R[row(?I, "gap") -> ?G].

?A[te#right -> true] <- ?A:te#Answer[te#case -> ?C; te#decision -> ?D] AND ?C[te#label -> ?D].

?A[te#missedGap -> ?G]
  <- ?A:te#Answer[te#case -> ?C] AND ?C[te#gap -> ?G] AND NOT ?A[te#named -> ?G].
\end{lstlisting}

\subsection*{\texttt{memory.fx}: pattern 11, Rule-Governed Memory}
\begin{lstlisting}[style=fxprog,language=framex]
// Memory library (pattern 11, Rule-Governed Memory). Load it before the program.
// The program states facts only:
//   ?K[mem#mayEstablish -> "slot"]                        what a kind of source may establish
//   ?U:mem#Utterance[mem#kind -> ?K; mem#seq -> N; mem#text -> "..."; mem#prompt -> ?P]
//   ?S:mem#Step[mem#about -> "who"; mem#needs -> "slot"]   what one step may see
//   ?Q:mem#Question[mem#step -> ?S; mem#text -> "..."; mem#prompt -> ?P]
//   ?U[mem#forgotten -> true]                             a person's request to forget
// The library derives statements st(?U, ?I) with mem#accepted, mem#rejected, mem#current,
// the context of each step, and ?Q[mem#answer -> "..."] or ?Q[mem#refused -> ...].
closed slot mem#mayEstablish. closed slot mem#forgotten. closed slot mem#hasNewer.
closed slot mem#current. closed slot mem#hasBad.
closed method mem#sentId/1.

// Storing (pattern 1): one statement object per quoted answer row.
external remember(?U) using _llm[prompt -> ?P; utterance -> ?T; groundIn -> ?T; tool#order -> 2]
  <- ?U:mem#Utterance[mem#text -> ?T; mem#prompt -> ?Pr] AND ?Pr[text -> ?P]
     AND NOT ?U[mem#forgotten -> true].

st(?U, ?I):mem#Statement[mem#from -> ?U; mem#about -> ?A; mem#slot -> ?S; mem#value -> ?V]
  <- ?U:mem#Utterance AND ?R[tool#name -> "_llm"; tool#arg1 -> ?U; row(?I, "about") -> ?A;
        row(?I, "slot") -> ?S; row(?I, "value") -> ?V; row(?I, "quoteFound") -> true].

// Admission, rejection, supersession.
?St[mem#accepted -> true]
  <- ?St:mem#Statement[mem#from -> ?U; mem#slot -> ?S] AND ?U[mem#kind -> ?K]
     AND ?K[mem#mayEstablish -> ?S] AND NOT ?U[mem#forgotten -> true].

?St[mem#rejected -> "this kind of source cannot establish it"]
  <- ?St:mem#Statement[mem#from -> ?U; mem#slot -> ?S] AND ?U[mem#kind -> ?K]
     AND NOT ?K[mem#mayEstablish -> ?S].

?St[mem#hasNewer -> true]
  <- ?St:mem#Statement[mem#accepted -> true; mem#from -> ?U; mem#about -> ?A; mem#slot -> ?S]
     AND ?U[mem#seq -> ?N]
     AND ?Nw:mem#Statement[mem#accepted -> true; mem#from -> ?W; mem#about -> ?A; mem#slot -> ?S]
     AND ?W[mem#seq -> ?M] AND ?M > ?N.

?St[mem#current -> true] <- ?St:mem#Statement[mem#accepted -> true] AND NOT ?St[mem#hasNewer -> true].

// Forgetting also reaches the answer store.
external forget(?U) using _forgetAnswers[text -> ?T; tool#order -> 1]
  <- ?U:mem#Utterance[mem#forgotten -> true; mem#text -> ?T].

// Recall (pattern 5): a step sees the current statements it needs, and nothing else.
?St[selectedFor -> ?K]
  <- ?K:mem#Step[mem#about -> ?A; mem#needs -> ?S]
     AND ?St:mem#Statement[mem#current -> true; mem#about -> ?A; mem#slot -> ?S].

external answer(?Q) using _llm[prompt -> ?P; question -> ?T; contextKey -> ?K; tool#order -> 3]
  <- ?Q:mem#Question[mem#text -> ?T; mem#step -> ?K; mem#prompt -> ?Pr] AND ?Pr[text -> ?P].

// Citations: every cited id must have been sent, and its statement must be current.
?R[mem#sentId(?C) -> true] <- ?R[tool#sent(?C) -> ?St].
?Q[mem#cites -> ?St]
  <- ?Q:mem#Question AND ?R[tool#arg1 -> ?Q; row(?I, "cited") -> ?C; tool#sent(?C) -> ?St].
?Q[mem#refused -> ?C]
  <- ?Q:mem#Question AND ?R[tool#arg1 -> ?Q; row(?I, "cited") -> ?C]
     AND NOT ?R[mem#sentId(?C) -> true].
?Q[mem#refused -> ?St] <- ?Q:mem#Question[mem#cites -> ?St] AND NOT ?St[mem#current -> true].
?Q[mem#hasBad -> true] <- ?Q:mem#Question[mem#refused -> ?X].
?Q[mem#answer -> ?Text]
  <- ?Q:mem#Question[mem#cites -> ?St] AND ?R[tool#arg1 -> ?Q; result("text") -> ?Text]
     AND NOT ?Q[mem#hasBad -> true].
\end{lstlisting}

\section{Rule graph legend}\label{sec:rglegend}

\Cref{tab:rglegend} lists the elements of the rule graphs and the flow diagram. The
script \texttt{rulegraph.py} of the companion folder draws them from a program with
Graphviz: \texttt{python3 rulegraph.py p01-sensor.fx > p01.dot}, then
\texttt{dot -Tpdf p01.dot -o p01.pdf}.

\tikzset{lgbox/.style={draw, rounded corners=2pt, minimum width=9mm, minimum height=4mm, font=\tiny\ttfamily}}
\begin{longtable}{@{}>{\centering\arraybackslash}m{2.6cm}>{\raggedright\arraybackslash}m{12cm}@{}}
\caption{Elements of the rule graphs and the flow diagram.}\label{tab:rglegend}\\
\toprule
Symbol & Meaning \\
\midrule\endfirsthead
\toprule
Symbol & Meaning \\
\midrule\endhead
\tikz\node[lgbox, fill=black!5]{Document}; & Facts of a class or object, with their slots, and the prompt objects. \\[2pt]
\tikz\node[lgbox, dashed]{read \textrightarrow{} \_llm}; & An external request and its tool. \\[2pt]
\tikz\node[lgbox, fill=orange!12]{receipt}; & The receipt the host writes, with the fields the rules read: \texttt{result("k")}, \texttt{row("k")}, \texttt{tool\#status}, \texttt{error}. \\[2pt]
\tikz\node[lgbox, fill=blue!8]{accepted(I)}; & Slots or classes that the rules derive. A box lists the slots of one rule head. \\[2pt]
\tikz\node[draw, circle, inner sep=0.5pt, font=\tiny\sffamily]{l.\,23}; & One rule. The label is the line of the program listing where the rule starts: ``l.~23'' is the rule in line 23. \\[2pt]
\tikz\node[draw, dashed, circle, inner sep=0.5pt, font=\tiny\sffamily]{l.\,14}; & A rule that issues an external request, labelled by its line in the same way. \\[2pt]
\tikz\draw[->, >={Stealth[length=1.5mm]}](0,0)--(9mm,0); & Into a rule: the rule reads it. Out of a rule: the rule derives it. \\[2pt]
\tikz\draw[dashed, -{Bar[width=2mm]}](0,0)--(9mm,0); & The rule reads it under \texttt{NOT}. \\[2pt]
\tikz\draw[dashed, ->, >={Stealth[length=1.5mm]}](0,0)--(9mm,0); & The rule issues the request. \\[2pt]
\tikz\draw[dotted, thick, ->, >={Stealth[length=1.5mm]}](0,0)--(9mm,0); & The host carries out the request and writes the receipt. \\[2pt]
\midrule
\tikz\node[lgbox, fill=blue!8, font=\tiny\sffamily]{Collect}; & Flow diagram: a step that asks a question and lets a judge decide its criterion. \\[2pt]
\tikz\node[lgbox, fill=black!5, font=\tiny\sffamily]{Say}; & Flow diagram: a step that shows a text. \\[2pt]
\tikz{\fill (0,0) circle (1.2pt); \draw (6mm,0) circle (1.4pt) (6mm,0) circle (2.2pt);} & Start and end of the flow. \\[2pt]
\tikz\draw[dashed, ->, >={Stealth[length=1.5mm]}](0,0) arc (200:-20:3mm); & A rejected attempt: the question is asked again while attempts remain. \\
\bottomrule
\end{longtable}

\printbibliography

@book{gamma1994design,
  author = {Gamma, Erich and Helm, Richard and Johnson, Ralph and Vlissides, John},
  title = {Design Patterns: Elements of Reusable Object-Oriented Software},
  publisher = {Addison-Wesley}, address = {Reading, MA}, year = {1994}}

@book{alexander1977pattern,
  author = {Alexander, Christopher and Ishikawa, Sara and Silverstein, Murray and Jacobson, Max and Fiksdahl-King, Ingrid and Angel, Shlomo},
  title = {A Pattern Language: Towns, Buildings, Construction},
  publisher = {Oxford University Press}, address = {New York}, year = {1977}}

@book{buschmann1996posa,
  author = {Buschmann, Frank and Meunier, Regine and Rohnert, Hans and Sommerlad, Peter and Stal, Michael},
  title = {Pattern-Oriented Software Architecture, Volume 1: A System of Patterns},
  publisher = {Wiley}, address = {Chichester}, year = {1996}}

@article{kifer1995flogic,
  author = {Kifer, Michael and Lausen, Georg and Wu, James},
  title = {Logical Foundations of Object-Oriented and Frame-Based Languages},
  journal = {Journal of the ACM}, volume = {42}, number = {4}, pages = {741--843}, year = {1995},
  doi = {10.1145/210332.210335}}

@article{vanharmelen2019boxology,
  author = {van Harmelen, Frank and ten Teije, Annette},
  title = {A Boxology of Design Patterns for Hybrid Learning and Reasoning Systems},
  journal = {Journal of Web Engineering}, volume = {18}, number = {1--3}, pages = {97--124}, year = {2019},
  doi = {10.13052/jwe1540-9589.18133}}

@article{vanbekkum2021modular,
  author = {van Bekkum, Michael and de Boer, Maaike and van Harmelen, Frank and Meyer-Vitali, Andr{\'e} and ten Teije, Annette},
  title = {Modular Design Patterns for Hybrid Learning and Reasoning Systems: A Taxonomy, Patterns and Use Cases},
  journal = {Applied Intelligence}, volume = {51}, number = {9}, pages = {6528--6546}, year = {2021},
  doi = {10.1007/s10489-021-02394-3}}

@article{deboer2025patterns,
  author = {de Boer, Maaike and Smit, Quirine and van Bekkum, Michael and Meyer-Vitali, Andr{\'e} and Schmid, Thomas},
  title = {Design Patterns for Large Language Model Based Neuro-Symbolic Systems},
  journal = {Neurosymbolic Artificial Intelligence}, volume = {1}, eid = {29498732251377499}, year = {2025},
  doi = {10.1177/29498732251377499}}

@article{kautz2022third,
  author = {Kautz, Henry A.},
  title = {The Third {AI} Summer: {AAAI} Robert S. Engelmore Memorial Lecture},
  journal = {AI Magazine}, volume = {43}, number = {1}, pages = {105--125}, year = {2022},
  doi = {10.1002/aaai.12036}}

@article{pan2024roadmap,
  author = {Pan, Shirui and Luo, Linhao and Wang, Yufei and Chen, Chen and Wang, Jiapu and Wu, Xindong},
  title = {Unifying Large Language Models and Knowledge Graphs: A Roadmap},
  journal = {IEEE Transactions on Knowledge and Data Engineering}, volume = {36}, number = {7},
  pages = {3580--3599}, year = {2024}, doi = {10.1109/TKDE.2024.3352100}}

@article{garcez2023third,
  author = {d'Avila Garcez, Artur and Lamb, Lu{\'i}s C.},
  title = {Neurosymbolic {AI}: The 3rd Wave},
  journal = {Artificial Intelligence Review}, volume = {56}, number = {11}, pages = {12387--12406}, year = {2023},
  doi = {10.1007/s10462-023-10448-w}}

@inproceedings{yang2025nesy,
  author = {Yang, Xiao-Wen and Shao, Jie-Jing and Guo, Lan-Zhe and Zhang, Bo-Wen and Zhou, Zhi and Jia, Lin-Han and Dai, Wang-Zhou and Li, Yu-Feng},
  title = {Neuro-Symbolic Artificial Intelligence: Towards Improving the Reasoning Abilities of Large Language Models},
  booktitle = {Proceedings of the Thirty-Fourth International Joint Conference on Artificial Intelligence ({IJCAI}-25), Survey Track},
  pages = {10770--10778}, year = {2025}, doi = {10.24963/ijcai.2025/1195}}

@article{rani2026nesyllm,
  author = {Rani, Maneeha and Mishra, Bhupesh Kumar and Thakker, Dhavalkumar},
  title = {Neurosymbolic Large Language Models: A Survey of Symbolic Integration, Reasoning and Explainability},
  journal = {Information Systems Frontiers}, year = {2026}, note = {Online first},
  doi = {10.1007/s10796-026-10794-4}}

@incollection{witschel2020human,
  author = {Witschel, Hans Friedrich and Pande, Charuta and Martin, Andreas and Laurenzi, Emanuele and Hinkelmann, Knut},
  title = {Visualization of Patterns for Hybrid Learning and Reasoning with Human Involvement},
  booktitle = {New Trends in Business Information Systems and Technology},
  series = {Studies in Systems, Decision and Control}, pages = {193--204},
  publisher = {Springer}, address = {Cham}, year = {2020}, doi = {10.1007/978-3-030-48332-6_13}}

@inproceedings{kambhampati2024llmmodulo,
  author = {Kambhampati, Subbarao and Valmeekam, Karthik and Guan, Lin and Verma, Mudit and Stechly, Kaya and Bhambri, Siddhant and Saldyt, Lucas Paul and Murthy, Anil B.},
  title = {Position: {LLMs} Can't Plan, But Can Help Planning in {LLM}-Modulo Frameworks},
  booktitle = {Proceedings of the 41st International Conference on Machine Learning},
  series = {PMLR}, volume = {235}, pages = {22895--22907}, year = {2024}}

@article{liu2025agentpatterns,
  author = {Liu, Yue and Lo, Sin Kit and Lu, Qinghua and Zhu, Liming and Zhao, Dehai and Xu, Xiwei and Harrer, Stefan and Whittle, Jon},
  title = {Agent Design Pattern Catalogue: A Collection of Architectural Patterns for Foundation Model Based Agents},
  journal = {Journal of Systems and Software}, volume = {220}, eid = {112278}, year = {2025},
  doi = {10.1016/j.jss.2024.112278}}

@online{schluntz2024agents,
  author = {Schluntz, Erik and Zhang, Barry},
  title = {Building Effective Agents}, organization = {Anthropic}, date = {2024-12-19},
  url = {https://www.anthropic.com/research/building-effective-agents}, urldate = {2026-10-12}}

@inproceedings{pan2023logiclm,
  author = {Pan, Liangming and Albalak, Alon and Wang, Xinyi and Wang, William},
  title = {Logic-{LM}: Empowering Large Language Models with Symbolic Solvers for Faithful Logical Reasoning},
  booktitle = {Findings of the Association for Computational Linguistics: EMNLP 2023}, pages = {3806--3824}, year = {2023},
  doi = {10.18653/v1/2023.findings-emnlp.248}}

@inproceedings{olausson2023linc,
  author = {Olausson, Theo and Gu, Alex and Lipkin, Ben and Zhang, Cedegao and Solar-Lezama, Armando and Tenenbaum, Joshua and Levy, Roger},
  title = {{LINC}: A Neurosymbolic Approach for Logical Reasoning by Combining Language Models with First-Order Logic Provers},
  booktitle = {Proceedings of the 2023 Conference on Empirical Methods in Natural Language Processing}, pages = {5153--5176}, year = {2023},
  doi = {10.18653/v1/2023.emnlp-main.313}}

@inproceedings{ye2023satlm,
  author = {Ye, Xi and Chen, Qiaochu and Dillig, Isil and Durrett, Greg},
  title = {{SatLM}: Satisfiability-Aided Language Models Using Declarative Prompting},
  booktitle = {Advances in Neural Information Processing Systems}, volume = {36}, pages = {45548--45580}, year = {2023}}

@inproceedings{madaan2023selfrefine,
  author = {Madaan, Aman and Tandon, Niket and Gupta, Prakhar and Hallinan, Skyler and Gao, Luyu and Wiegreffe, Sarah and Alon, Uri and Dziri, Nouha and Prabhumoye, Shrimai and Yang, Yiming and Gupta, Shashank and Majumder, Bodhisattwa Prasad and Hermann, Katherine and Welleck, Sean and Yazdanbakhsh, Amir and Clark, Peter},
  title = {Self-Refine: Iterative Refinement with Self-Feedback},
  booktitle = {Advances in Neural Information Processing Systems}, volume = {36}, pages = {46534--46594}, year = {2023}}

@inproceedings{shinn2023reflexion,
  author = {Shinn, Noah and Cassano, Federico and Gopinath, Ashwin and Narasimhan, Karthik and Yao, Shunyu},
  title = {Reflexion: Language Agents with Verbal Reinforcement Learning},
  booktitle = {Advances in Neural Information Processing Systems}, volume = {36}, pages = {8634--8652}, year = {2023}}

@inproceedings{huang2024cannot,
  author = {Huang, Jie and Chen, Xinyun and Mishra, Swaroop and Zheng, Huaixiu Steven and Yu, Adams Wei and Song, Xinying and Zhou, Denny},
  title = {Large Language Models Cannot Self-Correct Reasoning Yet},
  booktitle = {International Conference on Learning Representations}, year = {2024},
  url = {https://openreview.net/forum?id=IkmD3fKBPQ}}

@inproceedings{gou2024critic,
  author = {Gou, Zhibin and Shao, Zhihong and Gong, Yeyun and Shen, Yelong and Yang, Yujiu and Duan, Nan and Chen, Weizhu},
  title = {{CRITIC}: Large Language Models Can Self-Correct with Tool-Interactive Critiquing},
  booktitle = {International Conference on Learning Representations}, year = {2024},
  url = {https://openreview.net/forum?id=Sx038qxjek}}

@online{bai2022constitutional,
  author = {Bai, Yuntao and Kadavath, Saurav and Kundu, Sandipan and Askell, Amanda and others},
  title = {Constitutional {AI}: Harmlessness from {AI} Feedback}, year = {2022},
  eprint = {2212.08073}, eprinttype = {arXiv}}

@inproceedings{lee2024rlaif,
  author = {Lee, Harrison and Phatale, Samrat and Mansoor, Hassan and Mesnard, Thomas and Ferret, Johan and Lu, Kellie Ren and Bishop, Colton and Hall, Ethan and Carbune, Victor and Rastogi, Abhinav and Prakash, Sushant},
  title = {{RLAIF} vs. {RLHF}: Scaling Reinforcement Learning from Human Feedback with {AI} Feedback},
  booktitle = {Proceedings of the 41st International Conference on Machine Learning},
  series = {PMLR}, volume = {235}, pages = {26874--26901}, year = {2024}}

@inproceedings{zheng2023judge,
  author = {Zheng, Lianmin and Chiang, Wei-Lin and Sheng, Ying and Zhuang, Siyuan and Wu, Zhanghao and Zhuang, Yonghao and Lin, Zi and Li, Zhuohan and Li, Dacheng and Xing, Eric and Zhang, Hao and Gonzalez, Joseph and Stoica, Ion},
  title = {Judging {LLM}-as-a-Judge with {MT}-Bench and Chatbot Arena},
  booktitle = {Advances in Neural Information Processing Systems, Datasets and Benchmarks Track}, volume = {36}, pages = {46595--46623}, year = {2023}}

@inproceedings{rebedea2023nemo,
  author = {Rebedea, Traian and Dinu, Razvan and Sreedhar, Makesh Narsimhan and Parisien, Christopher and Cohen, Jonathan},
  title = {{NeMo} Guardrails: A Toolkit for Controllable and Safe {LLM} Applications with Programmable Rails},
  booktitle = {Proceedings of the 2023 Conference on Empirical Methods in Natural Language Processing: System Demonstrations},
  pages = {431--445}, year = {2023}, doi = {10.18653/v1/2023.emnlp-demo.40}}

@inproceedings{yao2023react,
  author = {Yao, Shunyu and Zhao, Jeffrey and Yu, Dian and Du, Nan and Shafran, Izhak and Narasimhan, Karthik and Cao, Yuan},
  title = {{ReAct}: Synergizing Reasoning and Acting in Language Models},
  booktitle = {International Conference on Learning Representations}, year = {2023}}

@inproceedings{schick2023toolformer,
  author = {Schick, Timo and Dwivedi-Yu, Jane and Dess{\`i}, Roberto and Raileanu, Roberta and Lomeli, Maria and Hambro, Eric and Zettlemoyer, Luke and Cancedda, Nicola and Scialom, Thomas},
  title = {Toolformer: Language Models Can Teach Themselves to Use Tools},
  booktitle = {Advances in Neural Information Processing Systems}, volume = {36}, pages = {68539--68551}, year = {2023}}

@inproceedings{shi2023lamp,
  author = {Shi, Xiaoming and Xue, Siqiao and Wang, Kangrui and Zhou, Fan and Zhang, James and Zhou, Jun and Tan, Chenhao and Mei, Hongyuan},
  title = {Language Models Can Improve Event Prediction by Few-Shot Abductive Reasoning},
  booktitle = {Advances in Neural Information Processing Systems}, volume = {36}, pages = {29532--29557}, year = {2023}}

@inproceedings{wang2024hypothesis,
  author = {Wang, Ruocheng and Zelikman, Eric and Poesia, Gabriel and Pu, Yewen and Haber, Nick and Goodman, Noah D.},
  title = {Hypothesis Search: Inductive Reasoning with Language Models},
  booktitle = {International Conference on Learning Representations}, year = {2024},
  url = {https://openreview.net/forum?id=G7UtIGQmjm}}

@article{washizaki2022sepml,
  author = {Washizaki, Hironori and Khomh, Foutse and Gu{\'e}h{\'e}neuc, Yann-Ga{\"e}l and Takeuchi, Hironori and Natori, Naotake and Doi, Takuo and Okuda, Satoshi},
  title = {Software-Engineering Design Patterns for Machine Learning Applications},
  journal = {Computer}, volume = {55}, number = {3}, pages = {30--39}, year = {2022}, doi = {10.1109/MC.2021.3137227}}

@inproceedings{washizaki2020sep4mla,
  author = {Washizaki, Hironori and Khomh, Foutse and Gu{\'e}h{\'e}neuc, Yann-Ga{\"e}l and Takeuchi, Hironori and Okuda, Satoshi and Natori, Naotake and Shioura, Naohisa},
  title = {Software Engineering Patterns for Machine Learning Applications ({SEP4MLA}) -- Part 2},
  booktitle = {Proceedings of the 27th Conference on Pattern Languages of Programs ({PLoP} '20)},
  publisher = {The Hillside Group}, year = {2020}}

@book{reiter2000nlg,
  author    = {Reiter, Ehud and Dale, Robert},
  title     = {Building Natural Language Generation Systems},
  publisher = {Cambridge University Press},
  address   = {Cambridge},
  year      = {2000}
}

@techreport{oasis2013xacml,
  author      = {{OASIS}},
  title       = {{eXtensible Access Control Markup Language (XACML) Version 3.0}},
  institution = {OASIS},
  type        = {OASIS Standard},
  year        = {2013},
  month       = jan
}
\end{document}